\pdfoutput=1

\documentclass[11pt,hyphens]{article}

\usepackage[final]{acl}

\usepackage[utf8]{inputenc}
\usepackage[T1]{fontenc}
\usepackage{times}
\usepackage{latexsym}
\usepackage{inconsolata}
\usepackage{url}
\usepackage{booktabs}
\usepackage{amsfonts}
\usepackage{nicefrac}
\usepackage{microtype}

\usepackage{xspace}
\usepackage{graphicx}
\usepackage[normalem]{ulem}
\usepackage{amsmath}
\usepackage{enumitem}

\usepackage{multirow}
\usepackage{makecell}
\usepackage{ragged2e}
\usepackage{tabularx}

\usepackage{subcaption}

\usepackage{listings}

\usepackage{tikz}
\usetikzlibrary{calc}
\usetikzlibrary{shapes.geometric}
\usetikzlibrary{decorations.pathreplacing}
\usetikzlibrary{positioning}

\usepackage{twemojis}
\usepackage{datetime}
\usepackage{bbm}
\usepackage{xfp}
\usepackage{siunitx}
\definecolor{gray}{RGB}{211,211,211}
\newcommand{\jbasicstyle}{\small\sffamily}

\newcommand{\jnumberstyle}{\scriptsize}

\lstdefinelanguage{pseudo}
{
morekeywords={},
keywordstyle=\bfseries,
lineskip=-0.1em,
numbers=left,
numberstyle=\jnumberstyle,
numbersep=4pt,
basicstyle=\jbasicstyle,
breaklines=true,
breakautoindent=true,
tabsize=2,
columns=fullflexible,
morecomment=*[l][\textsl]{//},
mathescape=true,
xleftmargin=10pt,
}

\lstdefinelanguage{todo-comment}
{
morekeywords={},
keywordstyle=\bfseries,
lineskip=-0.1em,
numbers=none,
basicstyle=\jbasicstyle,
breaklines=true,
breakautoindent=true,
tabsize=2,
columns=fullflexible,
morecomment=*[l][\textsl]{//},
mathescape=true,
xleftmargin=-10pt,
}

\lstdefinelanguage{java-pretty}
{
language=java,
numbers=left,
basicstyle=\scriptsize\ttfamily,
numberstyle=\scriptsize,
breaklines=true,
columns=fullflexible,
xleftmargin=16pt,
showstringspaces=false,
}

\definecolor{rust-keyword}{rgb}{0.74, 0.2, 0.64}
\definecolor{rust-modifier}{rgb}{0.2, 0.2, 0.6}
\definecolor{rust-type}{rgb}{0.0, 0.42, 0.24}
\definecolor{rust-fn}{rgb}{0.25, 0.5, 0.35}
\definecolor{rust-comment}{rgb}{0.45, 0.45, 0.45}
\definecolor{rust-string}{rgb}{0.65, 0.16, 0.16}
\definecolor{gitgreen}{rgb}{0.0, 0.5, 0.0}
\definecolor{gitred}{rgb}{0.7, 0.0, 0.0}

\lstdefinelanguage{rust-pretty}
{
morekeywords=[1]{if, else, while, loop, for, match, break, continue, return, in, as, where},
keywordstyle=[1]\color{rust-keyword}\bfseries,
morekeywords=[2]{pub, mut, ref, const, static, let, move, async, await, unsafe, crate, use, mod, super, self, Self},
keywordstyle=[2]\color{rust-modifier}\bfseries,
morekeywords=[3]{bool, char, i8, i16, i32, i64, i128, u8, u16, u32, u64, u128, usize, isize, f32, f64, str, String, Vec, Option, Result, Box, Cursor},
keywordstyle=[3]\color{rust-type}\bfseries,
morekeywords=[4]{fn, struct, enum, trait, impl, type, dyn, Some, None, Ok, Err, true, false},
keywordstyle=[4]\color{rust-fn},
morecomment=[l]{//},
commentstyle=\color{rust-comment}\itshape,
morestring=[b]",
stringstyle=\color{rust-string},
numbers=left,
basicstyle=\scriptsize\ttfamily,
numberstyle=\scriptsize,
breaklines=true,
columns=fullflexible,
xleftmargin=14pt,
tabsize=2,
showstringspaces=false,
keepspaces=true,
}

\lstdefinelanguage{rust-diff}
{
language=rust-pretty,
morecomment=[f][\color{gitgreen}][0]{+},
morecomment=[f][\color{gitred}][0]{-},
escapeinside={(*@}{@*)},
showstringspaces=false,
keepspaces=true,
breaklines=true,
}

\lstdefinestyle{ContextSnippet}{
basicstyle=\footnotesize\ttfamily,
breaklines=true,
breakautoindent=true,
breakatwhitespace=false,
columns=fullflexible,
keepspaces=true,
showstringspaces=false,
xleftmargin=4pt,
}

\makeatletter
\newcommand{\DefMacro}{\@ifstar\@DefMacroAllowRedefine\@DefMacro}
\newcommand{\@DefMacro}[2]{\expandafter\newcommand\csname rmk-#1\endcsname{#2}}
\newcommand{\@DefMacroAllowRedefine}[2]{\expandafter\providecommand\csname rmk-#1\endcsname{} \expandafter\renewcommand\csname rmk-#1\endcsname{#2}}
\makeatother
\newcommand{\UseMacro}[1]{\csname rmk-#1\endcsname}

\newcommand{\BiggerBold}[2]{\ifnum\fpeval{(#1)>=(#2)}=1 \textbf{#1}\else #1\fi}
\newcommand{\SmallerBold}[2]{\ifnum\fpeval{(#1)<=(#2)}=1 \textbf{#1}\else #1\fi}
\makeatletter
\newcommand{\SmallerBoldSafe}[2]{%
\edef\@sbsa{#1}\edef\@sbsb{#2}%
\def\@sbsdash{--}%
\ifx\@sbsa\@sbsdash\@sbsa
\else\ifx\@sbsb\@sbsdash\@sbsa
\else\SmallerBold{\@sbsa}{\@sbsb}\fi\fi}
\makeatother
\makeatletter
\newcommand{\SmallestBoldSafe}[3]{%
\edef\@stba{#1}\edef\@stbb{#2}\edef\@stbc{#3}%
\def\@stbdash{--}%
\ifx\@stba\@stbdash\@stba
\else\ifx\@stbb\@stbdash\@stba
\else\ifx\@stbc\@stbdash\@stba
\else
\ifnum\fpeval{(\@stba)<=(\@stbb)}=1
\ifnum\fpeval{(\@stba)<=(\@stbc)}=1
\textbf{\@stba}%
\else\@stba\fi
\else\@stba\fi
\fi\fi\fi}
\makeatother

\makeatletter
\newcommand{\LargestBoldSafe}[3]{%
\edef\@ltba{#1}\edef\@ltbb{#2}\edef\@ltbc{#3}%
\def\@ltbdash{--}%
\ifx\@ltba\@ltbdash\@ltba
\else\ifx\@ltbb\@ltbdash\@ltba
\else\ifx\@ltbc\@ltbdash\@ltba
\else
\ifnum\fpeval{(\@ltba)>=(\@ltbb)}=1
\ifnum\fpeval{(\@ltba)>=(\@ltbc)}=1
\textbf{\@ltba}%
\else\@ltba\fi
\else\@ltba\fi
\fi\fi\fi}
\makeatother

\usepackage[textsize=tiny]{todonotes}

\newcommand{\XSpace}[1]{}
\newcommand{\XComment}[1]{}

\newcommand{\MyPara}[1]{\noindent\textbf{#1}.}

\newcommand{\Code}[1]{{\ifmmode{\mathtt{#1}}\else$\mathtt{#1}$\fi}}
\newcommand{\CodeIn}[1]{{\ifmmode{\mathtt{#1}}\else$\mathtt{#1}$\fi}}

\newcolumntype{R}[1]{>{\RaggedLeft\arraybackslash}p{#1}}
\newcolumntype{L}[1]{>{\RaggedRight\arraybackslash}p{#1}}

\newcommand{\InputWithSpace}[1]{\bgroup\def\arraystretch{1.2}\input{#1}\egroup}

\newcommand{\olmoThreeBare}[2]{\texttt{Olmo#1\,#2B}\xspace}
\newcommand{\qwen}{\texttt{Qwen3\!.\!5\,35B\,A3B}\xspace}
\newcommand{\gpto}{\texttt{GPT-OSS\,20B}\xspace}
\newcommand{\olmos}{\olmoThreeBare{3}{7}}
\newcommand{\olmol}{\olmoThreeBare{3\!.\!1}{32}}
\newcommand{\olmosfull}{\olmoThreeBare{3}{7} (Thinking)}
\newcommand{\olmolfull}{\olmoThreeBare{3\!.\!1}{32} (Thinking)}
\newcommand{\haiku}{\texttt{Haiku\,4\!.\!5}\xspace}

\newcommand{\Tool}{{\sc Crocodil}\xspace}

\newcommand{\rust}{Rust\xspace}
\newcommand{\github}{GitHub\xspace}
\newcommand{\selfedit}{self-edit\xspace}
\newcommand{\crossedit}{cross-edit\xspace}
\newcommand{\Crossedit}{Cross-edit\xspace}
\newcommand{\selfediting}{self-editing\xspace}
\newcommand{\crossediting}{cross-editing\xspace}
\newcommand{\Crossediting}{Cross-editing\xspace}
\newcommand{\othercode}{foreign code\xspace}
\newcommand{\rewriter}{implementor\xspace}
\newcommand{\rewriters}{implementors\xspace}
\newcommand{\reimplement}{implement\xspace}
\newcommand{\reimplements}{implements\xspace}
\newcommand{\reimplemented}{implemented\xspace}
\newcommand{\reimplementing}{implementing\xspace}
\newcommand{\reimplementation}{implementation\xspace}
\newcommand{\reimplementations}{implementations\xspace}
\newcommand{\Reimplementation}{Implementation\xspace}
\newcommand{\editor}{editor\xspace}
\newcommand{\editors}{editors\xspace}
\newcommand{\task}{task\xspace}
\newcommand{\tasks}{tasks\xspace}

\newcommand{\numofeditors}{five\xspace}
\newcommand{\numofopeneditors}{four\xspace}
\newcommand{\levmetric}{Levenshtein edit distance\xspace}
\newcommand{\basecommit}{base commit\xspace}
\newcommand{\headcommit}{head commit\xspace}
\newcommand{\maxcommit}{max commit\xspace}
\newcommand{\draft}{draft\xspace}
\newcommand{\drafts}{drafts\xspace}
\newcommand{\draftstage}{draft stage\xspace}
\newcommand{\repairstage}{repair stage\xspace}

\newcommand{\Funcsig}{Function signature\xspace}

\newcommand{\Restprdiff}{Rest-of-PR diff\xspace}

\newcommand{\Callees}{Callee functions\xspace}

\newcommand{\Callsites}{Call sites\xspace}

\newcommand{\Usestmts}{Use statements\xspace}

\newcommand{\Structimpl}{Struct and impl\xspace}

\newcommand{\Repodesc}{Repo description\xspace}
\newcommand{\prdesc}{PR description\xspace}
\newcommand{\PRdesc}{PR description\xspace}

\newcommand{\Funcdesc}{Function description\xspace}
\newcommand{\Implagent}{Patch Agent\xspace}
\newcommand{\implagent}{patch agent\xspace}
\newcommand{\Instagent}{Instruct Agent\xspace}
\newcommand{\instagent}{instruct agent\xspace}

\newcommand{\IDFuncsig}{\ensuremath{C_{\text{FuncSig}}}\xspace}
\newcommand{\IDRestprdiff}{\ensuremath{C_{\text{Diff}}}\xspace}

\newcommand{\IDCallees}{\ensuremath{C_{\text{Callees}}}\xspace}
\newcommand{\IDCallsites}{\ensuremath{C_{\text{CallSites}}}\xspace}
\newcommand{\IDUsestmts}{\ensuremath{C_{\text{Use}}}\xspace}
\newcommand{\IDStructimpl}{\ensuremath{C_{\text{Impl}}}\xspace}
\newcommand{\IDRepodesc}{\ensuremath{C_{\text{RepoDesc}}}\xspace}
\newcommand{\IDPRdesc}{\ensuremath{C_{\text{PRDesc}}}\xspace}
\newcommand{\IDFuncdesc}{\ensuremath{C_{\text{FuncDesc}}}\xspace}
\newcommand{\selfcode}{native code\xspace}
\newcommand{\before}{pre-edit\xspace}
\newcommand{\after}{post-edit\xspace}
\newcommand{\beforecode}{\before code\xspace}
\newcommand{\exec}{execution\xspace}
\newcommand{\simi}{similarity\xspace}
\newcommand{\Simi}{Similarity\xspace}
\newcommand{\simreward}{\simi reward\xspace}
\newcommand{\execreward}{\exec reward\xspace}
\newcommand{\Exec}{Execution\xspace}

\newcommand{\SimReward}{\Simi Reward\xspace}

\newcommand{\ExecReward}{\Exec Reward\xspace}
\newcommand{\rlmodel}{RL trained model\xspace}

\newcommand{\basemodel}{base model\xspace}

\newcommand{\minprompt}{strict prompt\xspace}
\newcommand{\baselabel}{\olmos~Base\xspace}
\newcommand{\strictlabel}{\olmos~Strict\xspace}
\newcommand{\rllabel}{\olmos~RL\xspace}
\newcommand{\basestrictpass}{Base \& Strict pass\xspace}

\newcommand{\Minprompt}{Strict prompt\xspace}
\newcommand{\LLMs}{LLMs\xspace}
\newcommand{\LLM}{LLM\xspace}
\newcommand{\heterogeneous}{multiple\xspace}
\newcommand{\PR}{PR\xspace}
\newcommand{\PRs}{PRs\xspace}
\newcommand{\hunk}{diff\xspace}

\newcommand{\TColStack}[2]{\shortstack{\texttt{#1}\\\texttt{#2}}}
\DefMacro{TCol-qwen}{\TColStack{Qwen3.5}{35B\,A3B}}
\DefMacro{TCol-gpto}{\TColStack{GPT-OSS}{20B}}
\DefMacro{TCol-olmos}{\TColStack{Olmo3}{7B}}
\DefMacro{TCol-olmol}{\TColStack{Olmo3.1}{32B}}
\DefMacro{TCol-haiku}{\TColStack{Haiku}{4.5}}

\DefMacro{TColInline-qwen}{\texttt{Qwen3.5}}
\DefMacro{TColInline-gpto}{\texttt{GPT-OSS}}
\DefMacro{TColInline-olmos}{\texttt{Olmo3}}
\DefMacro{TColInline-olmol}{\texttt{Olmo3.1}}
\DefMacro{TColInline-haiku}{\texttt{Haiku}}

\DefMacro{TColOneLine-qwen}{\texttt{Qwen3.5 35B\,A3B}}
\DefMacro{TColOneLine-gpto}{\texttt{GPT-OSS 20B}}
\DefMacro{TColOneLine-olmos}{\texttt{Olmo3 7B}}
\DefMacro{TColOneLine-olmol}{\texttt{Olmo3.1 32B}}
\DefMacro{TColOneLine-haiku}{\texttt{Haiku 4.5}}

\newcommand{\THStack}[3]{\shortstack{\textbf{#1}\\\textbf{#2}\\\textbf{#3}}}
\newcommand{\THStackTwo}[2]{\shortstack{\textbf{#1}\\\textbf{#2}}}
\DefMacro{TCol-meantest}{\THStack{Mean}{Tests}{(\%)}}
\DefMacro{TCol-meantest-wide}{\THStackTwo{Mean Tests}{(\%)}}
\DefMacro{TCol-meantest-nc}{\THStack{Mean Tests}{no change}{(\%)}}
\DefMacro{TCol-meantest-wc}{\THStack{Mean Tests}{with change}{(\%)}}
\DefMacro{TColInline-meantest}{\textbf{Mean Tests}}
\DefMacro{TColInline-meantest-nc}{\textbf{Mean Tests no change}}
\DefMacro{TColInline-meantest-wc}{\textbf{Mean Tests with change}}

\DefMacro{TCol-meantest-nc-share}{\THStack{Mean Tests}{but no change}{share (\%)}}
\DefMacro{TCol-meantest-wc-share}{\THStack{Mean Tests}{with change}{share (\%)}}
\DefMacro{TColInline-meantest-nc-share}{\textbf{Mean Tests but no change share}}
\DefMacro{TColInline-meantest-wc-share}{\textbf{Mean Tests with change share}}

\DefMacro{TCol-build}{\THStackTwo{Build}{(\%)}}
\DefMacro{TCol-build-nc}{\THStack{Build}{no change}{(\%)}}
\DefMacro{TCol-build-wc}{\THStack{Build}{with change}{(\%)}}
\DefMacro{TColInline-build}{\textbf{Build}}
\DefMacro{TColInline-build-nc}{\textbf{Build no change}}
\DefMacro{TColInline-build-wc}{\textbf{Build with change}}

\DefMacro{TCol-alltest}{\THStackTwo{All Tests}{(\%)}}
\DefMacro{TColInline-alltest}{\textbf{All Tests}}

\DefMacro{TCol-build-nc-share}{\THStack{Build but}{no change}{share (\%)}}
\DefMacro{TCol-build-wc-share}{\THStack{Build with}{change}{share (\%)}}
\DefMacro{TColInline-build-nc-share}{\textbf{Build but no change share}}
\DefMacro{TColInline-build-wc-share}{\textbf{Build with change share}}

\DefMacro{lora-trainable-params}{80M}

\DefMacro{TCap-pr-lev-minmax-norm-pivot-rewrite-char}{Per-\task character-level \levmetric, min-max normalized within each \task across the \numofeditors \editors and averaged over \tasks. The minimum in each column is bolded.}
\DefMacro{TCap-verl-comparison-all-mean-lev-rustfmt-eval-rust-repos}{Mean \levmetric from the \reimplementation to the edited function. \Tool roughly halves the edit distance across all \numofopeneditors open-weight \rewriters. The smaller value of each base/RL pair is bolded.}
\DefMacro{TCap-verl-lev-raw-rustfmt-eval-rust-repos}{Mean \levmetric from the \reimplementation to the edited function in absolute form, at character and line granularity. \textbf{All \tasks} reports over every \task. Each remaining group restricts to the \tasks that the two \editors it names both pass, and the row left out of a comparison shows dashes there. The smallest value of each comparison is bolded. A dagger marks a cell averaged over ten or fewer \tasks. Table~\ref{tab:verl-lev-normfull-rustfmt-eval-rust-repos} is the min-max normalized form of the same cells.}
\DefMacro{TCap-verl-lev-normfull-rustfmt-eval-rust-repos}{Min-max normalized \levmetric from the \reimplementation to the edited function, at character and line granularity, over the same \tasks as Table~\ref{tab:verl-lev-raw-rustfmt-eval-rust-repos}. Table~\ref{tab:verl-lev-norm-eval-rust-repos} prints the \textbf{Chars} column of each group from this table.}
\DefMacro{TCap-verl-lev-norm-eval-rust-repos}{Min-max normalized \levmetric from the \reimplementation to the edited function. \textbf{All \tasks} reports over every \task. Each \textbf{Both pass} group restricts to the \tasks that the two \editors it names both pass, and the row left out of a comparison shows dashes there. The smallest value of each comparison is bolded. A dagger marks a cell averaged over ten or fewer \tasks.}
\DefMacro{TCap-verl-lev-both-pass-rustfmt-eval-rust-repos}{\levmetric restricted to \tasks where edits by both base and \rlmodel pass every test. The \rlmodel still makes smaller edits while reaching the same successful outcome, so the reduction is not an artifact of producing worse edits.}
\DefMacro{TCap-verl-lev-rl-pass-rustfmt-eval-rust-repos}{\levmetric restricted to \tasks where the \rlmodel passed every test (\basemodel need not pass). The \rlmodel edits roughly a third as much as the \basemodel on this subset.}
\DefMacro{TCap-verl-comparison-all-mean-eval-rust-repos}{Edit success of the \basemodel vs.\ the \rlmodel per \rewriter. \UseMacro{TColInline-build} is the percentage of \tasks whose edit compiles, \UseMacro{TColInline-alltest} is the percentage of \tasks whose edit passes every collected test, and \UseMacro{TColInline-meantest} is the mean per-\task fraction of tests passed, counting an edit that fails to build as zero.}
\DefMacro{TCap-verl-meantestnc-eval-rust-repos}{The mean per-\task fraction of tests passed over three populations, every \task (\UseMacro{TColInline-meantest}), the \tasks where the \rlmodel returns the \reimplementation unchanged (\UseMacro{TColInline-meantest-nc}), and the \tasks where the \rlmodel edits it (\UseMacro{TColInline-meantest-wc}).}
\DefMacro{TCap-verl-buildnc-eval-rust-repos}{The share of \tasks whose edit builds, over three populations, every \task (\UseMacro{TColInline-build}), the \tasks where the \rlmodel returns the \reimplementation unchanged (\UseMacro{TColInline-build-nc}), and the \tasks where the \rlmodel edits it (\UseMacro{TColInline-build-wc}).}
\DefMacro{TCap-verl-buildncshare-eval-rust-repos}{Table~\ref{tab:verl-buildnc-eval-rust-repos} with the rate split instead of the \tasks. \UseMacro{TColInline-build} is unchanged, and it separates into the \tasks that build while the \reimplementation is left untouched (\UseMacro{TColInline-build-nc-share}) and the \tasks that build after a change is applied (\UseMacro{TColInline-build-wc-share}).}
\DefMacro{TCap-verl-meantestncshare-eval-rust-repos}{Table~\ref{tab:verl-meantestnc-eval-rust-repos} with the mean split instead of the \tasks, exactly as Table~\ref{tab:verl-buildncshare-eval-rust-repos} does for the build rate. \UseMacro{TColInline-meantest} is unchanged, and it separates into the fraction of tests passed on the \tasks the \editor leaves untouched (\UseMacro{TColInline-meantest-nc-share}) and on the \tasks it changes (\UseMacro{TColInline-meantest-wc-share}), both divided by all \tasks.}
\DefMacro{TCap-corpus-attrition-before}{Functions remaining after filters in Section~\ref{sec:char-crossediting}, before \reimplementation by the \LLM \rewriters.}
\DefMacro{TCap-corpus-attrition-after}{Functions remaining after each filter in Section~\ref{sec:char-crossediting}, after \reimplementation by the \LLM \rewriters. Each \rewriter, implements its own, so each has its own count. The bold row is the edit \tasks it contributes.}
\DefMacro{TCap-verl-trsplit-pooled-eval-rust-repos}{Passed test counts after excluding \tasks where either model makes no edit. The \rlmodel passes more tests on every \rewriter other than \olmos itself and fewer on its own \reimplementations, showing that smaller edits do not sacrifice correctness. The \minprompt is the \basemodel prompted for the smallest possible edit, on the same \tasks.}
\DefMacro{FCap-verl-trsplit-dist}{Per-\task test-pass ratio densities for base (orange) vs.\ RL (green). \Tool shifts mass from the zero spike toward the one spike on every \rewriter other than \olmos itself, converting more failed edits into fully passing ones, and moves the other way on \selfediting.}
\DefMacro{FCap-verl-venn}{Overlap between tasks where \basemodel (orange), \minprompt (blue), and \rlmodel (green) edits pass all tests. The restricted columns of Table~\ref{tab:verl-lev-norm-eval-rust-repos} are the pairwise overlaps bolded here.}
\DefMacro{FCap-instruct-plan-example-candidate}{The failing candidate the \instagent{} was asked to diagnose. \texttt{rustc} rejects the first line with \texttt{E0449}, visibility qualifiers are not permitted on a trait method.}
\DefMacro{FCap-instruct-plan-example-plan}{The plan the \instagent{} produced for it, in full.}
\DefMacro{FCap-instruct-plan-example}{A repair round on \texttt{fmt} in \texttt{fancy-regex/fancy-regex} \#120, with \olmol as the \rewriter. The plan names the bug and describes the fix in prose.}
\DefMacro{FCap-prompt-edit-limit-guidelines}{The guideline the \minprompt adds to the \editor system prompt.}
\DefMacro{TCap-pr-edit-pass-rate-pivot}{Percentage of \tasks whose edit passes every test, for each (\rewriter, \editor) pair of Table~\ref{tab:pr-lev-minmax-norm-pivot-rewrite-char}. The maximum in each column is bolded.}
\DefMacro{TCap-pr-lev-selfcross-mwu}{One-sided Mann-Whitney U $p$ values for the off-diagonal cells of Table~\ref{tab:pr-lev-minmax-norm-pivot-rewrite-char}. Each cell tests whether the column \editor's normalized distances on its own \reimplementations are smaller than its distances on the row \rewriter's \reimplementations. Stars mark $p<0.05$ ($^{*}$), $p<0.01$ ($^{**}$), and $p<0.001$ ($^{***}$).}

\IfFileExists{numbers/verl/numbers_verl_lev_common_pass_macros.tex}{\input{numbers/verl/numbers_verl_lev_common_pass_macros.tex}}{}
\IfFileExists{numbers/verl/numbers_verl_lev_common_pass_minmax_macros.tex}{\input{numbers/verl/numbers_verl_lev_common_pass_minmax_macros.tex}}{}
\IfFileExists{numbers/verl/numbers_verl_lev_rl_pass_raw_macros.tex}{\input{numbers/verl/numbers_verl_lev_rl_pass_raw_macros.tex}}{}
\IfFileExists{numbers/verl/numbers_verl_lev_both_pass_minmax_macros.tex}{\input{numbers/verl/numbers_verl_lev_both_pass_minmax_macros.tex}}{}
\IfFileExists{numbers/verl/numbers_verl_comparison_all_mean_lev_rustfmt_macros.tex}{\DefMacro{fig-verl-comparison-allmean-rustfmt-base-qwen-lev-char}{413.29}
\DefMacro{fig-verl-comparison-allmean-rustfmt-base-qwen-lev-line}{16.59}
\DefMacro{fig-verl-comparison-allmean-rustfmt-prompt-qwen-lev-char}{399.82}
\DefMacro{fig-verl-comparison-allmean-rustfmt-prompt-qwen-lev-line}{15.66}
\DefMacro{fig-verl-comparison-allmean-rustfmt-verl-step62-qwen-lev-char}{233.28}
\DefMacro{fig-verl-comparison-allmean-rustfmt-verl-step62-qwen-lev-line}{8.47}
\DefMacro{fig-verl-comparison-allmean-rustfmt-base-gpt-lev-char}{413.54}
\DefMacro{fig-verl-comparison-allmean-rustfmt-base-gpt-lev-line}{16.38}
\DefMacro{fig-verl-comparison-allmean-rustfmt-prompt-gpt-lev-char}{446.19}
\DefMacro{fig-verl-comparison-allmean-rustfmt-prompt-gpt-lev-line}{17.18}
\DefMacro{fig-verl-comparison-allmean-rustfmt-verl-step62-gpt-lev-char}{200.82}
\DefMacro{fig-verl-comparison-allmean-rustfmt-verl-step62-gpt-lev-line}{8.23}
\DefMacro{fig-verl-comparison-allmean-rustfmt-base-olmo7b-lev-char}{290.72}
\DefMacro{fig-verl-comparison-allmean-rustfmt-base-olmo7b-lev-line}{10.89}
\DefMacro{fig-verl-comparison-allmean-rustfmt-prompt-olmo7b-lev-char}{354.80}
\DefMacro{fig-verl-comparison-allmean-rustfmt-prompt-olmo7b-lev-line}{12.46}
\DefMacro{fig-verl-comparison-allmean-rustfmt-verl-step62-olmo7b-lev-char}{165.86}
\DefMacro{fig-verl-comparison-allmean-rustfmt-verl-step62-olmo7b-lev-line}{6.74}
\DefMacro{fig-verl-comparison-allmean-rustfmt-base-olmo32b-lev-char}{376.13}
\DefMacro{fig-verl-comparison-allmean-rustfmt-base-olmo32b-lev-line}{13.69}
\DefMacro{fig-verl-comparison-allmean-rustfmt-prompt-olmo32b-lev-char}{437.67}
\DefMacro{fig-verl-comparison-allmean-rustfmt-prompt-olmo32b-lev-line}{14.96}
\DefMacro{fig-verl-comparison-allmean-rustfmt-verl-step62-olmo32b-lev-char}{221.14}
\DefMacro{fig-verl-comparison-allmean-rustfmt-verl-step62-olmo32b-lev-line}{8.14}
}{}
\IfFileExists{numbers/verl/numbers_verl_lev_rl_pass_rustfmt_macros.tex}{\input{numbers/verl/numbers_verl_lev_rl_pass_rustfmt_macros.tex}}{}
\IfFileExists{numbers/verl/numbers_verl_lev_both_pass_rustfmt_macros.tex}{\DefMacro{fig-verl-lev-both-pass-rustfmt-qwen-small}{}
\DefMacro{fig-verl-lev-both-pass-rustfmt-qwen-base-mean-char}{92.93}
\DefMacro{fig-verl-lev-both-pass-rustfmt-qwen-base-mean-line}{3.81}
\DefMacro{fig-verl-lev-both-pass-rustfmt-qwen-rl-mean-char}{74.22}
\DefMacro{fig-verl-lev-both-pass-rustfmt-qwen-rl-mean-line}{2.59}
\DefMacro{fig-verl-lev-both-pass-rustfmt-gpt-small}{}
\DefMacro{fig-verl-lev-both-pass-rustfmt-gpt-base-mean-char}{156.82}
\DefMacro{fig-verl-lev-both-pass-rustfmt-gpt-base-mean-line}{5.36}
\DefMacro{fig-verl-lev-both-pass-rustfmt-gpt-rl-mean-char}{105.77}
\DefMacro{fig-verl-lev-both-pass-rustfmt-gpt-rl-mean-line}{3.82}
\DefMacro{fig-verl-lev-both-pass-rustfmt-olmo7b-small}{\dag}
\DefMacro{fig-verl-lev-both-pass-rustfmt-olmo7b-base-mean-char}{106.11}
\DefMacro{fig-verl-lev-both-pass-rustfmt-olmo7b-base-mean-line}{4.33}
\DefMacro{fig-verl-lev-both-pass-rustfmt-olmo7b-rl-mean-char}{57.78}
\DefMacro{fig-verl-lev-both-pass-rustfmt-olmo7b-rl-mean-line}{2.00}
\DefMacro{fig-verl-lev-both-pass-rustfmt-olmo32b-small}{}
\DefMacro{fig-verl-lev-both-pass-rustfmt-olmo32b-base-mean-char}{94.00}
\DefMacro{fig-verl-lev-both-pass-rustfmt-olmo32b-base-mean-line}{4.23}
\DefMacro{fig-verl-lev-both-pass-rustfmt-olmo32b-rl-mean-char}{96.31}
\DefMacro{fig-verl-lev-both-pass-rustfmt-olmo32b-rl-mean-line}{4.08}
}{}
\IfFileExists{numbers/verl/numbers_verl_lev_prompt_pass_rustfmt_macros.tex}{\DefMacro{fig-verl-lev-prompt-pass-rustfmt-qwen-small}{}
\DefMacro{fig-verl-lev-prompt-pass-rustfmt-qwen-prompt-mean-char}{65.74}
\DefMacro{fig-verl-lev-prompt-pass-rustfmt-qwen-prompt-mean-line}{2.48}
\DefMacro{fig-verl-lev-prompt-pass-rustfmt-qwen-rl-mean-char}{72.19}
\DefMacro{fig-verl-lev-prompt-pass-rustfmt-qwen-rl-mean-line}{2.49}
\DefMacro{fig-verl-lev-prompt-pass-rustfmt-gpt-small}{}
\DefMacro{fig-verl-lev-prompt-pass-rustfmt-gpt-prompt-mean-char}{131.67}
\DefMacro{fig-verl-lev-prompt-pass-rustfmt-gpt-prompt-mean-line}{5.33}
\DefMacro{fig-verl-lev-prompt-pass-rustfmt-gpt-rl-mean-char}{86.28}
\DefMacro{fig-verl-lev-prompt-pass-rustfmt-gpt-rl-mean-line}{3.26}
\DefMacro{fig-verl-lev-prompt-pass-rustfmt-olmo7b-small}{\dag}
\DefMacro{fig-verl-lev-prompt-pass-rustfmt-olmo7b-prompt-mean-char}{39.00}
\DefMacro{fig-verl-lev-prompt-pass-rustfmt-olmo7b-prompt-mean-line}{2.20}
\DefMacro{fig-verl-lev-prompt-pass-rustfmt-olmo7b-rl-mean-char}{42.40}
\DefMacro{fig-verl-lev-prompt-pass-rustfmt-olmo7b-rl-mean-line}{2.00}
\DefMacro{fig-verl-lev-prompt-pass-rustfmt-olmo32b-small}{}
\DefMacro{fig-verl-lev-prompt-pass-rustfmt-olmo32b-prompt-mean-char}{111.39}
\DefMacro{fig-verl-lev-prompt-pass-rustfmt-olmo32b-prompt-mean-line}{5.14}
\DefMacro{fig-verl-lev-prompt-pass-rustfmt-olmo32b-rl-mean-char}{103.82}
\DefMacro{fig-verl-lev-prompt-pass-rustfmt-olmo32b-rl-mean-line}{4.54}
}{}
\IfFileExists{numbers/verl/numbers_verl_lev_prompt_pass_rustfmt_minmax_macros.tex}{\DefMacro{fig-verl-lev-prompt-pass-rustfmt-minmax-qwen-lev-char-small}{}
\DefMacro{fig-verl-lev-prompt-pass-rustfmt-minmax-prompt-qwen-lev-char}{0.37}
\DefMacro{fig-verl-lev-prompt-pass-rustfmt-minmax-rl-qwen-lev-char}{0.26}
\DefMacro{fig-verl-lev-prompt-pass-rustfmt-minmax-qwen-lev-line-small}{}
\DefMacro{fig-verl-lev-prompt-pass-rustfmt-minmax-prompt-qwen-lev-line}{0.35}
\DefMacro{fig-verl-lev-prompt-pass-rustfmt-minmax-rl-qwen-lev-line}{0.20}
\DefMacro{fig-verl-lev-prompt-pass-rustfmt-minmax-gpt-lev-char-small}{}
\DefMacro{fig-verl-lev-prompt-pass-rustfmt-minmax-prompt-gpt-lev-char}{0.42}
\DefMacro{fig-verl-lev-prompt-pass-rustfmt-minmax-rl-gpt-lev-char}{0.29}
\DefMacro{fig-verl-lev-prompt-pass-rustfmt-minmax-gpt-lev-line-small}{}
\DefMacro{fig-verl-lev-prompt-pass-rustfmt-minmax-prompt-gpt-lev-line}{0.42}
\DefMacro{fig-verl-lev-prompt-pass-rustfmt-minmax-rl-gpt-lev-line}{0.29}
\DefMacro{fig-verl-lev-prompt-pass-rustfmt-minmax-olmo7b-lev-char-n}{4}
\DefMacro{fig-verl-lev-prompt-pass-rustfmt-minmax-olmo7b-lev-char-small}{\dag}
\DefMacro{fig-verl-lev-prompt-pass-rustfmt-minmax-prompt-olmo7b-lev-char}{0.21}
\DefMacro{fig-verl-lev-prompt-pass-rustfmt-minmax-rl-olmo7b-lev-char}{0.29}
\DefMacro{fig-verl-lev-prompt-pass-rustfmt-minmax-olmo7b-lev-line-small}{\dag}
\DefMacro{fig-verl-lev-prompt-pass-rustfmt-minmax-prompt-olmo7b-lev-line}{0.25}
\DefMacro{fig-verl-lev-prompt-pass-rustfmt-minmax-rl-olmo7b-lev-line}{0.17}
\DefMacro{fig-verl-lev-prompt-pass-rustfmt-minmax-olmo32b-lev-char-small}{}
\DefMacro{fig-verl-lev-prompt-pass-rustfmt-minmax-prompt-olmo32b-lev-char}{0.30}
\DefMacro{fig-verl-lev-prompt-pass-rustfmt-minmax-rl-olmo32b-lev-char}{0.23}
\DefMacro{fig-verl-lev-prompt-pass-rustfmt-minmax-olmo32b-lev-line-small}{}
\DefMacro{fig-verl-lev-prompt-pass-rustfmt-minmax-prompt-olmo32b-lev-line}{0.26}
\DefMacro{fig-verl-lev-prompt-pass-rustfmt-minmax-rl-olmo32b-lev-line}{0.22}
\DefMacro{fig-verl-lev-prompt-pass-rustfmt-minmax-haiku-lev-char-small}{}
\DefMacro{fig-verl-lev-prompt-pass-rustfmt-minmax-prompt-haiku-lev-char}{0.35}
\DefMacro{fig-verl-lev-prompt-pass-rustfmt-minmax-rl-haiku-lev-char}{0.11}
}{}
\IfFileExists{numbers/verl/numbers_verl_lev_base_strict_rustfmt_macros.tex}{\DefMacro{fig-verl-lev-base-strict-rustfmt-qwen-small}{}
\DefMacro{fig-verl-lev-base-strict-rustfmt-qwen-base-mean-char}{71.83}
\DefMacro{fig-verl-lev-base-strict-rustfmt-qwen-base-mean-line}{2.92}
\DefMacro{fig-verl-lev-base-strict-rustfmt-qwen-prompt-mean-char}{82.45}
\DefMacro{fig-verl-lev-base-strict-rustfmt-qwen-prompt-mean-line}{3.08}
\DefMacro{fig-verl-lev-base-strict-rustfmt-gpt-small}{}
\DefMacro{fig-verl-lev-base-strict-rustfmt-gpt-base-mean-char}{123.03}
\DefMacro{fig-verl-lev-base-strict-rustfmt-gpt-base-mean-line}{4.33}
\DefMacro{fig-verl-lev-base-strict-rustfmt-gpt-prompt-mean-char}{126.83}
\DefMacro{fig-verl-lev-base-strict-rustfmt-gpt-prompt-mean-line}{5.25}
\DefMacro{fig-verl-lev-base-strict-rustfmt-olmo7b-small}{\dag}
\DefMacro{fig-verl-lev-base-strict-rustfmt-olmo7b-base-mean-char}{108.83}
\DefMacro{fig-verl-lev-base-strict-rustfmt-olmo7b-base-mean-line}{6.17}
\DefMacro{fig-verl-lev-base-strict-rustfmt-olmo7b-prompt-mean-char}{149.83}
\DefMacro{fig-verl-lev-base-strict-rustfmt-olmo7b-prompt-mean-line}{7.33}
\DefMacro{fig-verl-lev-base-strict-rustfmt-olmo32b-small}{}
\DefMacro{fig-verl-lev-base-strict-rustfmt-olmo32b-base-mean-char}{97.50}
\DefMacro{fig-verl-lev-base-strict-rustfmt-olmo32b-base-mean-line}{4.21}
\DefMacro{fig-verl-lev-base-strict-rustfmt-olmo32b-prompt-mean-char}{105.33}
\DefMacro{fig-verl-lev-base-strict-rustfmt-olmo32b-prompt-mean-line}{4.50}
}{}
\IfFileExists{numbers/verl/numbers_verl_lev_base_strict_rustfmt_minmax_macros.tex}{\DefMacro{fig-verl-lev-base-strict-rustfmt-minmax-qwen-lev-char-small}{}
\DefMacro{fig-verl-lev-base-strict-rustfmt-minmax-base-qwen-lev-char}{0.43}
\DefMacro{fig-verl-lev-base-strict-rustfmt-minmax-prompt-qwen-lev-char}{0.45}
\DefMacro{fig-verl-lev-base-strict-rustfmt-minmax-qwen-lev-line-small}{}
\DefMacro{fig-verl-lev-base-strict-rustfmt-minmax-base-qwen-lev-line}{0.43}
\DefMacro{fig-verl-lev-base-strict-rustfmt-minmax-prompt-qwen-lev-line}{0.41}
\DefMacro{fig-verl-lev-base-strict-rustfmt-minmax-gpt-lev-char-small}{}
\DefMacro{fig-verl-lev-base-strict-rustfmt-minmax-base-gpt-lev-char}{0.34}
\DefMacro{fig-verl-lev-base-strict-rustfmt-minmax-prompt-gpt-lev-char}{0.40}
\DefMacro{fig-verl-lev-base-strict-rustfmt-minmax-gpt-lev-line-small}{}
\DefMacro{fig-verl-lev-base-strict-rustfmt-minmax-base-gpt-lev-line}{0.28}
\DefMacro{fig-verl-lev-base-strict-rustfmt-minmax-prompt-gpt-lev-line}{0.37}
\DefMacro{fig-verl-lev-base-strict-rustfmt-minmax-olmo7b-lev-char-small}{\dag}
\DefMacro{fig-verl-lev-base-strict-rustfmt-minmax-base-olmo7b-lev-char}{0.18}
\DefMacro{fig-verl-lev-base-strict-rustfmt-minmax-prompt-olmo7b-lev-char}{0.32}
\DefMacro{fig-verl-lev-base-strict-rustfmt-minmax-olmo7b-lev-line-small}{\dag}
\DefMacro{fig-verl-lev-base-strict-rustfmt-minmax-base-olmo7b-lev-line}{0.34}
\DefMacro{fig-verl-lev-base-strict-rustfmt-minmax-prompt-olmo7b-lev-line}{0.35}
\DefMacro{fig-verl-lev-base-strict-rustfmt-minmax-olmo32b-lev-char-small}{}
\DefMacro{fig-verl-lev-base-strict-rustfmt-minmax-base-olmo32b-lev-char}{0.34}
\DefMacro{fig-verl-lev-base-strict-rustfmt-minmax-prompt-olmo32b-lev-char}{0.29}
\DefMacro{fig-verl-lev-base-strict-rustfmt-minmax-olmo32b-lev-line-small}{}
\DefMacro{fig-verl-lev-base-strict-rustfmt-minmax-base-olmo32b-lev-line}{0.27}
\DefMacro{fig-verl-lev-base-strict-rustfmt-minmax-prompt-olmo32b-lev-line}{0.22}
\DefMacro{fig-verl-lev-base-strict-rustfmt-minmax-haiku-lev-char-small}{}
\DefMacro{fig-verl-lev-base-strict-rustfmt-minmax-base-haiku-lev-char}{0.33}
\DefMacro{fig-verl-lev-base-strict-rustfmt-minmax-prompt-haiku-lev-char}{0.37}
}{}
\IfFileExists{numbers/verl/numbers_verl_comparison_all_mean_lev_rustfmt_minmax_macros.tex}{\DefMacro{fig-verl-comparison-allmean-rustfmt-minmax-qwen-lev-char-n}{545}
\DefMacro{fig-verl-comparison-allmean-rustfmt-minmax-base-qwen-lev-char}{0.54}
\DefMacro{fig-verl-comparison-allmean-rustfmt-minmax-prompt-qwen-lev-char}{0.53}
\DefMacro{fig-verl-comparison-allmean-rustfmt-minmax-rl-qwen-lev-char}{0.20}
\DefMacro{fig-verl-comparison-allmean-rustfmt-minmax-base-qwen-lev-line}{0.56}
\DefMacro{fig-verl-comparison-allmean-rustfmt-minmax-prompt-qwen-lev-line}{0.55}
\DefMacro{fig-verl-comparison-allmean-rustfmt-minmax-rl-qwen-lev-line}{0.21}
\DefMacro{fig-verl-comparison-allmean-rustfmt-minmax-gpt-lev-char-n}{415}
\DefMacro{fig-verl-comparison-allmean-rustfmt-minmax-base-gpt-lev-char}{0.54}
\DefMacro{fig-verl-comparison-allmean-rustfmt-minmax-prompt-gpt-lev-char}{0.51}
\DefMacro{fig-verl-comparison-allmean-rustfmt-minmax-rl-gpt-lev-char}{0.22}
\DefMacro{fig-verl-comparison-allmean-rustfmt-minmax-base-gpt-lev-line}{0.56}
\DefMacro{fig-verl-comparison-allmean-rustfmt-minmax-prompt-gpt-lev-line}{0.54}
\DefMacro{fig-verl-comparison-allmean-rustfmt-minmax-rl-gpt-lev-line}{0.24}
\DefMacro{fig-verl-comparison-allmean-rustfmt-minmax-olmo7b-lev-char-n}{69}
\DefMacro{fig-verl-comparison-allmean-rustfmt-minmax-base-olmo7b-lev-char}{0.47}
\DefMacro{fig-verl-comparison-allmean-rustfmt-minmax-prompt-olmo7b-lev-char}{0.51}
\DefMacro{fig-verl-comparison-allmean-rustfmt-minmax-rl-olmo7b-lev-char}{0.21}
\DefMacro{fig-verl-comparison-allmean-rustfmt-minmax-base-olmo7b-lev-line}{0.47}
\DefMacro{fig-verl-comparison-allmean-rustfmt-minmax-prompt-olmo7b-lev-line}{0.51}
\DefMacro{fig-verl-comparison-allmean-rustfmt-minmax-rl-olmo7b-lev-line}{0.19}
\DefMacro{fig-verl-comparison-allmean-rustfmt-minmax-olmo32b-lev-char-n}{203}
\DefMacro{fig-verl-comparison-allmean-rustfmt-minmax-base-olmo32b-lev-char}{0.51}
\DefMacro{fig-verl-comparison-allmean-rustfmt-minmax-prompt-olmo32b-lev-char}{0.49}
\DefMacro{fig-verl-comparison-allmean-rustfmt-minmax-rl-olmo32b-lev-char}{0.24}
\DefMacro{fig-verl-comparison-allmean-rustfmt-minmax-base-olmo32b-lev-line}{0.52}
\DefMacro{fig-verl-comparison-allmean-rustfmt-minmax-prompt-olmo32b-lev-line}{0.53}
\DefMacro{fig-verl-comparison-allmean-rustfmt-minmax-rl-olmo32b-lev-line}{0.25}
\DefMacro{fig-verl-comparison-allmean-rustfmt-minmax-haiku-lev-char-n}{441}
\DefMacro{fig-verl-comparison-allmean-rustfmt-minmax-base-haiku-lev-char}{0.45}
\DefMacro{fig-verl-comparison-allmean-rustfmt-minmax-prompt-haiku-lev-char}{0.52}
\DefMacro{fig-verl-comparison-allmean-rustfmt-minmax-rl-haiku-lev-char}{0.15}
}{}
\IfFileExists{numbers/verl/numbers_verl_lev_rl_pass_rustfmt_minmax_macros.tex}{\input{numbers/verl/numbers_verl_lev_rl_pass_rustfmt_minmax_macros.tex}}{}
\IfFileExists{numbers/verl/numbers_verl_lev_both_pass_rustfmt_minmax_macros.tex}{\DefMacro{fig-verl-lev-both-pass-rustfmt-minmax-qwen-lev-char-small}{}
\DefMacro{fig-verl-lev-both-pass-rustfmt-minmax-base-qwen-lev-char}{0.31}
\DefMacro{fig-verl-lev-both-pass-rustfmt-minmax-rl-qwen-lev-char}{0.22}
\DefMacro{fig-verl-lev-both-pass-rustfmt-minmax-qwen-lev-line-small}{}
\DefMacro{fig-verl-lev-both-pass-rustfmt-minmax-base-qwen-lev-line}{0.32}
\DefMacro{fig-verl-lev-both-pass-rustfmt-minmax-rl-qwen-lev-line}{0.14}
\DefMacro{fig-verl-lev-both-pass-rustfmt-minmax-gpt-lev-char-small}{}
\DefMacro{fig-verl-lev-both-pass-rustfmt-minmax-base-gpt-lev-char}{0.35}
\DefMacro{fig-verl-lev-both-pass-rustfmt-minmax-rl-gpt-lev-char}{0.22}
\DefMacro{fig-verl-lev-both-pass-rustfmt-minmax-gpt-lev-line-small}{}
\DefMacro{fig-verl-lev-both-pass-rustfmt-minmax-base-gpt-lev-line}{0.31}
\DefMacro{fig-verl-lev-both-pass-rustfmt-minmax-rl-gpt-lev-line}{0.24}
\DefMacro{fig-verl-lev-both-pass-rustfmt-minmax-olmo7b-lev-char-small}{\dag}
\DefMacro{fig-verl-lev-both-pass-rustfmt-minmax-base-olmo7b-lev-char}{0.22}
\DefMacro{fig-verl-lev-both-pass-rustfmt-minmax-rl-olmo7b-lev-char}{0.02}
\DefMacro{fig-verl-lev-both-pass-rustfmt-minmax-olmo7b-lev-line-small}{\dag}
\DefMacro{fig-verl-lev-both-pass-rustfmt-minmax-base-olmo7b-lev-line}{0.23}
\DefMacro{fig-verl-lev-both-pass-rustfmt-minmax-rl-olmo7b-lev-line}{0.02}
\DefMacro{fig-verl-lev-both-pass-rustfmt-minmax-olmo32b-lev-char-small}{}
\DefMacro{fig-verl-lev-both-pass-rustfmt-minmax-base-olmo32b-lev-char}{0.24}
\DefMacro{fig-verl-lev-both-pass-rustfmt-minmax-rl-olmo32b-lev-char}{0.14}
\DefMacro{fig-verl-lev-both-pass-rustfmt-minmax-olmo32b-lev-line-small}{}
\DefMacro{fig-verl-lev-both-pass-rustfmt-minmax-base-olmo32b-lev-line}{0.18}
\DefMacro{fig-verl-lev-both-pass-rustfmt-minmax-rl-olmo32b-lev-line}{0.13}
\DefMacro{fig-verl-lev-both-pass-rustfmt-minmax-haiku-lev-char-small}{}
\DefMacro{fig-verl-lev-both-pass-rustfmt-minmax-base-haiku-lev-char}{0.34}
\DefMacro{fig-verl-lev-both-pass-rustfmt-minmax-rl-haiku-lev-char}{0.25}
}{}
\IfFileExists{numbers/pr_overlap_counts/numbers_pr_lev_minmax_norm_pivot_rewrite_rustfmt_overlap_macros.tex}{\input{numbers/pr_overlap_counts/numbers_pr_lev_minmax_norm_pivot_rewrite_rustfmt_overlap_macros.tex}}{}
\IfFileExists{numbers/pr_overlap_counts/numbers_pr_selfedit_decrease_macros.tex}{\DefMacro{intro-selfedit-decrease-open-max}{14}
}{}
\IfFileExists{numbers/pr_overlap_counts/numbers_pr_lev_selfcross_mwu_macros.tex}{\DefMacro{fig-pr-lev-selfcross-mwu-open-n-tests}{16}
\DefMacro{fig-pr-lev-selfcross-mwu-open-n-significant}{7}
\DefMacro{fig-pr-lev-selfcross-mwu-open-n-right-direction}{14}
\DefMacro{fig-pr-lev-selfcross-mwu-qwen3.5-thinking:35b-a3b-q4-k-m-gpt-oss:20b-q4-k-m-p}{$.126$}
\DefMacro{fig-pr-lev-selfcross-mwu-qwen3.5-thinking:35b-a3b-q4-k-m-gpt-oss:20b-q4-k-m-stars}{}
\DefMacro{fig-pr-lev-selfcross-mwu-qwen3.5-thinking:35b-a3b-q4-k-m-olmo3-thinking:7b.think-q4-k-m-p}{$.209$}
\DefMacro{fig-pr-lev-selfcross-mwu-qwen3.5-thinking:35b-a3b-q4-k-m-olmo3-thinking:7b.think-q4-k-m-stars}{}
\DefMacro{fig-pr-lev-selfcross-mwu-qwen3.5-thinking:35b-a3b-q4-k-m-olmo3.1-thinking:32b.think-q4-k-m-p}{$.003$}
\DefMacro{fig-pr-lev-selfcross-mwu-qwen3.5-thinking:35b-a3b-q4-k-m-olmo3.1-thinking:32b.think-q4-k-m-stars}{$^{**}$}
\DefMacro{fig-pr-lev-selfcross-mwu-qwen3.5-thinking:35b-a3b-q4-k-m-claude-haiku-4-5-p}{$.858$}
\DefMacro{fig-pr-lev-selfcross-mwu-qwen3.5-thinking:35b-a3b-q4-k-m-claude-haiku-4-5-stars}{}
\DefMacro{fig-pr-lev-selfcross-mwu-gpt-oss:20b-q4-k-m-qwen3.5-thinking:35b-a3b-q4-k-m-p}{$.318$}
\DefMacro{fig-pr-lev-selfcross-mwu-gpt-oss:20b-q4-k-m-qwen3.5-thinking:35b-a3b-q4-k-m-stars}{}
\DefMacro{fig-pr-lev-selfcross-mwu-gpt-oss:20b-q4-k-m-olmo3-thinking:7b.think-q4-k-m-p}{$.123$}
\DefMacro{fig-pr-lev-selfcross-mwu-gpt-oss:20b-q4-k-m-olmo3-thinking:7b.think-q4-k-m-stars}{}
\DefMacro{fig-pr-lev-selfcross-mwu-gpt-oss:20b-q4-k-m-olmo3.1-thinking:32b.think-q4-k-m-p}{$.035$}
\DefMacro{fig-pr-lev-selfcross-mwu-gpt-oss:20b-q4-k-m-olmo3.1-thinking:32b.think-q4-k-m-stars}{$^{*}$}
\DefMacro{fig-pr-lev-selfcross-mwu-gpt-oss:20b-q4-k-m-claude-haiku-4-5-p}{$.964$}
\DefMacro{fig-pr-lev-selfcross-mwu-gpt-oss:20b-q4-k-m-claude-haiku-4-5-stars}{}
\DefMacro{fig-pr-lev-selfcross-mwu-olmo3-thinking:7b.think-q4-k-m-qwen3.5-thinking:35b-a3b-q4-k-m-p}{$.017$}
\DefMacro{fig-pr-lev-selfcross-mwu-olmo3-thinking:7b.think-q4-k-m-qwen3.5-thinking:35b-a3b-q4-k-m-stars}{$^{*}$}
\DefMacro{fig-pr-lev-selfcross-mwu-olmo3-thinking:7b.think-q4-k-m-gpt-oss:20b-q4-k-m-p}{$.068$}
\DefMacro{fig-pr-lev-selfcross-mwu-olmo3-thinking:7b.think-q4-k-m-gpt-oss:20b-q4-k-m-stars}{}
\DefMacro{fig-pr-lev-selfcross-mwu-olmo3-thinking:7b.think-q4-k-m-olmo3.1-thinking:32b.think-q4-k-m-p}{$.199$}
\DefMacro{fig-pr-lev-selfcross-mwu-olmo3-thinking:7b.think-q4-k-m-olmo3.1-thinking:32b.think-q4-k-m-stars}{}
\DefMacro{fig-pr-lev-selfcross-mwu-olmo3-thinking:7b.think-q4-k-m-claude-haiku-4-5-p}{$.827$}
\DefMacro{fig-pr-lev-selfcross-mwu-olmo3-thinking:7b.think-q4-k-m-claude-haiku-4-5-stars}{}
\DefMacro{fig-pr-lev-selfcross-mwu-olmo3.1-thinking:32b.think-q4-k-m-qwen3.5-thinking:35b-a3b-q4-k-m-p}{$.102$}
\DefMacro{fig-pr-lev-selfcross-mwu-olmo3.1-thinking:32b.think-q4-k-m-qwen3.5-thinking:35b-a3b-q4-k-m-stars}{}
\DefMacro{fig-pr-lev-selfcross-mwu-olmo3.1-thinking:32b.think-q4-k-m-gpt-oss:20b-q4-k-m-p}{$.026$}
\DefMacro{fig-pr-lev-selfcross-mwu-olmo3.1-thinking:32b.think-q4-k-m-gpt-oss:20b-q4-k-m-stars}{$^{*}$}
\DefMacro{fig-pr-lev-selfcross-mwu-olmo3.1-thinking:32b.think-q4-k-m-olmo3-thinking:7b.think-q4-k-m-p}{$.547$}
\DefMacro{fig-pr-lev-selfcross-mwu-olmo3.1-thinking:32b.think-q4-k-m-olmo3-thinking:7b.think-q4-k-m-stars}{}
\DefMacro{fig-pr-lev-selfcross-mwu-olmo3.1-thinking:32b.think-q4-k-m-claude-haiku-4-5-p}{$.279$}
\DefMacro{fig-pr-lev-selfcross-mwu-olmo3.1-thinking:32b.think-q4-k-m-claude-haiku-4-5-stars}{}
\DefMacro{fig-pr-lev-selfcross-mwu-claude-haiku-4-5-qwen3.5-thinking:35b-a3b-q4-k-m-p}{$.004$}
\DefMacro{fig-pr-lev-selfcross-mwu-claude-haiku-4-5-qwen3.5-thinking:35b-a3b-q4-k-m-stars}{$^{**}$}
\DefMacro{fig-pr-lev-selfcross-mwu-claude-haiku-4-5-gpt-oss:20b-q4-k-m-p}{$<.001$}
\DefMacro{fig-pr-lev-selfcross-mwu-claude-haiku-4-5-gpt-oss:20b-q4-k-m-stars}{$^{***}$}
\DefMacro{fig-pr-lev-selfcross-mwu-claude-haiku-4-5-olmo3-thinking:7b.think-q4-k-m-p}{$.818$}
\DefMacro{fig-pr-lev-selfcross-mwu-claude-haiku-4-5-olmo3-thinking:7b.think-q4-k-m-stars}{}
\DefMacro{fig-pr-lev-selfcross-mwu-claude-haiku-4-5-olmo3.1-thinking:32b.think-q4-k-m-p}{$.003$}
\DefMacro{fig-pr-lev-selfcross-mwu-claude-haiku-4-5-olmo3.1-thinking:32b.think-q4-k-m-stars}{$^{**}$}
}{}
\IfFileExists{numbers/pr_overlap_counts/numbers_pr_lev_minmax_norm_pivot_all_macros.tex}{\DefMacro{fig-pr-lev-minmax-norm-pivot-all-qwen3.5-thinking:35b-a3b-q4-k-m-qwen3.5-thinking:35b-a3b-q4-k-m-lev-char}{0.24}
\DefMacro{fig-pr-lev-minmax-norm-pivot-all-qwen3.5-thinking:35b-a3b-q4-k-m-qwen3.5-thinking:35b-a3b-q4-k-m-lev-char-n}{549}
\DefMacro{fig-pr-lev-minmax-norm-pivot-all-qwen3.5-thinking:35b-a3b-q4-k-m-gpt-oss:20b-q4-k-m-lev-char}{0.32}
\DefMacro{fig-pr-lev-minmax-norm-pivot-all-qwen3.5-thinking:35b-a3b-q4-k-m-olmo3-thinking:7b.think-q4-k-m-lev-char}{0.64}
\DefMacro{fig-pr-lev-minmax-norm-pivot-all-qwen3.5-thinking:35b-a3b-q4-k-m-olmo3.1-thinking:32b.think-q4-k-m-lev-char}{0.32}
\DefMacro{fig-pr-lev-minmax-norm-pivot-all-qwen3.5-thinking:35b-a3b-q4-k-m-claude-haiku-4-5-lev-char}{0.28}
\DefMacro{fig-pr-lev-minmax-norm-pivot-all-gpt-oss:20b-q4-k-m-qwen3.5-thinking:35b-a3b-q4-k-m-lev-char}{0.27}
\DefMacro{fig-pr-lev-minmax-norm-pivot-all-gpt-oss:20b-q4-k-m-gpt-oss:20b-q4-k-m-lev-char}{0.30}
\DefMacro{fig-pr-lev-minmax-norm-pivot-all-gpt-oss:20b-q4-k-m-gpt-oss:20b-q4-k-m-lev-char-n}{424}
\DefMacro{fig-pr-lev-minmax-norm-pivot-all-gpt-oss:20b-q4-k-m-olmo3-thinking:7b.think-q4-k-m-lev-char}{0.66}
\DefMacro{fig-pr-lev-minmax-norm-pivot-all-gpt-oss:20b-q4-k-m-olmo3.1-thinking:32b.think-q4-k-m-lev-char}{0.29}
\DefMacro{fig-pr-lev-minmax-norm-pivot-all-gpt-oss:20b-q4-k-m-claude-haiku-4-5-lev-char}{0.27}
\DefMacro{fig-pr-lev-minmax-norm-pivot-all-olmo3-thinking:7b.think-q4-k-m-qwen3.5-thinking:35b-a3b-q4-k-m-lev-char}{0.38}
\DefMacro{fig-pr-lev-minmax-norm-pivot-all-olmo3-thinking:7b.think-q4-k-m-gpt-oss:20b-q4-k-m-lev-char}{0.39}
\DefMacro{fig-pr-lev-minmax-norm-pivot-all-olmo3-thinking:7b.think-q4-k-m-olmo3-thinking:7b.think-q4-k-m-lev-char}{0.60}
\DefMacro{fig-pr-lev-minmax-norm-pivot-all-olmo3-thinking:7b.think-q4-k-m-olmo3-thinking:7b.think-q4-k-m-lev-char-n}{68}
\DefMacro{fig-pr-lev-minmax-norm-pivot-all-olmo3-thinking:7b.think-q4-k-m-olmo3.1-thinking:32b.think-q4-k-m-lev-char}{0.27}
\DefMacro{fig-pr-lev-minmax-norm-pivot-all-olmo3-thinking:7b.think-q4-k-m-claude-haiku-4-5-lev-char}{0.29}
\DefMacro{fig-pr-lev-minmax-norm-pivot-all-olmo3.1-thinking:32b.think-q4-k-m-qwen3.5-thinking:35b-a3b-q4-k-m-lev-char}{0.30}
\DefMacro{fig-pr-lev-minmax-norm-pivot-all-olmo3.1-thinking:32b.think-q4-k-m-gpt-oss:20b-q4-k-m-lev-char}{0.38}
\DefMacro{fig-pr-lev-minmax-norm-pivot-all-olmo3.1-thinking:32b.think-q4-k-m-olmo3-thinking:7b.think-q4-k-m-lev-char}{0.59}
\DefMacro{fig-pr-lev-minmax-norm-pivot-all-olmo3.1-thinking:32b.think-q4-k-m-olmo3.1-thinking:32b.think-q4-k-m-lev-char}{0.25}
\DefMacro{fig-pr-lev-minmax-norm-pivot-all-olmo3.1-thinking:32b.think-q4-k-m-olmo3.1-thinking:32b.think-q4-k-m-lev-char-n}{202}
\DefMacro{fig-pr-lev-minmax-norm-pivot-all-olmo3.1-thinking:32b.think-q4-k-m-claude-haiku-4-5-lev-char}{0.33}
\DefMacro{fig-pr-lev-minmax-norm-pivot-all-claude-haiku-4-5-qwen3.5-thinking:35b-a3b-q4-k-m-lev-char}{0.32}
\DefMacro{fig-pr-lev-minmax-norm-pivot-all-claude-haiku-4-5-gpt-oss:20b-q4-k-m-lev-char}{0.39}
\DefMacro{fig-pr-lev-minmax-norm-pivot-all-claude-haiku-4-5-olmo3-thinking:7b.think-q4-k-m-lev-char}{0.54}
\DefMacro{fig-pr-lev-minmax-norm-pivot-all-claude-haiku-4-5-olmo3.1-thinking:32b.think-q4-k-m-lev-char}{0.32}
\DefMacro{fig-pr-lev-minmax-norm-pivot-all-claude-haiku-4-5-claude-haiku-4-5-lev-char}{0.32}
\DefMacro{fig-pr-lev-minmax-norm-pivot-all-claude-haiku-4-5-claude-haiku-4-5-lev-char-n}{453}
}{}
\IfFileExists{numbers/verl/numbers_verl_lev_common_pass_mwu_macros.tex}{\input{numbers/verl/numbers_verl_lev_common_pass_mwu_macros.tex}}{}
\IfFileExists{numbers/verl/numbers_verl_comparison_all_mean_macros.tex}{\DefMacro{fig-verl-comparison-allmean-base-qwen-n}{603}
\DefMacro{fig-verl-comparison-allmean-base-qwen-build}{32.67}
\DefMacro{fig-verl-comparison-allmean-base-qwen-tests}{16.92}
\DefMacro{fig-verl-comparison-allmean-base-qwen-mean-tests}{21.12}
\DefMacro{fig-verl-comparison-allmean-prompt-qwen-build}{29.52}
\DefMacro{fig-verl-comparison-allmean-prompt-qwen-tests}{12.27}
\DefMacro{fig-verl-comparison-allmean-prompt-qwen-mean-tests}{17.95}
\DefMacro{fig-verl-comparison-allmean-verl-step62-qwen-build}{41.46}
\DefMacro{fig-verl-comparison-allmean-verl-step62-qwen-tests}{19.07}
\DefMacro{fig-verl-comparison-allmean-verl-step62-qwen-mean-tests}{26.97}
\DefMacro{fig-verl-comparison-allmean-base-gpt-n}{456}
\DefMacro{fig-verl-comparison-allmean-base-gpt-build}{28.51}
\DefMacro{fig-verl-comparison-allmean-base-gpt-tests}{16.23}
\DefMacro{fig-verl-comparison-allmean-base-gpt-mean-tests}{20.76}
\DefMacro{fig-verl-comparison-allmean-prompt-gpt-build}{30.70}
\DefMacro{fig-verl-comparison-allmean-prompt-gpt-tests}{16.45}
\DefMacro{fig-verl-comparison-allmean-prompt-gpt-mean-tests}{21.45}
\DefMacro{fig-verl-comparison-allmean-verl-step62-gpt-build}{37.28}
\DefMacro{fig-verl-comparison-allmean-verl-step62-gpt-tests}{18.42}
\DefMacro{fig-verl-comparison-allmean-verl-step62-gpt-mean-tests}{25.76}
\DefMacro{fig-verl-comparison-allmean-base-olmo7b-n}{79}
\DefMacro{fig-verl-comparison-allmean-base-olmo7b-build}{39.24}
\DefMacro{fig-verl-comparison-allmean-base-olmo7b-tests}{24.05}
\DefMacro{fig-verl-comparison-allmean-base-olmo7b-mean-tests}{28.06}
\DefMacro{fig-verl-comparison-allmean-prompt-olmo7b-build}{37.97}
\DefMacro{fig-verl-comparison-allmean-prompt-olmo7b-tests}{12.66}
\DefMacro{fig-verl-comparison-allmean-prompt-olmo7b-mean-tests}{22.66}
\DefMacro{fig-verl-comparison-allmean-verl-step62-olmo7b-build}{37.97}
\DefMacro{fig-verl-comparison-allmean-verl-step62-olmo7b-tests}{20.25}
\DefMacro{fig-verl-comparison-allmean-verl-step62-olmo7b-mean-tests}{26.63}
\DefMacro{fig-verl-comparison-allmean-base-olmo32b-n}{225}
\DefMacro{fig-verl-comparison-allmean-base-olmo32b-build}{31.11}
\DefMacro{fig-verl-comparison-allmean-base-olmo32b-tests}{20.00}
\DefMacro{fig-verl-comparison-allmean-base-olmo32b-mean-tests}{24.84}
\DefMacro{fig-verl-comparison-allmean-prompt-olmo32b-build}{36.00}
\DefMacro{fig-verl-comparison-allmean-prompt-olmo32b-tests}{21.33}
\DefMacro{fig-verl-comparison-allmean-prompt-olmo32b-mean-tests}{28.01}
\DefMacro{fig-verl-comparison-allmean-verl-step62-olmo32b-build}{37.78}
\DefMacro{fig-verl-comparison-allmean-verl-step62-olmo32b-tests}{25.33}
\DefMacro{fig-verl-comparison-allmean-verl-step62-olmo32b-mean-tests}{31.90}
\DefMacro{fig-verl-comparison-allmean-base-haiku-build}{32.63}
\DefMacro{fig-verl-comparison-allmean-base-haiku-tests}{12.36}
\DefMacro{fig-verl-comparison-allmean-base-haiku-mean-tests}{18.56}
\DefMacro{fig-verl-comparison-allmean-prompt-haiku-build}{35.14}
\DefMacro{fig-verl-comparison-allmean-prompt-haiku-tests}{19.31}
\DefMacro{fig-verl-comparison-allmean-prompt-haiku-mean-tests}{23.65}
\DefMacro{fig-verl-comparison-allmean-verl-step62-haiku-build}{45.75}
\DefMacro{fig-verl-comparison-allmean-verl-step62-haiku-tests}{20.85}
\DefMacro{fig-verl-comparison-allmean-verl-step62-haiku-mean-tests}{30.81}
}{}
\IfFileExists{numbers/verl/numbers_verl_comparison_all_mean_minmax_macros.tex}{\input{numbers/verl/numbers_verl_comparison_all_mean_minmax_macros.tex}}{}
\IfFileExists{numbers/verl/numbers_verl_comparison_all_mean_mwu_macros.tex}{\input{numbers/verl/numbers_verl_comparison_all_mean_mwu_macros.tex}}{}
\IfFileExists{numbers/verl/numbers_verl_test_pass_by_rewrite_split_macros.tex}{\input{numbers/verl/numbers_verl_test_pass_by_rewrite_split_macros.tex}}{}
\IfFileExists{numbers/verl/numbers_verl_meantest_nochange_macros.tex}{\DefMacro{fig-verl-meantestnc-base-qwen-mean-tests}{21.12}
\DefMacro{fig-verl-meantestnc-base-qwen-no-change}{11.97}
\DefMacro{fig-verl-meantestnc-base-qwen-mean-tests-change}{22.77}
\DefMacro{fig-verl-meantestnc-prompt-qwen-mean-tests}{17.95}
\DefMacro{fig-verl-meantestnc-prompt-qwen-no-change}{10.18}
\DefMacro{fig-verl-meantestnc-prompt-qwen-mean-tests-change}{19.35}
\DefMacro{fig-verl-meantestnc-verl-step62-qwen-mean-tests}{26.97}
\DefMacro{fig-verl-meantestnc-verl-step62-qwen-no-change}{14.45}
\DefMacro{fig-verl-meantestnc-verl-step62-qwen-mean-tests-change}{29.23}
\DefMacro{fig-verl-meantestnc-base-gpt-mean-tests}{20.76}
\DefMacro{fig-verl-meantestnc-base-gpt-no-change}{12.06}
\DefMacro{fig-verl-meantestnc-base-gpt-mean-tests-change}{22.61}
\DefMacro{fig-verl-meantestnc-prompt-gpt-mean-tests}{21.45}
\DefMacro{fig-verl-meantestnc-prompt-gpt-no-change}{12.69}
\DefMacro{fig-verl-meantestnc-prompt-gpt-mean-tests-change}{23.31}
\DefMacro{fig-verl-meantestnc-verl-step62-gpt-mean-tests}{25.76}
\DefMacro{fig-verl-meantestnc-verl-step62-gpt-no-change}{12.64}
\DefMacro{fig-verl-meantestnc-verl-step62-gpt-mean-tests-change}{28.55}
\DefMacro{fig-verl-meantestnc-base-olmo7b-mean-tests}{28.06}
\DefMacro{fig-verl-meantestnc-base-olmo7b-no-change}{16.67}
\DefMacro{fig-verl-meantestnc-base-olmo7b-mean-tests-change}{31.43}
\DefMacro{fig-verl-meantestnc-prompt-olmo7b-mean-tests}{22.66}
\DefMacro{fig-verl-meantestnc-prompt-olmo7b-no-change}{10.26}
\DefMacro{fig-verl-meantestnc-prompt-olmo7b-mean-tests-change}{26.32}
\DefMacro{fig-verl-meantestnc-verl-step62-olmo7b-mean-tests}{26.63}
\DefMacro{fig-verl-meantestnc-verl-step62-olmo7b-no-change}{3.70}
\DefMacro{fig-verl-meantestnc-verl-step62-olmo7b-mean-tests-change}{33.39}
\DefMacro{fig-verl-meantestnc-base-olmo32b-mean-tests}{24.84}
\DefMacro{fig-verl-meantestnc-base-olmo32b-no-change}{6.25}
\DefMacro{fig-verl-meantestnc-base-olmo32b-mean-tests-change}{27.92}
\DefMacro{fig-verl-meantestnc-prompt-olmo32b-mean-tests}{28.01}
\DefMacro{fig-verl-meantestnc-prompt-olmo32b-no-change}{12.45}
\DefMacro{fig-verl-meantestnc-prompt-olmo32b-mean-tests-change}{30.59}
\DefMacro{fig-verl-meantestnc-verl-step62-olmo32b-mean-tests}{31.90}
\DefMacro{fig-verl-meantestnc-verl-step62-olmo32b-no-change}{8.53}
\DefMacro{fig-verl-meantestnc-verl-step62-olmo32b-mean-tests-change}{35.77}
\DefMacro{fig-verl-meantestnc-base-haiku-mean-tests}{18.56}
\DefMacro{fig-verl-meantestnc-base-haiku-no-change}{14.54}
\DefMacro{fig-verl-meantestnc-base-haiku-mean-tests-change}{20.25}
\DefMacro{fig-verl-meantestnc-prompt-haiku-mean-tests}{23.65}
\DefMacro{fig-verl-meantestnc-prompt-haiku-no-change}{17.97}
\DefMacro{fig-verl-meantestnc-prompt-haiku-mean-tests-change}{26.04}
\DefMacro{fig-verl-meantestnc-verl-step62-haiku-mean-tests}{30.81}
\DefMacro{fig-verl-meantestnc-verl-step62-haiku-no-change}{18.62}
\DefMacro{fig-verl-meantestnc-verl-step62-haiku-mean-tests-change}{35.92}
}{}
\IfFileExists{numbers/verl/numbers_verl_build_nochange_macros.tex}{\DefMacro{fig-verl-buildnc-base-qwen-build}{32.67}
\DefMacro{fig-verl-buildnc-base-qwen-no-change}{23.91}
\DefMacro{fig-verl-buildnc-base-qwen-build-change}{34.25}
\DefMacro{fig-verl-buildnc-prompt-qwen-build}{29.52}
\DefMacro{fig-verl-buildnc-prompt-qwen-no-change}{27.17}
\DefMacro{fig-verl-buildnc-prompt-qwen-build-change}{29.94}
\DefMacro{fig-verl-buildnc-verl-step62-qwen-build}{41.46}
\DefMacro{fig-verl-buildnc-verl-step62-qwen-no-change}{33.70}
\DefMacro{fig-verl-buildnc-verl-step62-qwen-build-change}{42.86}
\DefMacro{fig-verl-buildnc-base-gpt-build}{28.51}
\DefMacro{fig-verl-buildnc-base-gpt-no-change}{20.00}
\DefMacro{fig-verl-buildnc-base-gpt-build-change}{30.32}
\DefMacro{fig-verl-buildnc-prompt-gpt-build}{30.70}
\DefMacro{fig-verl-buildnc-prompt-gpt-no-change}{22.50}
\DefMacro{fig-verl-buildnc-prompt-gpt-build-change}{32.45}
\DefMacro{fig-verl-buildnc-verl-step62-gpt-build}{37.28}
\DefMacro{fig-verl-buildnc-verl-step62-gpt-no-change}{27.50}
\DefMacro{fig-verl-buildnc-verl-step62-gpt-build-change}{39.36}
\DefMacro{fig-verl-buildnc-base-olmo7b-build}{39.24}
\DefMacro{fig-verl-buildnc-base-olmo7b-no-change}{27.78}
\DefMacro{fig-verl-buildnc-base-olmo7b-build-change}{42.62}
\DefMacro{fig-verl-buildnc-prompt-olmo7b-build}{37.97}
\DefMacro{fig-verl-buildnc-prompt-olmo7b-no-change}{22.22}
\DefMacro{fig-verl-buildnc-prompt-olmo7b-build-change}{42.62}
\DefMacro{fig-verl-buildnc-verl-step62-olmo7b-build}{37.97}
\DefMacro{fig-verl-buildnc-verl-step62-olmo7b-no-change}{22.22}
\DefMacro{fig-verl-buildnc-verl-step62-olmo7b-build-change}{42.62}
\DefMacro{fig-verl-buildnc-base-olmo32b-build}{31.11}
\DefMacro{fig-verl-buildnc-base-olmo32b-no-change}{9.38}
\DefMacro{fig-verl-buildnc-base-olmo32b-build-change}{34.72}
\DefMacro{fig-verl-buildnc-prompt-olmo32b-build}{36.00}
\DefMacro{fig-verl-buildnc-prompt-olmo32b-no-change}{18.75}
\DefMacro{fig-verl-buildnc-prompt-olmo32b-build-change}{38.86}
\DefMacro{fig-verl-buildnc-verl-step62-olmo32b-build}{37.78}
\DefMacro{fig-verl-buildnc-verl-step62-olmo32b-no-change}{18.75}
\DefMacro{fig-verl-buildnc-verl-step62-olmo32b-build-change}{40.93}
\DefMacro{fig-verl-buildnc-base-haiku-build}{32.63}
\DefMacro{fig-verl-buildnc-base-haiku-no-change}{31.37}
\DefMacro{fig-verl-buildnc-base-haiku-build-change}{33.15}
\DefMacro{fig-verl-buildnc-prompt-haiku-build}{35.14}
\DefMacro{fig-verl-buildnc-prompt-haiku-no-change}{30.07}
\DefMacro{fig-verl-buildnc-prompt-haiku-build-change}{37.26}
\DefMacro{fig-verl-buildnc-verl-step62-haiku-build}{45.75}
\DefMacro{fig-verl-buildnc-verl-step62-haiku-no-change}{37.25}
\DefMacro{fig-verl-buildnc-verl-step62-haiku-build-change}{49.32}
\DefMacro{fig-verl-buildnc-pooled-nochange-share}{19.94}
}{}
\IfFileExists{numbers/verl/numbers_verl_build_nochange_shares_macros.tex}{\input{numbers/verl/numbers_verl_build_nochange_shares_macros.tex}}{}
\IfFileExists{numbers/verl/numbers_verl_meantest_nochange_shares_macros.tex}{\input{numbers/verl/numbers_verl_meantest_nochange_shares_macros.tex}}{}
\IfFileExists{numbers/corpus_stats/numbers_corpus_filter_stats_eval_rust_repos_macros.tex}{\DefMacro{corpus-eval-rust-repos-split-functions}{8272}
\DefMacro{corpus-eval-rust-repos-readme-prs}{4514}
\DefMacro{corpus-eval-rust-repos-readme-functions}{8235}
\DefMacro{corpus-eval-rust-repos-readme-repos}{356}
\DefMacro{corpus-eval-rust-repos-pretests-prs}{1404}
\DefMacro{corpus-eval-rust-repos-pretests-functions}{2458}
\DefMacro{corpus-eval-rust-repos-pretests-repos}{207}
\DefMacro{corpus-eval-rust-repos-draft-target-functions}{2442}
\DefMacro{corpus-eval-rust-repos-target-callsites-removed}{12}
\DefMacro{corpus-eval-rust-repos-target-after-callsites}{2446}
\DefMacro{corpus-eval-rust-repos-target-funcdesc-removed}{4}
\DefMacro{corpus-eval-rust-repos-funcdesc-ctx}{10000}
\DefMacro{corpus-eval-rust-repos-callsite-window}{5}
\DefMacro{corpus-eval-rust-repos-repair-rounds}{6}
\DefMacro{corpus-eval-rust-repos-haiku-repair-rounds}{2}
}{}
\IfFileExists{numbers/corpus_stats/numbers_edit_filter_stats_eval_rust_repos_macros.tex}{\DefMacro{edit-filter-eval-rust-repos-qwen-after-presolved}{612}
\DefMacro{edit-filter-eval-rust-repos-qwen-after-nodesc}{603}
\DefMacro{edit-filter-eval-rust-repos-qwen-wide-passing}{1602}
\DefMacro{edit-filter-eval-rust-repos-qwen-wide-after-overlap}{1169}
\DefMacro{edit-filter-eval-rust-repos-gpt-after-presolved}{466}
\DefMacro{edit-filter-eval-rust-repos-gpt-after-nodesc}{456}
\DefMacro{edit-filter-eval-rust-repos-gpt-wide-passing}{1102}
\DefMacro{edit-filter-eval-rust-repos-gpt-wide-after-overlap}{789}
\DefMacro{edit-filter-eval-rust-repos-olmo7b-after-presolved}{80}
\DefMacro{edit-filter-eval-rust-repos-olmo7b-after-nodesc}{79}
\DefMacro{edit-filter-eval-rust-repos-olmo7b-wide-passing}{199}
\DefMacro{edit-filter-eval-rust-repos-olmo7b-wide-after-overlap}{153}
\DefMacro{edit-filter-eval-rust-repos-olmo32b-after-presolved}{226}
\DefMacro{edit-filter-eval-rust-repos-olmo32b-after-nodesc}{225}
\DefMacro{edit-filter-eval-rust-repos-olmo32b-wide-passing}{459}
\DefMacro{edit-filter-eval-rust-repos-olmo32b-wide-after-overlap}{349}
\DefMacro{edit-filter-eval-rust-repos-haiku-after-presolved}{534}
\DefMacro{edit-filter-eval-rust-repos-haiku-after-nodesc}{518}
\DefMacro{edit-filter-eval-rust-repos-haiku-wide-passing}{1139}
\DefMacro{edit-filter-eval-rust-repos-haiku-wide-after-overlap}{728}
}{}
\IfFileExists{numbers/corpus_stats/numbers_train_dataset_training_rust_repos_macros.tex}{\DefMacro{corpus-training-rust-repos-fn-level-prs}{4666}
\DefMacro{corpus-training-rust-repos-fn-level-functions}{9150}
\DefMacro{corpus-training-rust-repos-fn-level-repos}{388}
\DefMacro{corpus-training-rust-repos-tests-prs}{942}
\DefMacro{corpus-training-rust-repos-tests-functions}{1636}
\DefMacro{corpus-training-rust-repos-tests-repos}{150}
\DefMacro{corpus-training-rust-repos-repair-rounds}{10}
\DefMacro{verl-max-seq-length}{12288}
\DefMacro{verl-max-new-tokens}{6144}
\DefMacro{verl-max-prompt-tokens}{6144}
\DefMacro{verl-train-prompts}{355}
\DefMacro{verl-val-prompts}{39}
}{}
\IfFileExists{numbers/corpus_stats/numbers_gt_test_coverage_eval_rust_repos_macros.tex}{\DefMacro{gt-cov-eval-rust-repos-pretests-mean}{87.9}
\DefMacro{gt-cov-eval-rust-repos-pretests-median}{94.5}
}{}
\IfFileExists{numbers/pr_overlap_counts/numbers_pr_lev_minmax_norm_pivot_rewrite_macros.tex}{\input{numbers/pr_overlap_counts/numbers_pr_lev_minmax_norm_pivot_rewrite_macros.tex}}{}
\IfFileExists{numbers/pr_cargo_ablate/numbers_pr_edit_pass_rate_pivot_macros.tex}{\DefMacro{fig-pr-edit-pass-rate-pivot-qwen3.5-thinking:35b-a3b-q4-k-m-qwen3.5-thinking:35b-a3b-q4-k-m-tests}{66.00}
\DefMacro{fig-pr-edit-pass-rate-pivot-qwen3.5-thinking:35b-a3b-q4-k-m-gpt-oss:20b-q4-k-m-tests}{63.68}
\DefMacro{fig-pr-edit-pass-rate-pivot-qwen3.5-thinking:35b-a3b-q4-k-m-olmo3-thinking:7b.think-q4-k-m-tests}{16.92}
\DefMacro{fig-pr-edit-pass-rate-pivot-qwen3.5-thinking:35b-a3b-q4-k-m-olmo3.1-thinking:32b.think-q4-k-m-tests}{39.97}
\DefMacro{fig-pr-edit-pass-rate-pivot-qwen3.5-thinking:35b-a3b-q4-k-m-claude-haiku-4-5-tests}{71.64}
\DefMacro{fig-pr-edit-pass-rate-pivot-gpt-oss:20b-q4-k-m-qwen3.5-thinking:35b-a3b-q4-k-m-tests}{67.32}
\DefMacro{fig-pr-edit-pass-rate-pivot-gpt-oss:20b-q4-k-m-gpt-oss:20b-q4-k-m-tests}{64.47}
\DefMacro{fig-pr-edit-pass-rate-pivot-gpt-oss:20b-q4-k-m-olmo3-thinking:7b.think-q4-k-m-tests}{16.23}
\DefMacro{fig-pr-edit-pass-rate-pivot-gpt-oss:20b-q4-k-m-olmo3.1-thinking:32b.think-q4-k-m-tests}{40.57}
\DefMacro{fig-pr-edit-pass-rate-pivot-gpt-oss:20b-q4-k-m-claude-haiku-4-5-tests}{67.32}
\DefMacro{fig-pr-edit-pass-rate-pivot-olmo3-thinking:7b.think-q4-k-m-qwen3.5-thinking:35b-a3b-q4-k-m-tests}{60.76}
\DefMacro{fig-pr-edit-pass-rate-pivot-olmo3-thinking:7b.think-q4-k-m-gpt-oss:20b-q4-k-m-tests}{59.49}
\DefMacro{fig-pr-edit-pass-rate-pivot-olmo3-thinking:7b.think-q4-k-m-olmo3-thinking:7b.think-q4-k-m-tests}{24.05}
\DefMacro{fig-pr-edit-pass-rate-pivot-olmo3-thinking:7b.think-q4-k-m-olmo3.1-thinking:32b.think-q4-k-m-tests}{37.97}
\DefMacro{fig-pr-edit-pass-rate-pivot-olmo3-thinking:7b.think-q4-k-m-claude-haiku-4-5-tests}{67.09}
\DefMacro{fig-pr-edit-pass-rate-pivot-olmo3.1-thinking:32b.think-q4-k-m-qwen3.5-thinking:35b-a3b-q4-k-m-tests}{69.78}
\DefMacro{fig-pr-edit-pass-rate-pivot-olmo3.1-thinking:32b.think-q4-k-m-gpt-oss:20b-q4-k-m-tests}{67.56}
\DefMacro{fig-pr-edit-pass-rate-pivot-olmo3.1-thinking:32b.think-q4-k-m-olmo3-thinking:7b.think-q4-k-m-tests}{20.00}
\DefMacro{fig-pr-edit-pass-rate-pivot-olmo3.1-thinking:32b.think-q4-k-m-olmo3.1-thinking:32b.think-q4-k-m-tests}{42.22}
\DefMacro{fig-pr-edit-pass-rate-pivot-olmo3.1-thinking:32b.think-q4-k-m-claude-haiku-4-5-tests}{69.78}
\DefMacro{fig-pr-edit-pass-rate-pivot-claude-haiku-4-5-qwen3.5-thinking:35b-a3b-q4-k-m-tests}{49.61}
\DefMacro{fig-pr-edit-pass-rate-pivot-claude-haiku-4-5-gpt-oss:20b-q4-k-m-tests}{49.03}
\DefMacro{fig-pr-edit-pass-rate-pivot-claude-haiku-4-5-olmo3-thinking:7b.think-q4-k-m-tests}{12.36}
\DefMacro{fig-pr-edit-pass-rate-pivot-claude-haiku-4-5-olmo3.1-thinking:32b.think-q4-k-m-tests}{27.61}
\DefMacro{fig-pr-edit-pass-rate-pivot-claude-haiku-4-5-claude-haiku-4-5-tests}{57.14}
}{}

\begin{document}

\title{\twemoji[scale=1]{crocodile} \Tool: Cross-Model Code Editing with LLMs}

\author{
Linghan Zhong{\normalfont\textsuperscript{1}},\,
Aditya Thimmaiah{\normalfont\textsuperscript{1}},
\\
\textbf{Jayanth Srinivasa}{\normalfont\textsuperscript{2}},\,
\textbf{Milos Gligoric}{\normalfont\textsuperscript{1}},\,
\textbf{Junyi Jessy Li}{\normalfont\textsuperscript{1}}
\\[0.5em]
\texttt{linghanz@cs.utexas.edu},\,
\texttt{auditt@utexas.edu},\,
\\[-0.2em]
\texttt{jasriniv@cisco.com},\,
\texttt{gligoric@utexas.edu},\,
\texttt{jessy@utexas.edu}
\\[0.5em]
\textsuperscript{1}The University of Texas at Austin
\;
\textsuperscript{2}Cisco Research
}

\maketitle

\begin{abstract}
Large language models (\LLMs) have become ubiquitous tools for code
generation and editing. However, development teams often use
\heterogeneous \LLM assistants. Different developers may prefer
different models, and individual developers may switch between
models across different coding sessions. Because of this, the edits any one
model makes are frequently applied to \othercode originally
generated by another model. These \LLMs are often trained on different
datasets, and as a result have different stylistic preferences. Do \LLMs behave differently when they edit
\othercode originally written by a different \LLM with a different
coding style? We find that models tend
to make more, and often excessive, edits on \othercode. We introduce \Tool (\textbf{Cro}ss-model \textbf{Cod}e Ed\textbf{i}ting with \textbf{L}LMs), a
post-training framework for reducing excessive edits while
preserving functional correctness. \Tool's similarity reward
penalizes large changes, while its execution reward scores build and
test success. We use the product of these two rewards to encourage
the policy to decrease the edit size without decreasing the edit
\task success rate. \Tool is available at \url{https://github.com/EngineeringSoftware/Crocodil}.
\end{abstract}

\section{Introduction}

\begin{figure}[!t]
\centering
\includegraphics[width=\columnwidth]{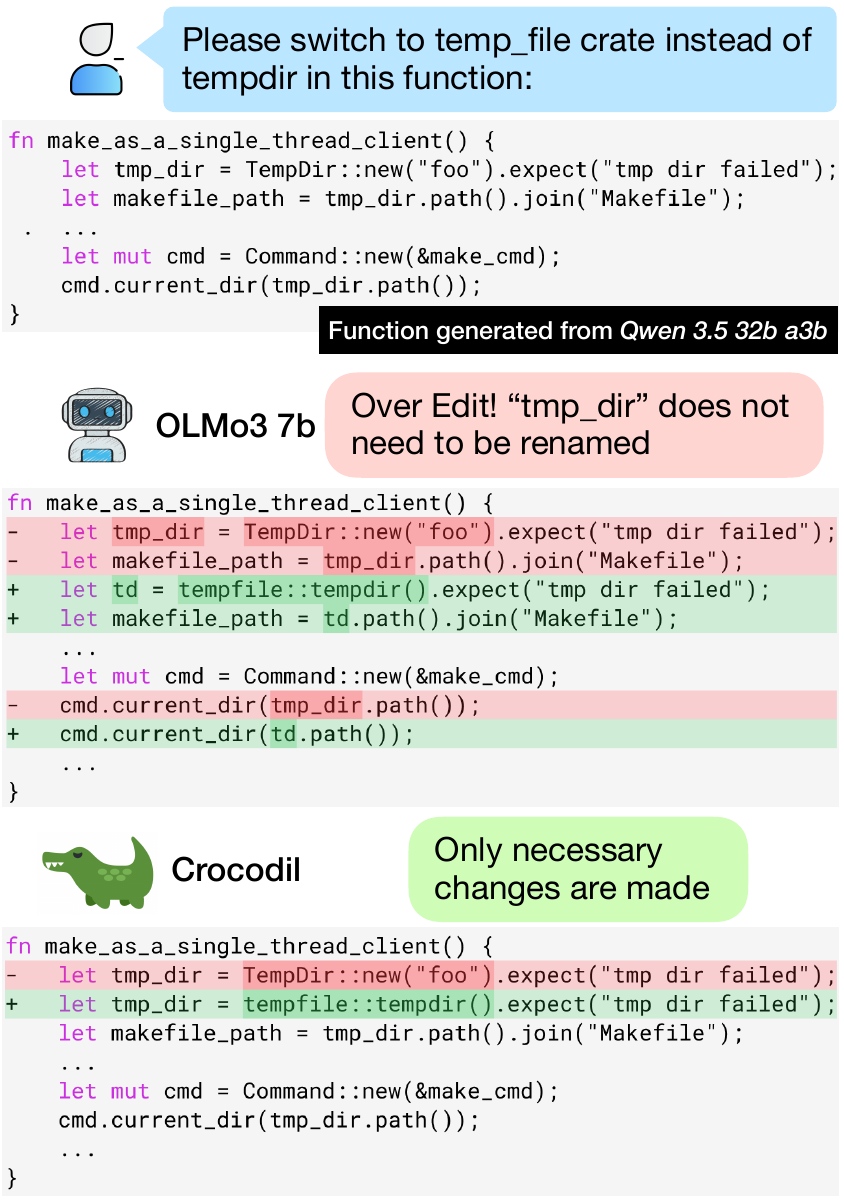}
\caption{\Crossedit example: vanilla \olmos over-edits a
\qwen-generated function by renaming the identifier \texttt{tmp\_dir} to
\texttt{td}. Trained with \Tool, it makes only the requested
change.}
\label{fig:intro}
\end{figure}

Large language models (\LLMs) have become ubiquitous tools for code
generation and editing. Developers use them to draft new
functions~\cite{chen2021humaneval,austin2021mbpp,nijkamp2022codegen}, refactor existing modules~\cite{cui2025refactorbench}, and patch failing
tests~\cite{xia2022alpharepair,xia2023apr,yang2024sweagent,tufano2019empirical}. With a wide range of models available, development teams
often use \heterogeneous \LLMs. Different developers may prefer different models~\cite{liang2024survey}, and the same developer may switch between models
from one session to the next as task demands, latency, or cost shift~\cite{cai2026oneisnotenough}. As a
result, the \beforecode each \LLM assistant is tasked to edit not
only includes \emph{\selfcode} authored by itself (\emph{\selfedit}) but also
\emph{\othercode} that another model originally generated (\emph{\crossedit}). These
models are trained on different corpora and tuned under different
post-training objectives~\cite{ouyang2022instructgpt,roziere2023codellama,hui2024qwen25coder}. They accordingly hold their own
preferences in style. Do these preferences lead an \LLM to make more edits
during \crossedit than during \selfedit, \reimplementing \othercode toward
its own conventions?

Prior work on benchmarking \LLM code editing ability~\cite{just2014defects4j,widyasari2020bugsinpy,Cassano2023CanItEdit} is
ill-suited for answering this question. Such benchmarks usually support varying only the \editor \LLM for each edit task, but not the author of the \before code. With this limitation, we cannot easily evaluate how much of an \editor's output is driven by
who wrote the \before code.

We first reveal that \textbf{(1) \LLMs are excessive code editors}, and \textbf{(2) mixing and matching \rewriter and \editor models when coding makes the \LLMs' edits more excessive}.
To systematically assess how the editing behavior of \LLMs differs on
\selfcode compared to \othercode,
we introduce a novel data-collection framework that gathers
\beforecode from merged \github pull requests (\PRs) on popular Rust
crates, and use a set of models to \reimplement the \before code themselves. We then run
existing tests to ensure the \reimplementations are correct.
Each of the models is then asked to apply the original \PR's edit
to every \reimplementation.

Using \numofeditors \LLMs on this corpus, we find that the
\numofopeneditors open-weight models edit their own \reimplementations up to
\UseMacro{intro-selfedit-decrease-open-max}\% less than they edit the
other models' \reimplementations, on
\UseMacro{fig-pr-lev-selfcross-mwu-open-n-right-direction} of
\UseMacro{fig-pr-lev-selfcross-mwu-open-n-tests} model pairings.
The gap cannot be closed by prompt engineering: a system prompt asking for the smallest
possible edit cannot consistently make edits smaller or improve
edit success.
This
motivates a training signal that explicitly
penalizes excessive editing without sacrificing correctness.
Fewer changes per \PR could mean faster code reviews, fewer tests to run
for that \PR, and thus faster overall development.

To address this gap, we introduce \Tool
(\textbf{Cro}ss-model \textbf{Cod}e
Ed\textbf{i}ting with \textbf{L}LMs),
a post-training framework for code editing models with a
novel reward function to reduce this excessive
editing while preserving functional correctness. \Tool's components
penalize large changes while rewarding build and test success.
We reuse our data-collection framework to gather the
training \tasks, so the model can be
trained against code by another \LLM whose style it is not familiar with.
We show that \Tool halves \olmos's edit size on \reimplementations by every model, while
improving build and test pass rates on every \rewriter but \olmos
itself. When restricted to \tasks where edits by both
the base model and the \Tool-trained model pass every test, \Tool still
produces smaller edits, demonstrating that our model indeed reduces
excessive non-task-related edits.

\section{Related Work}

\MyPara{LLM-based code editing and repair}
\citet{ZhangETAL22CoditT5} introduced CoditT5, which proposed a pretraining objective that
explicitly models code edits, and used it to train an \LLM on a large amount of
source code and natural language comments. Can It Edit~\cite{Cassano2023CanItEdit} builds a
fine-tuning dataset, EditPackFT, by collecting and cleaning commits on
Python repositories, and provides a benchmark with
human-written code, tasks, and test suites. EDIT-Bench~\cite{chi2026editbench}
builds a benchmark by collecting instructed edit tasks from the users of their VS Code extension,
pairing each real user instruction with the corresponding file, the highlighted
region, and the cursor position. A team of programmers then writes the test harnesses by hand to check the \LLM-generated edits on those tasks.
SWE-agent~\cite{yang2024sweagent} creates an
Agent-Computer Interface to provide \LLMs with tools that help them view,
search through, and edit files. Agentless~\cite{xiaAgentlessDemystifyingLLMbased2024} employs a three-phase process of localization, repair, and patch
validation, without letting the \LLM decide future actions or operate
with complex tools. Instead of focusing on improving \LLMs at solving editing tasks, \Tool trains
\LLMs to make fewer excessive edits, especially on \othercode.

\MyPara{Reinforcement learning for coding \LLMs}
CodeRL~\cite{le2022coderl} trains an actor that generates programs alongside a critic
that provides dense execution feedback to the actor during both
training and inference. RLTF~\cite{liu2023rltf} introduces a fine-grained reward from
unit-test outcomes, so the policy can attribute failure to specific
lines rather than only to the final program. StepCoder~\cite{dou2024stepcoder} splits
generation into a curriculum of shorter subtasks and masks
unexecuted code in the compiler-feedback reward. These works design their reward around the correctness of the output. \Tool pairs an \execreward
with a \simreward, so a model that edits \othercode is encouraged to keep generating functionally correct edits while being penalized for changing the \before code excessively.

\section{Characterizing \Crossediting}
\label{sec:char-crossediting}

\subsection{Corpus}
\label{sec:corpus}
We first construct a dataset of \github pull requests (\PRs) with both
the \before and \after versions of each function, and their associated
test cases.

\MyPara{Data selection}
We collect \PRs from Rust projects registered on crates.io
\cite{crates-io} and hosted on \github. Each \PR must satisfy: (1)~at most five files changed when merged,
and (2)~at least 75\% of the changed files are in Rust. We focus on Rust here because it is widely used, its projects are generally well tested, it compiler could provide rich feedback for \repairstage, improving success rate for \LLM \reimplementations. To bound the scope of the \reimplementation
and edit, we break each \PR into per-function updates. We include
global functions, associated functions, which are generally defined on a
type, and methods, which are associated functions that are called on a
particular instance of a type. For simplicity, we refer to all three as \emph{functions} in this work.

Each extracted function must exist before (\before) the \PR and must
not be removed or renamed after (\after) the \PR. We match \before and
\after through file path, name, and \CodeIn{impl} header. Both the \before and
\after functions are limited to between 20 and 200 non-empty, non-comment
lines. These limits keep the context size tractable while retaining
functions complex enough to carry meaningful edits. We believe this setup can
generalize to real-world multi-file edit settings, since each multi-file change is
made up of many function edits, but we leave that direction for future work.
On top of these, we keep only the \PRs whose repository has a
README, to make sure we can easily generate a description of the repository for the models.
In total, we collected
\num{\UseMacro{corpus-eval-rust-repos-readme-prs}} \PRs
and \num{\UseMacro{corpus-eval-rust-repos-readme-functions}} functions
across \num{\UseMacro{corpus-eval-rust-repos-readme-repos}} repositories
that satisfied these constraints.

For every collected function we prepare the additional context items
listed in Table~\ref{tab:context-items}. Every \rewriter
(Section~\ref{sec:reimplement}) and \editor
(Section~\ref{sec:edit}) uses a subset of this list as its input
context.

\begin{table}[t]
\centering
\small
\setlength{\tabcolsep}{2pt}
\begin{tabularx}{\columnwidth}{@{}llX@{}}
\toprule
\textbf{ID} & \textbf{Item} & \textbf{Description} \\
\midrule
\IDFuncsig & \Funcsig &
The \before or \after function's signature. \\
\midrule
\IDRestprdiff & \Restprdiff &
Unified \PR diff with the target function's own \hunk removed. \\
\midrule
\IDCallees & \Callees &
Other Rust functions in the same repo that the \before or \after function calls (signature and body). \\
\midrule
\IDCallsites & \Callsites &
Surrounding code at each place the function is called
($\pm$\UseMacro{corpus-eval-rust-repos-callsite-window} LOC). \\
\midrule
\IDUsestmts & \Usestmts &
The file's \CodeIn{use} declarations, showing the modules in scope. \\
\midrule
\IDStructimpl & \Structimpl &
Header of the \CodeIn{impl} block containing the function and the \CodeIn{struct} it implements (e.g., \CodeIn{impl\; Tra\;for\;Val}). \\
\midrule
\IDRepodesc & \Repodesc &
\LLM-generated description of the project derived from the repository's README. \\
\midrule
\IDPRdesc & \PRdesc &
\LLM-generated description of the \PR's overall purpose. \\
\midrule
\IDFuncdesc & \Funcdesc &
\LLM-generated description of the function. \\
\bottomrule
\end{tabularx}
\caption{Per-function context items prepared once and reused as a
pool from which \rewriter and \editor each draw a subset. Repo, PR, and Function description generated with \qwen.}
\label{tab:context-items}
\end{table}

\begin{figure}[t]
\centering
\includegraphics[width=\columnwidth]{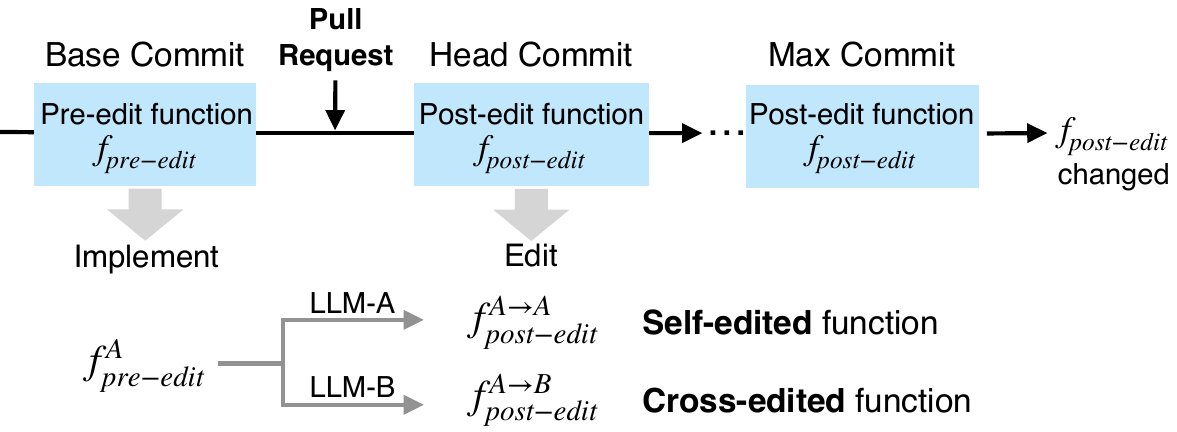}
\caption{Our data collection, \reimplementation, and edit framework.}
\label{fig:approach-framework}
\vspace{-12pt}
\end{figure}

\MyPara{Test collection}
To check the correctness of the \reimplementation and the success of
the edit, we collect tests that check each function from the three
commits illustrated in Figure~\ref{fig:approach-framework}. The
\textbf{\basecommit} is the parent of the \PR, where the
function is in its \before form, so its tests can check \reimplementations.
The \textbf{\headcommit} is the tip of the \PR, where the function is
in its \after form and includes any tests the \PR added, so its tests can
check the edits. The \textbf{\maxcommit} is the last commit on the
main branch where the function still matches the \after version exactly, and we
use its tests to augment the \headcommit tests.

We use a line-coverage collector for Rust to record which lines of code
are executed when each test runs. This allows us to keep only the tests
that check the behavior of the \before and \after functions.
Test collection
succeeds for
\num{\UseMacro{corpus-eval-rust-repos-pretests-functions}} of them
(\num{\UseMacro{corpus-eval-rust-repos-pretests-prs}} \PRs,
\num{\UseMacro{corpus-eval-rust-repos-pretests-repos}} repositories).
On these functions, the collected tests execute
\UseMacro{gt-cov-eval-rust-repos-pretests-mean}\%
of the lines of the
developer's \after function on average, with a median of
\UseMacro{gt-cov-eval-rust-repos-pretests-median}\%.

\MyPara{LLMs}
We use \numofeditors \LLMs in our study: \numofopeneditors open-weight models
including \qwen~\cite{yang2025qwen3}, \gpto~\cite{openai2025gptoss},
\olmosfull~\cite{olmoteam2025olmo3}, \olmolfull~\cite{olmoteam2025olmo3}, and one
closed-source model, \haiku~\cite{anthropic2025haiku}.
Each \LLM we study serves in two roles: as an
\textit{\rewriter} that \reimplements the collected \before code, and as an
\textit{\editor} that applies the changes to the \reimplemented code.

As shown in Figure~\ref{fig:approach-framework}, we derive an editing
\task for every (\reimplementation, \PR change) pair, and have every
\editor perform every \task, including the \tasks of editing its own outputs as \rewriter. This lets us compare model behavior on
\crossediting and \selfediting.

\subsection{\Reimplementation}
\label{sec:reimplement}

During \reimplementation, we use each \rewriter to convert each
developer-written function into a set of model-authored equivalents.
The \reimplementation process is split into a \draftstage where the
\rewriter generates an initial \draft of the function, and a
\repairstage where it repairs that \draft until it builds and passes
the collected tests, for up to
\UseMacro{corpus-eval-rust-repos-repair-rounds} rounds. For \haiku,
we run only \UseMacro{corpus-eval-rust-repos-haiku-repair-rounds}
rounds instead of \UseMacro{corpus-eval-rust-repos-repair-rounds} to
bound API cost.

\MyPara{Draft stage}
We invoke each \rewriter once per function to produce an initial
\draft of the \before function, conditioned on seven items from
Table~\ref{tab:context-items}: \IDFuncsig, \IDCallees, \IDCallsites,
\IDUsestmts, \IDStructimpl, \IDRepodesc, and \IDFuncdesc.  The developer's original code is
not given to the \rewriter to ensure the output reflects only the
\rewriter's own choices.
We filter out
\UseMacro{corpus-eval-rust-repos-target-callsites-removed}
functions where \IDCallsites is not available as we find no caller of
the \before function inside the repository, and
\UseMacro{corpus-eval-rust-repos-target-funcdesc-removed}
functions whose code and associated context are too long to fit inside the
\num{\UseMacro{corpus-eval-rust-repos-funcdesc-ctx}}-token window we
use to generate \IDFuncdesc. After this,
\num{\UseMacro{corpus-eval-rust-repos-draft-target-functions}}
functions remain.

\MyPara{Repair stage}
The generated \drafts often fail to build or fail the tests.  Thus,
all \drafts that fail to build or fail testing enter an iterative
repair loop
in the style of Self-Refine~\cite{madaan2023selfrefine}. In each round, we set up each
\rewriter model as two agents to fix its own \draft:
\begin{itemize}[topsep=0pt,itemsep=0ex,partopsep=0ex,parsep=0ex,leftmargin=*]
\item \textbf{\Instagent}: Receives the same context used
for the initial \draft, together with the previous round's
failing candidate, the build-and-test failure log, and the
developer's \before function. It produces a natural-language
repair plan that names the bug and prescribes the fix.
\item \textbf{\Implagent}: Receives the same context used
for the initial \draft, the previous round's failing
candidate, and the \instagent's repair plan. It outputs
a new repaired function.
\end{itemize}
We use the developer's \before function to help expedite the repair
process, and only the \instagent sees it during the loop.
This layer of indirection lets the \instagent diagnose the
candidate's bug against the ground-truth oracle while preventing the
\implagent from copying pieces of the original code verbatim.
Additionally, the \instagent's plan describes the failure in natural
language rather than the code that fixes it, making it unlikely to leak the
developer's style into the \reimplementation.
Figure~\ref{fig:instruct-plan-example} in
Appendix~\ref{sec:app-instruct-plan-example} shows one such plan with the corresponding
failing candidate.

The number of successfully \reimplemented functions varies across
\rewriters: \qwen produces passing \reimplementations for
\num{\UseMacro{edit-filter-eval-rust-repos-qwen-wide-passing}}
of those \before functions, \gpto for
\num{\UseMacro{edit-filter-eval-rust-repos-gpt-wide-passing}},
\olmol for
\num{\UseMacro{edit-filter-eval-rust-repos-olmo32b-wide-passing}},
\olmos for
\num{\UseMacro{edit-filter-eval-rust-repos-olmo7b-wide-passing}}, and
\haiku for
\num{\UseMacro{edit-filter-eval-rust-repos-haiku-wide-passing}}.

\MyPara{Filtering} We apply three filtering steps to the \reimplementations. First,
to ensure the \reimplemented function reflects the model's style and
does not include remembered code from training or any leaked style choices, we discard any
\reimplementation whose overlap with the \before function exceeds
50\%. Here overlap is the gestalt-pattern-matching score
implemented in Python's \CodeIn{difflib}.
Second, we discard any \task
on which the \reimplementation
already passes every \after test at \headcommit / \maxcommit, since then no edit is needed for that \reimplementation.
This happens when the
collected tests do not target the change at all, or when the
\reimplementation already exhibits the \after behavior before any
edit is applied.
Third, we discard any
\task whose \PR diff and related context are too big to fit inside the
\num{\UseMacro{corpus-eval-rust-repos-funcdesc-ctx}}-token window we
use to generate the \IDPRdesc (note that this is separate from \IDFuncdesc generation).

In the end, we are left with
\num{\UseMacro{fig-verl-comparison-allmean-base-qwen-n}} \tasks for
\qwen, \num{\UseMacro{fig-verl-comparison-allmean-base-gpt-n}} for \gpto,
\num{\UseMacro{fig-verl-comparison-allmean-base-olmo32b-n}} for \olmol,
\num{\UseMacro{fig-verl-comparison-allmean-base-olmo7b-n}} for \olmos, and
\num{\UseMacro{edit-filter-eval-rust-repos-haiku-after-nodesc}} for \haiku as
\rewriter. Every evaluation later is performed on these \tasks.
Table~\ref{tab:corpus-attrition-before} and
Table~\ref{tab:corpus-attrition-after} in
Appendix~\ref{sec:app-corpus-attrition} break down how many functions
remain after each filter.

\subsection{Edit}
\label{sec:edit}

\begin{table}[t]
\centering
\small
\setlength{\tabcolsep}{3pt}
\begin{tabular}{l c c c c c}
\toprule
\multirow{2}{*}{\textbf{Impl.}} & \multicolumn{5}{c}{\textbf{Editor}} \\
\cmidrule(lr){2-6}
& \textbf{\UseMacro{TCol-qwen}} & \textbf{\UseMacro{TCol-gpto}} & \textbf{\UseMacro{TCol-olmos}} & \textbf{\UseMacro{TCol-olmol}} & \textbf{\UseMacro{TCol-haiku}} \\
\midrule
\UseMacro{TColInline-qwen} & \textbf{\UseMacro{fig-pr-lev-minmax-norm-pivot-all-qwen3.5-thinking:35b-a3b-q4-k-m-qwen3.5-thinking:35b-a3b-q4-k-m-lev-char}} & \UseMacro{fig-pr-lev-minmax-norm-pivot-all-qwen3.5-thinking:35b-a3b-q4-k-m-gpt-oss:20b-q4-k-m-lev-char} & \UseMacro{fig-pr-lev-minmax-norm-pivot-all-qwen3.5-thinking:35b-a3b-q4-k-m-olmo3-thinking:7b.think-q4-k-m-lev-char} & \UseMacro{fig-pr-lev-minmax-norm-pivot-all-qwen3.5-thinking:35b-a3b-q4-k-m-olmo3.1-thinking:32b.think-q4-k-m-lev-char} & \UseMacro{fig-pr-lev-minmax-norm-pivot-all-qwen3.5-thinking:35b-a3b-q4-k-m-claude-haiku-4-5-lev-char} \\
\UseMacro{TColInline-gpto} & \UseMacro{fig-pr-lev-minmax-norm-pivot-all-gpt-oss:20b-q4-k-m-qwen3.5-thinking:35b-a3b-q4-k-m-lev-char} & \textbf{\UseMacro{fig-pr-lev-minmax-norm-pivot-all-gpt-oss:20b-q4-k-m-gpt-oss:20b-q4-k-m-lev-char}} & \UseMacro{fig-pr-lev-minmax-norm-pivot-all-gpt-oss:20b-q4-k-m-olmo3-thinking:7b.think-q4-k-m-lev-char} & \UseMacro{fig-pr-lev-minmax-norm-pivot-all-gpt-oss:20b-q4-k-m-olmo3.1-thinking:32b.think-q4-k-m-lev-char} & \textbf{\UseMacro{fig-pr-lev-minmax-norm-pivot-all-gpt-oss:20b-q4-k-m-claude-haiku-4-5-lev-char}} \\
\UseMacro{TColInline-olmos} & \UseMacro{fig-pr-lev-minmax-norm-pivot-all-olmo3-thinking:7b.think-q4-k-m-qwen3.5-thinking:35b-a3b-q4-k-m-lev-char} & \UseMacro{fig-pr-lev-minmax-norm-pivot-all-olmo3-thinking:7b.think-q4-k-m-gpt-oss:20b-q4-k-m-lev-char} & \UseMacro{fig-pr-lev-minmax-norm-pivot-all-olmo3-thinking:7b.think-q4-k-m-olmo3-thinking:7b.think-q4-k-m-lev-char} & \UseMacro{fig-pr-lev-minmax-norm-pivot-all-olmo3-thinking:7b.think-q4-k-m-olmo3.1-thinking:32b.think-q4-k-m-lev-char} & \UseMacro{fig-pr-lev-minmax-norm-pivot-all-olmo3-thinking:7b.think-q4-k-m-claude-haiku-4-5-lev-char} \\
\UseMacro{TColInline-olmol} & \UseMacro{fig-pr-lev-minmax-norm-pivot-all-olmo3.1-thinking:32b.think-q4-k-m-qwen3.5-thinking:35b-a3b-q4-k-m-lev-char} & \UseMacro{fig-pr-lev-minmax-norm-pivot-all-olmo3.1-thinking:32b.think-q4-k-m-gpt-oss:20b-q4-k-m-lev-char} & \UseMacro{fig-pr-lev-minmax-norm-pivot-all-olmo3.1-thinking:32b.think-q4-k-m-olmo3-thinking:7b.think-q4-k-m-lev-char} & \textbf{\UseMacro{fig-pr-lev-minmax-norm-pivot-all-olmo3.1-thinking:32b.think-q4-k-m-olmo3.1-thinking:32b.think-q4-k-m-lev-char}} & \UseMacro{fig-pr-lev-minmax-norm-pivot-all-olmo3.1-thinking:32b.think-q4-k-m-claude-haiku-4-5-lev-char} \\
\UseMacro{TColInline-haiku} & \UseMacro{fig-pr-lev-minmax-norm-pivot-all-claude-haiku-4-5-qwen3.5-thinking:35b-a3b-q4-k-m-lev-char} & \UseMacro{fig-pr-lev-minmax-norm-pivot-all-claude-haiku-4-5-gpt-oss:20b-q4-k-m-lev-char} & \textbf{\UseMacro{fig-pr-lev-minmax-norm-pivot-all-claude-haiku-4-5-olmo3-thinking:7b.think-q4-k-m-lev-char}} & \UseMacro{fig-pr-lev-minmax-norm-pivot-all-claude-haiku-4-5-olmo3.1-thinking:32b.think-q4-k-m-lev-char} & \UseMacro{fig-pr-lev-minmax-norm-pivot-all-claude-haiku-4-5-claude-haiku-4-5-lev-char} \\
\bottomrule
\end{tabular}
\caption{\UseMacro{TCap-pr-lev-minmax-norm-pivot-rewrite-char}}
\label{tab:pr-lev-minmax-norm-pivot-rewrite-char}
\end{table}

Each \editor then edits every \reimplemented \before function,
including the \reimplementations produced by itself as \rewriter,
according to the \prdesc. The \editors are given four additional items from
Table~\ref{tab:context-items} in the prompt: \after \IDFuncsig, \IDPRdesc, \IDRestprdiff, and
\IDCallees.
The \editor is asked to generate the full edited function and is allowed to add extra helper functions or structs, and those are counted as added characters and lines.

\subsection{Quantifying \selfediting vs. \crossediting}
\label{sec:quantify-crossedit}

\MyPara{Metric}
For each edit \task, we measure \levmetric between the \reimplementation and
the \editor's output at character granularity. Before
evaluating the distance, both the \reimplementation and the \editor's
output are passed through \texttt{rustfmt}, and then we strip comments and
blank lines from both sides, so that pure formatting differences and
comments do not inflate the distance. We use this formatted, comment-stripped \levmetric
throughout the rest of the paper. Since different edit \tasks can lead
to very different edit sizes, we min-max normalize each \task across
the \numofeditors \editors, mapping each edit to a value in $[0,1]$
where 0 is the smallest edit on that \task across the \numofeditors
\editors and 1 is the largest:
$$\frac{\mathrm{lev}_i^{\mathit{model}} - \min_m \mathrm{lev}_i^{m}}{\max_m \mathrm{lev}_i^{m} - \min_m \mathrm{lev}_i^{m}}.$$
Here $i$ indexes \tasks and $m$ indexes \editors.

Additionally, to test the statistical significance of whether an
\editor's \selfediting distances are smaller than its \crossediting
ones, we pair each \editor with each \rewriter other than itself and
run a one-sided Mann-Whitney U (MWU) test. The test compares every
\selfediting distance against every \crossediting one and counts how
often the \selfediting distance is smaller. In this way, it
detects whether one set tends to fall below the other.

\begin{table}[t]
\centering
\small
\setlength{\tabcolsep}{3pt}
\begin{tabular}{l c c c c c}
\toprule
\multirow{2}{*}{\textbf{Impl.}} & \multicolumn{5}{c}{\textbf{Editor}} \\
\cmidrule(lr){2-6}
& \textbf{\UseMacro{TCol-qwen}} & \textbf{\UseMacro{TCol-gpto}} & \textbf{\UseMacro{TCol-olmos}} & \textbf{\UseMacro{TCol-olmol}} & \textbf{\UseMacro{TCol-haiku}} \\
\midrule
\UseMacro{TColInline-qwen} & -- & \UseMacro{fig-pr-lev-selfcross-mwu-qwen3.5-thinking:35b-a3b-q4-k-m-gpt-oss:20b-q4-k-m-p}\UseMacro{fig-pr-lev-selfcross-mwu-qwen3.5-thinking:35b-a3b-q4-k-m-gpt-oss:20b-q4-k-m-stars} & \UseMacro{fig-pr-lev-selfcross-mwu-qwen3.5-thinking:35b-a3b-q4-k-m-olmo3-thinking:7b.think-q4-k-m-p}\UseMacro{fig-pr-lev-selfcross-mwu-qwen3.5-thinking:35b-a3b-q4-k-m-olmo3-thinking:7b.think-q4-k-m-stars} & \UseMacro{fig-pr-lev-selfcross-mwu-qwen3.5-thinking:35b-a3b-q4-k-m-olmo3.1-thinking:32b.think-q4-k-m-p}\UseMacro{fig-pr-lev-selfcross-mwu-qwen3.5-thinking:35b-a3b-q4-k-m-olmo3.1-thinking:32b.think-q4-k-m-stars} & \UseMacro{fig-pr-lev-selfcross-mwu-qwen3.5-thinking:35b-a3b-q4-k-m-claude-haiku-4-5-p}\UseMacro{fig-pr-lev-selfcross-mwu-qwen3.5-thinking:35b-a3b-q4-k-m-claude-haiku-4-5-stars} \\
\UseMacro{TColInline-gpto} & \UseMacro{fig-pr-lev-selfcross-mwu-gpt-oss:20b-q4-k-m-qwen3.5-thinking:35b-a3b-q4-k-m-p}\UseMacro{fig-pr-lev-selfcross-mwu-gpt-oss:20b-q4-k-m-qwen3.5-thinking:35b-a3b-q4-k-m-stars} & -- & \UseMacro{fig-pr-lev-selfcross-mwu-gpt-oss:20b-q4-k-m-olmo3-thinking:7b.think-q4-k-m-p}\UseMacro{fig-pr-lev-selfcross-mwu-gpt-oss:20b-q4-k-m-olmo3-thinking:7b.think-q4-k-m-stars} & \UseMacro{fig-pr-lev-selfcross-mwu-gpt-oss:20b-q4-k-m-olmo3.1-thinking:32b.think-q4-k-m-p}\UseMacro{fig-pr-lev-selfcross-mwu-gpt-oss:20b-q4-k-m-olmo3.1-thinking:32b.think-q4-k-m-stars} & \UseMacro{fig-pr-lev-selfcross-mwu-gpt-oss:20b-q4-k-m-claude-haiku-4-5-p}\UseMacro{fig-pr-lev-selfcross-mwu-gpt-oss:20b-q4-k-m-claude-haiku-4-5-stars} \\
\UseMacro{TColInline-olmos} & \UseMacro{fig-pr-lev-selfcross-mwu-olmo3-thinking:7b.think-q4-k-m-qwen3.5-thinking:35b-a3b-q4-k-m-p}\UseMacro{fig-pr-lev-selfcross-mwu-olmo3-thinking:7b.think-q4-k-m-qwen3.5-thinking:35b-a3b-q4-k-m-stars} & \UseMacro{fig-pr-lev-selfcross-mwu-olmo3-thinking:7b.think-q4-k-m-gpt-oss:20b-q4-k-m-p}\UseMacro{fig-pr-lev-selfcross-mwu-olmo3-thinking:7b.think-q4-k-m-gpt-oss:20b-q4-k-m-stars} & -- & \UseMacro{fig-pr-lev-selfcross-mwu-olmo3-thinking:7b.think-q4-k-m-olmo3.1-thinking:32b.think-q4-k-m-p}\UseMacro{fig-pr-lev-selfcross-mwu-olmo3-thinking:7b.think-q4-k-m-olmo3.1-thinking:32b.think-q4-k-m-stars} & \UseMacro{fig-pr-lev-selfcross-mwu-olmo3-thinking:7b.think-q4-k-m-claude-haiku-4-5-p}\UseMacro{fig-pr-lev-selfcross-mwu-olmo3-thinking:7b.think-q4-k-m-claude-haiku-4-5-stars} \\
\UseMacro{TColInline-olmol} & \UseMacro{fig-pr-lev-selfcross-mwu-olmo3.1-thinking:32b.think-q4-k-m-qwen3.5-thinking:35b-a3b-q4-k-m-p}\UseMacro{fig-pr-lev-selfcross-mwu-olmo3.1-thinking:32b.think-q4-k-m-qwen3.5-thinking:35b-a3b-q4-k-m-stars} & \UseMacro{fig-pr-lev-selfcross-mwu-olmo3.1-thinking:32b.think-q4-k-m-gpt-oss:20b-q4-k-m-p}\UseMacro{fig-pr-lev-selfcross-mwu-olmo3.1-thinking:32b.think-q4-k-m-gpt-oss:20b-q4-k-m-stars} & \UseMacro{fig-pr-lev-selfcross-mwu-olmo3.1-thinking:32b.think-q4-k-m-olmo3-thinking:7b.think-q4-k-m-p}\UseMacro{fig-pr-lev-selfcross-mwu-olmo3.1-thinking:32b.think-q4-k-m-olmo3-thinking:7b.think-q4-k-m-stars} & -- & \UseMacro{fig-pr-lev-selfcross-mwu-olmo3.1-thinking:32b.think-q4-k-m-claude-haiku-4-5-p}\UseMacro{fig-pr-lev-selfcross-mwu-olmo3.1-thinking:32b.think-q4-k-m-claude-haiku-4-5-stars} \\
\UseMacro{TColInline-haiku} & \UseMacro{fig-pr-lev-selfcross-mwu-claude-haiku-4-5-qwen3.5-thinking:35b-a3b-q4-k-m-p}\UseMacro{fig-pr-lev-selfcross-mwu-claude-haiku-4-5-qwen3.5-thinking:35b-a3b-q4-k-m-stars} & \UseMacro{fig-pr-lev-selfcross-mwu-claude-haiku-4-5-gpt-oss:20b-q4-k-m-p}\UseMacro{fig-pr-lev-selfcross-mwu-claude-haiku-4-5-gpt-oss:20b-q4-k-m-stars} & \UseMacro{fig-pr-lev-selfcross-mwu-claude-haiku-4-5-olmo3-thinking:7b.think-q4-k-m-p}\UseMacro{fig-pr-lev-selfcross-mwu-claude-haiku-4-5-olmo3-thinking:7b.think-q4-k-m-stars} & \UseMacro{fig-pr-lev-selfcross-mwu-claude-haiku-4-5-olmo3.1-thinking:32b.think-q4-k-m-p}\UseMacro{fig-pr-lev-selfcross-mwu-claude-haiku-4-5-olmo3.1-thinking:32b.think-q4-k-m-stars} & -- \\
\bottomrule
\end{tabular}
\caption{\UseMacro{TCap-pr-lev-selfcross-mwu}}
\label{tab:pr-lev-selfcross-mwu}
\end{table}

\MyPara{Results}
Table~\ref{tab:pr-lev-minmax-norm-pivot-rewrite-char} reports the
normalized matrix, and Table~\ref{tab:pr-lev-selfcross-mwu} the $p$
values of the MWU test.
The normalization is defined only on \tasks where every \editor
actually generated an edit and where the \editors did not all make identically sized
edits, which would cause the denominator to vanish, so we restrict this analysis to those. That leaves
\num{\UseMacro{fig-pr-lev-minmax-norm-pivot-all-qwen3.5-thinking:35b-a3b-q4-k-m-qwen3.5-thinking:35b-a3b-q4-k-m-lev-char-n}} \tasks for \qwen,
\num{\UseMacro{fig-pr-lev-minmax-norm-pivot-all-gpt-oss:20b-q4-k-m-gpt-oss:20b-q4-k-m-lev-char-n}} for \gpto,
\num{\UseMacro{fig-pr-lev-minmax-norm-pivot-all-olmo3.1-thinking:32b.think-q4-k-m-olmo3.1-thinking:32b.think-q4-k-m-lev-char-n}} for \olmol,
\num{\UseMacro{fig-pr-lev-minmax-norm-pivot-all-olmo3-thinking:7b.think-q4-k-m-olmo3-thinking:7b.think-q4-k-m-lev-char-n}} for \olmos, and
\num{\UseMacro{fig-pr-lev-minmax-norm-pivot-all-claude-haiku-4-5-claude-haiku-4-5-lev-char-n}} for \haiku as \rewriter.
We observe that the \numofopeneditors open-weight \editors tend to edit their own
\reimplementations the least: among the
\UseMacro{fig-pr-lev-selfcross-mwu-open-n-tests} self-versus-cross
comparisons,
\UseMacro{fig-pr-lev-selfcross-mwu-open-n-right-direction} show
\selfediting making the smaller change, and
\UseMacro{fig-pr-lev-selfcross-mwu-open-n-significant} are significant
at $p<0.05$. Both exceptions are \olmos, whose \selfediting change is
larger than its \crossediting change on \olmol's and on \haiku's
\reimplementations.

\haiku does not follow this pattern. Its column minimum falls on
\gpto's \reimplementations. Why \haiku, unlike the open-weight models, shows little difference in edit
distance between its own code and \othercode is an
interesting question, and we believe it is an exciting
direction for future work.

The \editors also differ in
overall edit size: \olmos is the most aggressive \editor, editing
about twice as much as the other models on \reimplementations by
\qwen and \gpto. The gap shrinks substantially on \olmos's own
\reimplementations, reinforcing the broader pattern that the
open-weight models edit more on code from other models than on their own.
These results indicate that \othercode induces excessive edits. This holds
even for the \editors that make smaller edits in general, which still change
more of \othercode than of their own.

Additionally, we report the test pass rate for all combinations of \editors
and \rewriters in Appendix~\ref{sec:app-edit-success-matrix}, where the
diagonal entry is not always the column maximum.

\section{\Tool}

To mitigate the excessive editing problem observed in
Section~\ref{sec:quantify-crossedit}, we propose \Tool,
a post-training framework for code editing models with a novel reward function
consisting of two components: a \simreward that penalizes
large edits, and an \execreward that scores whether the edits successfully complete
the given \tasks. We optimize the model with Group Relative Policy
Optimization (GRPO)~\cite{shao2024deepseekmath} to reduce the
magnitude of edits made by \LLMs while preserving functional
correctness.

\subsection{Task Formulation}

We frame function editing as a sequence-level reinforcement learning
problem. Given the \reimplemented \before function
$f^A_{\text{\before}}$, the \task description $d$, and the context
$\mathcal{C}$, a policy $\pi_\theta$ generates an edited function
$f^{A\to B}_{\text{\after}} = \pi_\theta(f^A_{\text{\before}}, d,
\mathcal{C})$. Here, the context is the same as that supplied
in the edit step of Section~\ref{sec:edit}. Each rollout is scored
with a single scalar reward $R_{\text{\Tool}} = R_{\text{sim}} \times
R_{\text{exec}}$. This reward is then used with GRPO to optimize
$\pi_\theta$. The objective is to produce edits that satisfy the
\task's functional requirements while introducing minimal unnecessary
change.

\vspace{-5pt}
\subsection{\SimReward}

The similarity reward penalizes excessive changes. For each function
edit made by the \LLM being trained, let $c_m$ and $l_m$ be the number
of characters and lines changed by the model. We compute
$D_m = (c_m + 1)(l_m + 1)$ to capture both how much text changed, via
the character count, and how many locations were touched, via the line
count. Then, to regularize it, we take a similar approach to GR3's
length-rescaling reward~\cite{li2026gr3}. Their reward deals with the problem of response-length inflation by rescaling a base reward
by response length normalized against the GRPO group mean, discouraging the
model from producing arbitrarily long generations. We instead
normalize against the per-\task developer edit magnitude
$D_h = (c_h + 1)(l_h + 1)$, where $c_h$ and $l_h$ are the characters
and lines changed inside the developer-written function in the
original \PR. $D_h$ estimates the ``\task size'', i.e., how much change
is needed to complete the edit \task. Our similarity reward is
$$R_{\text{sim}} = \frac{1}{1 + \alpha\frac{D_m}{D_h}}.$$
This allows us to penalize large edits proportionally to the estimated
edit size of the \task. The coefficient $\alpha$ controls the penalty
steepness. An edit matching the developer's edit size scores
$\frac{1}{1+\alpha}$.

\subsection{\ExecReward}

The execution reward $R_{\text{exec}}$ scores whether the edited code
builds and passes the function's tests. We compute $R_{\text{exec}} =
w_b b + w_{\text{pre}} p_{\text{pre-edit}} + w_{\text{post}}
p_{\text{post-edit}}$, where $b$ is build success (0 or 1),
$p_{\text{pre-edit}}$ is the edit's pass rate on tests that the
\reimplemented function already passes (regressions the edit should
avoid), and $p_{\text{post-edit}}$ is the edit's pass rate on tests
that the \reimplemented function fails (new behavior the edit must
introduce). The weights $w_b$, $w_{\text{pre}}$, and $w_{\text{post}}$ allow us
to control which metrics to prioritize. In training, we set $w_b = 0.2$,
$w_{\text{pre}} = 0.4$, and $w_{\text{post}} = 0.4$.

\begin{table}[t]
\centering
\small
\setlength{\tabcolsep}{3pt}
\begin{tabular}{l c c c c}
\toprule
\multirow{2}{*}{\shortstack[l]{\textbf{Implementor} \\ \quad\textbf{Editor}}} & \multirow{2}{*}{\shortstack{\textbf{All} \\ \textbf{\tasks}}} & \multicolumn{3}{c}{\textbf{Both pass}} \\
\cmidrule(lr){3-5}
 & & \shortstack{\textbf{Base \&} \\ \textbf{Strict}} & \shortstack{\textbf{Base \&} \\ \textbf{RL}} & \shortstack{\textbf{Strict} \\ \textbf{\& RL}} \\
\midrule
\multicolumn{5}{@{}l}{\textbf{\qwen}} \\
\quad \baselabel & \SmallestBoldSafe{\UseMacro{fig-verl-comparison-allmean-rustfmt-minmax-base-qwen-lev-char}}{\UseMacro{fig-verl-comparison-allmean-rustfmt-minmax-prompt-qwen-lev-char}}{\UseMacro{fig-verl-comparison-allmean-rustfmt-minmax-rl-qwen-lev-char}} & \SmallerBoldSafe{\UseMacro{fig-verl-lev-base-strict-rustfmt-minmax-base-qwen-lev-char}}{\UseMacro{fig-verl-lev-base-strict-rustfmt-minmax-prompt-qwen-lev-char}}\UseMacro{fig-verl-lev-base-strict-rustfmt-minmax-qwen-lev-char-small} & \SmallerBoldSafe{\UseMacro{fig-verl-lev-both-pass-rustfmt-minmax-base-qwen-lev-char}}{\UseMacro{fig-verl-lev-both-pass-rustfmt-minmax-rl-qwen-lev-char}}\UseMacro{fig-verl-lev-both-pass-rustfmt-minmax-qwen-lev-char-small} & -- \\
\quad \strictlabel & \SmallestBoldSafe{\UseMacro{fig-verl-comparison-allmean-rustfmt-minmax-prompt-qwen-lev-char}}{\UseMacro{fig-verl-comparison-allmean-rustfmt-minmax-base-qwen-lev-char}}{\UseMacro{fig-verl-comparison-allmean-rustfmt-minmax-rl-qwen-lev-char}} & \SmallerBoldSafe{\UseMacro{fig-verl-lev-base-strict-rustfmt-minmax-prompt-qwen-lev-char}}{\UseMacro{fig-verl-lev-base-strict-rustfmt-minmax-base-qwen-lev-char}}\UseMacro{fig-verl-lev-base-strict-rustfmt-minmax-qwen-lev-char-small} & -- & \SmallerBoldSafe{\UseMacro{fig-verl-lev-prompt-pass-rustfmt-minmax-prompt-qwen-lev-char}}{\UseMacro{fig-verl-lev-prompt-pass-rustfmt-minmax-rl-qwen-lev-char}}\UseMacro{fig-verl-lev-prompt-pass-rustfmt-minmax-qwen-lev-char-small} \\
\quad \rllabel & \SmallestBoldSafe{\UseMacro{fig-verl-comparison-allmean-rustfmt-minmax-rl-qwen-lev-char}}{\UseMacro{fig-verl-comparison-allmean-rustfmt-minmax-base-qwen-lev-char}}{\UseMacro{fig-verl-comparison-allmean-rustfmt-minmax-prompt-qwen-lev-char}} & -- & \SmallerBoldSafe{\UseMacro{fig-verl-lev-both-pass-rustfmt-minmax-rl-qwen-lev-char}}{\UseMacro{fig-verl-lev-both-pass-rustfmt-minmax-base-qwen-lev-char}}\UseMacro{fig-verl-lev-both-pass-rustfmt-minmax-qwen-lev-char-small} & \SmallerBoldSafe{\UseMacro{fig-verl-lev-prompt-pass-rustfmt-minmax-rl-qwen-lev-char}}{\UseMacro{fig-verl-lev-prompt-pass-rustfmt-minmax-prompt-qwen-lev-char}}\UseMacro{fig-verl-lev-prompt-pass-rustfmt-minmax-qwen-lev-char-small} \\
\midrule
\multicolumn{5}{@{}l}{\textbf{\gpto}} \\
\quad \baselabel & \SmallestBoldSafe{\UseMacro{fig-verl-comparison-allmean-rustfmt-minmax-base-gpt-lev-char}}{\UseMacro{fig-verl-comparison-allmean-rustfmt-minmax-prompt-gpt-lev-char}}{\UseMacro{fig-verl-comparison-allmean-rustfmt-minmax-rl-gpt-lev-char}} & \SmallerBoldSafe{\UseMacro{fig-verl-lev-base-strict-rustfmt-minmax-base-gpt-lev-char}}{\UseMacro{fig-verl-lev-base-strict-rustfmt-minmax-prompt-gpt-lev-char}}\UseMacro{fig-verl-lev-base-strict-rustfmt-minmax-gpt-lev-char-small} & \SmallerBoldSafe{\UseMacro{fig-verl-lev-both-pass-rustfmt-minmax-base-gpt-lev-char}}{\UseMacro{fig-verl-lev-both-pass-rustfmt-minmax-rl-gpt-lev-char}}\UseMacro{fig-verl-lev-both-pass-rustfmt-minmax-gpt-lev-char-small} & -- \\
\quad \strictlabel & \SmallestBoldSafe{\UseMacro{fig-verl-comparison-allmean-rustfmt-minmax-prompt-gpt-lev-char}}{\UseMacro{fig-verl-comparison-allmean-rustfmt-minmax-base-gpt-lev-char}}{\UseMacro{fig-verl-comparison-allmean-rustfmt-minmax-rl-gpt-lev-char}} & \SmallerBoldSafe{\UseMacro{fig-verl-lev-base-strict-rustfmt-minmax-prompt-gpt-lev-char}}{\UseMacro{fig-verl-lev-base-strict-rustfmt-minmax-base-gpt-lev-char}}\UseMacro{fig-verl-lev-base-strict-rustfmt-minmax-gpt-lev-char-small} & -- & \SmallerBoldSafe{\UseMacro{fig-verl-lev-prompt-pass-rustfmt-minmax-prompt-gpt-lev-char}}{\UseMacro{fig-verl-lev-prompt-pass-rustfmt-minmax-rl-gpt-lev-char}}\UseMacro{fig-verl-lev-prompt-pass-rustfmt-minmax-gpt-lev-char-small} \\
\quad \rllabel & \SmallestBoldSafe{\UseMacro{fig-verl-comparison-allmean-rustfmt-minmax-rl-gpt-lev-char}}{\UseMacro{fig-verl-comparison-allmean-rustfmt-minmax-base-gpt-lev-char}}{\UseMacro{fig-verl-comparison-allmean-rustfmt-minmax-prompt-gpt-lev-char}} & -- & \SmallerBoldSafe{\UseMacro{fig-verl-lev-both-pass-rustfmt-minmax-rl-gpt-lev-char}}{\UseMacro{fig-verl-lev-both-pass-rustfmt-minmax-base-gpt-lev-char}}\UseMacro{fig-verl-lev-both-pass-rustfmt-minmax-gpt-lev-char-small} & \SmallerBoldSafe{\UseMacro{fig-verl-lev-prompt-pass-rustfmt-minmax-rl-gpt-lev-char}}{\UseMacro{fig-verl-lev-prompt-pass-rustfmt-minmax-prompt-gpt-lev-char}}\UseMacro{fig-verl-lev-prompt-pass-rustfmt-minmax-gpt-lev-char-small} \\
\midrule
\multicolumn{5}{@{}l}{\textbf{\olmos}} \\
\quad \baselabel & \SmallestBoldSafe{\UseMacro{fig-verl-comparison-allmean-rustfmt-minmax-base-olmo7b-lev-char}}{\UseMacro{fig-verl-comparison-allmean-rustfmt-minmax-prompt-olmo7b-lev-char}}{\UseMacro{fig-verl-comparison-allmean-rustfmt-minmax-rl-olmo7b-lev-char}} & \SmallerBoldSafe{\UseMacro{fig-verl-lev-base-strict-rustfmt-minmax-base-olmo7b-lev-char}}{\UseMacro{fig-verl-lev-base-strict-rustfmt-minmax-prompt-olmo7b-lev-char}}\UseMacro{fig-verl-lev-base-strict-rustfmt-minmax-olmo7b-lev-char-small} & \SmallerBoldSafe{\UseMacro{fig-verl-lev-both-pass-rustfmt-minmax-base-olmo7b-lev-char}}{\UseMacro{fig-verl-lev-both-pass-rustfmt-minmax-rl-olmo7b-lev-char}}\UseMacro{fig-verl-lev-both-pass-rustfmt-minmax-olmo7b-lev-char-small} & -- \\
\quad \strictlabel & \SmallestBoldSafe{\UseMacro{fig-verl-comparison-allmean-rustfmt-minmax-prompt-olmo7b-lev-char}}{\UseMacro{fig-verl-comparison-allmean-rustfmt-minmax-base-olmo7b-lev-char}}{\UseMacro{fig-verl-comparison-allmean-rustfmt-minmax-rl-olmo7b-lev-char}} & \SmallerBoldSafe{\UseMacro{fig-verl-lev-base-strict-rustfmt-minmax-prompt-olmo7b-lev-char}}{\UseMacro{fig-verl-lev-base-strict-rustfmt-minmax-base-olmo7b-lev-char}}\UseMacro{fig-verl-lev-base-strict-rustfmt-minmax-olmo7b-lev-char-small} & -- & \SmallerBoldSafe{\UseMacro{fig-verl-lev-prompt-pass-rustfmt-minmax-prompt-olmo7b-lev-char}}{\UseMacro{fig-verl-lev-prompt-pass-rustfmt-minmax-rl-olmo7b-lev-char}}\UseMacro{fig-verl-lev-prompt-pass-rustfmt-minmax-olmo7b-lev-char-small} \\
\quad \rllabel & \SmallestBoldSafe{\UseMacro{fig-verl-comparison-allmean-rustfmt-minmax-rl-olmo7b-lev-char}}{\UseMacro{fig-verl-comparison-allmean-rustfmt-minmax-base-olmo7b-lev-char}}{\UseMacro{fig-verl-comparison-allmean-rustfmt-minmax-prompt-olmo7b-lev-char}} & -- & \SmallerBoldSafe{\UseMacro{fig-verl-lev-both-pass-rustfmt-minmax-rl-olmo7b-lev-char}}{\UseMacro{fig-verl-lev-both-pass-rustfmt-minmax-base-olmo7b-lev-char}}\UseMacro{fig-verl-lev-both-pass-rustfmt-minmax-olmo7b-lev-char-small} & \SmallerBoldSafe{\UseMacro{fig-verl-lev-prompt-pass-rustfmt-minmax-rl-olmo7b-lev-char}}{\UseMacro{fig-verl-lev-prompt-pass-rustfmt-minmax-prompt-olmo7b-lev-char}}\UseMacro{fig-verl-lev-prompt-pass-rustfmt-minmax-olmo7b-lev-char-small} \\
\midrule
\multicolumn{5}{@{}l}{\textbf{\olmol}} \\
\quad \baselabel & \SmallestBoldSafe{\UseMacro{fig-verl-comparison-allmean-rustfmt-minmax-base-olmo32b-lev-char}}{\UseMacro{fig-verl-comparison-allmean-rustfmt-minmax-prompt-olmo32b-lev-char}}{\UseMacro{fig-verl-comparison-allmean-rustfmt-minmax-rl-olmo32b-lev-char}} & \SmallerBoldSafe{\UseMacro{fig-verl-lev-base-strict-rustfmt-minmax-base-olmo32b-lev-char}}{\UseMacro{fig-verl-lev-base-strict-rustfmt-minmax-prompt-olmo32b-lev-char}}\UseMacro{fig-verl-lev-base-strict-rustfmt-minmax-olmo32b-lev-char-small} & \SmallerBoldSafe{\UseMacro{fig-verl-lev-both-pass-rustfmt-minmax-base-olmo32b-lev-char}}{\UseMacro{fig-verl-lev-both-pass-rustfmt-minmax-rl-olmo32b-lev-char}}\UseMacro{fig-verl-lev-both-pass-rustfmt-minmax-olmo32b-lev-char-small} & -- \\
\quad \strictlabel & \SmallestBoldSafe{\UseMacro{fig-verl-comparison-allmean-rustfmt-minmax-prompt-olmo32b-lev-char}}{\UseMacro{fig-verl-comparison-allmean-rustfmt-minmax-base-olmo32b-lev-char}}{\UseMacro{fig-verl-comparison-allmean-rustfmt-minmax-rl-olmo32b-lev-char}} & \SmallerBoldSafe{\UseMacro{fig-verl-lev-base-strict-rustfmt-minmax-prompt-olmo32b-lev-char}}{\UseMacro{fig-verl-lev-base-strict-rustfmt-minmax-base-olmo32b-lev-char}}\UseMacro{fig-verl-lev-base-strict-rustfmt-minmax-olmo32b-lev-char-small} & -- & \SmallerBoldSafe{\UseMacro{fig-verl-lev-prompt-pass-rustfmt-minmax-prompt-olmo32b-lev-char}}{\UseMacro{fig-verl-lev-prompt-pass-rustfmt-minmax-rl-olmo32b-lev-char}}\UseMacro{fig-verl-lev-prompt-pass-rustfmt-minmax-olmo32b-lev-char-small} \\
\quad \rllabel & \SmallestBoldSafe{\UseMacro{fig-verl-comparison-allmean-rustfmt-minmax-rl-olmo32b-lev-char}}{\UseMacro{fig-verl-comparison-allmean-rustfmt-minmax-base-olmo32b-lev-char}}{\UseMacro{fig-verl-comparison-allmean-rustfmt-minmax-prompt-olmo32b-lev-char}} & -- & \SmallerBoldSafe{\UseMacro{fig-verl-lev-both-pass-rustfmt-minmax-rl-olmo32b-lev-char}}{\UseMacro{fig-verl-lev-both-pass-rustfmt-minmax-base-olmo32b-lev-char}}\UseMacro{fig-verl-lev-both-pass-rustfmt-minmax-olmo32b-lev-char-small} & \SmallerBoldSafe{\UseMacro{fig-verl-lev-prompt-pass-rustfmt-minmax-rl-olmo32b-lev-char}}{\UseMacro{fig-verl-lev-prompt-pass-rustfmt-minmax-prompt-olmo32b-lev-char}}\UseMacro{fig-verl-lev-prompt-pass-rustfmt-minmax-olmo32b-lev-char-small} \\
\midrule
\multicolumn{5}{@{}l}{\textbf{\haiku}} \\
\quad \baselabel & \SmallestBoldSafe{\UseMacro{fig-verl-comparison-allmean-rustfmt-minmax-base-haiku-lev-char}}{\UseMacro{fig-verl-comparison-allmean-rustfmt-minmax-prompt-haiku-lev-char}}{\UseMacro{fig-verl-comparison-allmean-rustfmt-minmax-rl-haiku-lev-char}} & \SmallerBoldSafe{\UseMacro{fig-verl-lev-base-strict-rustfmt-minmax-base-haiku-lev-char}}{\UseMacro{fig-verl-lev-base-strict-rustfmt-minmax-prompt-haiku-lev-char}}\UseMacro{fig-verl-lev-base-strict-rustfmt-minmax-haiku-lev-char-small} & \SmallerBoldSafe{\UseMacro{fig-verl-lev-both-pass-rustfmt-minmax-base-haiku-lev-char}}{\UseMacro{fig-verl-lev-both-pass-rustfmt-minmax-rl-haiku-lev-char}}\UseMacro{fig-verl-lev-both-pass-rustfmt-minmax-haiku-lev-char-small} & -- \\
\quad \strictlabel & \SmallestBoldSafe{\UseMacro{fig-verl-comparison-allmean-rustfmt-minmax-prompt-haiku-lev-char}}{\UseMacro{fig-verl-comparison-allmean-rustfmt-minmax-base-haiku-lev-char}}{\UseMacro{fig-verl-comparison-allmean-rustfmt-minmax-rl-haiku-lev-char}} & \SmallerBoldSafe{\UseMacro{fig-verl-lev-base-strict-rustfmt-minmax-prompt-haiku-lev-char}}{\UseMacro{fig-verl-lev-base-strict-rustfmt-minmax-base-haiku-lev-char}}\UseMacro{fig-verl-lev-base-strict-rustfmt-minmax-haiku-lev-char-small} & -- & \SmallerBoldSafe{\UseMacro{fig-verl-lev-prompt-pass-rustfmt-minmax-prompt-haiku-lev-char}}{\UseMacro{fig-verl-lev-prompt-pass-rustfmt-minmax-rl-haiku-lev-char}}\UseMacro{fig-verl-lev-prompt-pass-rustfmt-minmax-haiku-lev-char-small} \\
\quad \rllabel & \SmallestBoldSafe{\UseMacro{fig-verl-comparison-allmean-rustfmt-minmax-rl-haiku-lev-char}}{\UseMacro{fig-verl-comparison-allmean-rustfmt-minmax-base-haiku-lev-char}}{\UseMacro{fig-verl-comparison-allmean-rustfmt-minmax-prompt-haiku-lev-char}} & -- & \SmallerBoldSafe{\UseMacro{fig-verl-lev-both-pass-rustfmt-minmax-rl-haiku-lev-char}}{\UseMacro{fig-verl-lev-both-pass-rustfmt-minmax-base-haiku-lev-char}}\UseMacro{fig-verl-lev-both-pass-rustfmt-minmax-haiku-lev-char-small} & \SmallerBoldSafe{\UseMacro{fig-verl-lev-prompt-pass-rustfmt-minmax-rl-haiku-lev-char}}{\UseMacro{fig-verl-lev-prompt-pass-rustfmt-minmax-prompt-haiku-lev-char}}\UseMacro{fig-verl-lev-prompt-pass-rustfmt-minmax-haiku-lev-char-small} \\
\bottomrule
\end{tabular}
\caption{\UseMacro{TCap-verl-lev-norm-eval-rust-repos}}
\label{tab:verl-lev-norm-eval-rust-repos}
\end{table}

\section{Experiment}
\label{sec:experiment}

\subsection{Training Dataset}

The training data for \Tool is built with the same pipeline described
in Section~\ref{sec:char-crossediting}, applied to a separate pool of
repositories to ensure there is no project-level overlap with the evaluation dataset.
The starting set contains
\num{\UseMacro{corpus-training-rust-repos-fn-level-functions}} functions, \num{\UseMacro{corpus-training-rust-repos-fn-level-prs}} \PRs,
and \num{\UseMacro{corpus-training-rust-repos-fn-level-repos}} repositories.
We are able to collect tests for
\num{\UseMacro{corpus-training-rust-repos-tests-functions}} functions, \num{\UseMacro{corpus-training-rust-repos-tests-prs}} \PRs,
and \num{\UseMacro{corpus-training-rust-repos-tests-repos}} repositories. To \reimplement the collected \before functions,
we use the \qwen model with the same \draft and repair procedure as in
Section~\ref{sec:reimplement}, with the repair loop run for up to
\UseMacro{corpus-training-rust-repos-repair-rounds} rounds rather than
\UseMacro{corpus-eval-rust-repos-repair-rounds}. We drop the overlap
filter that caps the similarity between the \reimplemented function
and the original developer's code, since the concern about bias from
the model remembering does not apply at training time. We add an additional filter for the \tasks whose prompts do not fit within the
\num{\UseMacro{verl-max-prompt-tokens}}-token limit (\num{\UseMacro{verl-max-seq-length}}-token limit on total context, \num{\UseMacro{verl-max-new-tokens}} tokens reserved for the output). In total,
we collected \num{\UseMacro{verl-train-prompts}} \crossedit \tasks for
training and \num{\UseMacro{verl-val-prompts}} for validation.

\subsection{Training Setup}

We use \olmos as our \basemodel for training, since it makes the most aggressive
edits in the evaluation of Section~\ref{sec:quantify-crossedit}. It is
adapted with LoRA~\cite{hu2022lora} at rank 32 and alpha 64 on all
linear modules. Since LoRA keeps the base weights frozen, users have the option to load the adapter on demand at inference time for \crossedit
\tasks. GRPO training is implemented with the VERL
framework~\cite{sheng2025verl}. We train for 3 epochs on our dataset
with a batch size of 32 prompts and 8 rollouts per prompt, sampled at
temperature 1.0. Our optimizer uses a learning rate of
$3 \times 10^{-5}$ with a KL regularization coefficient of 0.001. The
similarity reward sets $\alpha = 0.33$ following GR3~\cite{li2026gr3}.

\subsection{Baselines}
\label{sec:experiment-baselines}

We compare the \rlmodel (denoted as \rllabel) against two baselines on
the same \tasks. The first is the untrained \basemodel (denoted as
\baselabel). The second is the \minprompt (denoted as \strictlabel),
which explores whether excessive editing can be resolved by prompt
engineering. Its system prompt states the same instruction as for the other \editors but adds strict
guidelines to change only what the edit requires and to leave the rest
of the function as it is, while the edit prompt and the decoding
settings stay unchanged. The guidelines list out what should be left
alone for a minimal edit when possible, among them the control flow, the type and
variable names, and the error handling style. We show the prompt in
Figure~\ref{fig:prompt-edit-limit-guidelines} in Appendix~\ref{sec:app-edit-limit-prompt}.

\section{Results}
\label{sec:results}

We analyze \Tool from two quantitative angles, the size of the edits
it produces (Section~\ref{sec:results-edit-size}) and whether those
edits build and pass tests (Section~\ref{sec:results-edit-success}).
We then complement these with a qualitative analysis
(Section~\ref{sec:results-qualitative}) of cases where \Tool and the
\basemodel diverge.

\subsection{Effect on Edit Size}
\label{sec:results-edit-size}

\begin{figure*}[t]
\centering
\begin{minipage}{\linewidth}
\includegraphics[width=0.41\linewidth]{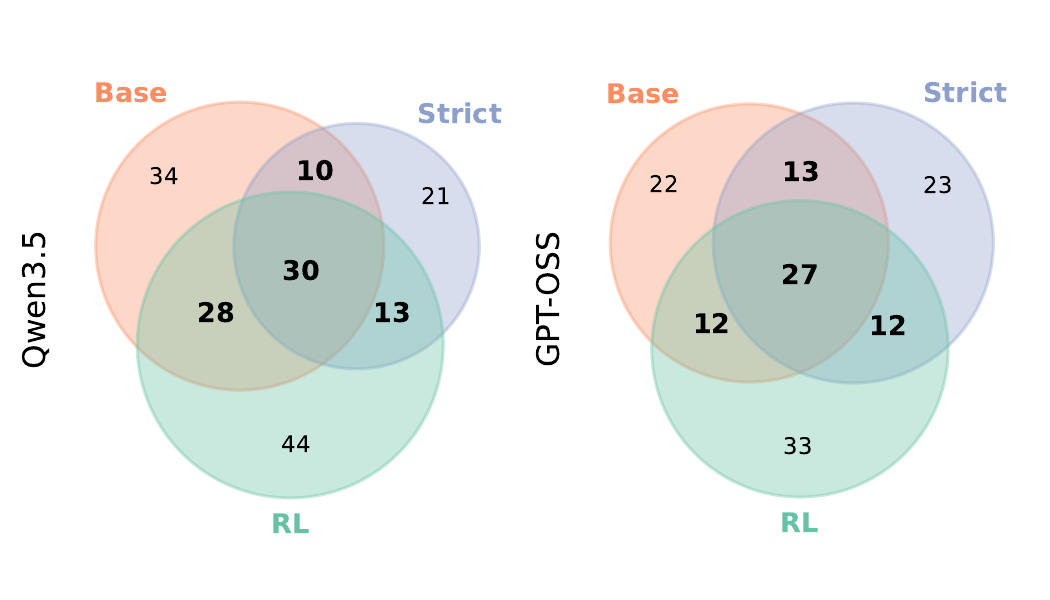}
\hspace{-4mm}
\includegraphics[width=0.41\linewidth]{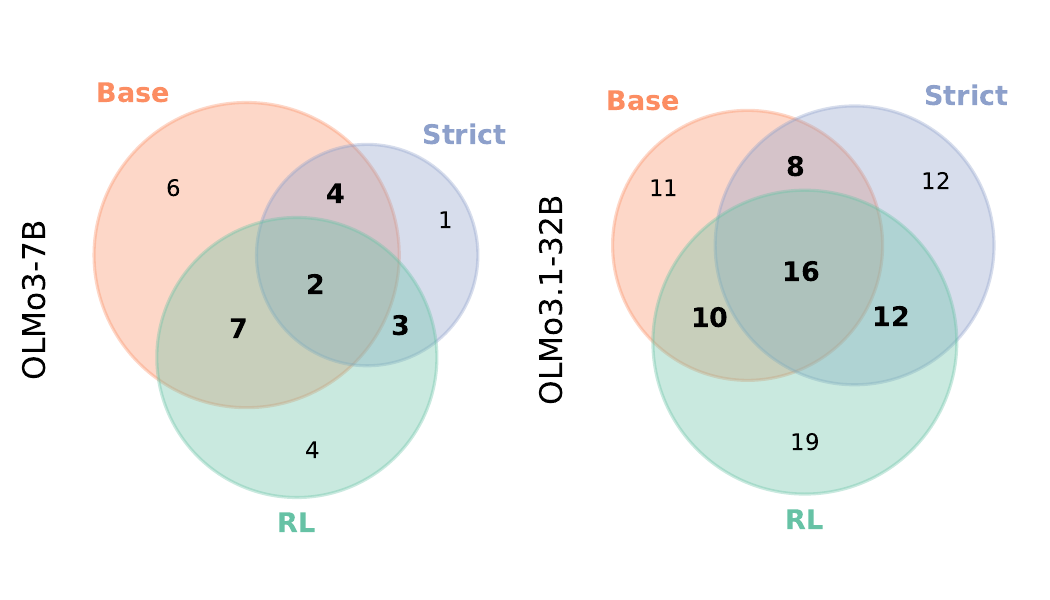}
\hspace{-3mm}
\begin{minipage}{0.19\linewidth}
\vspace{-3.5cm}
\includegraphics[width=\linewidth]{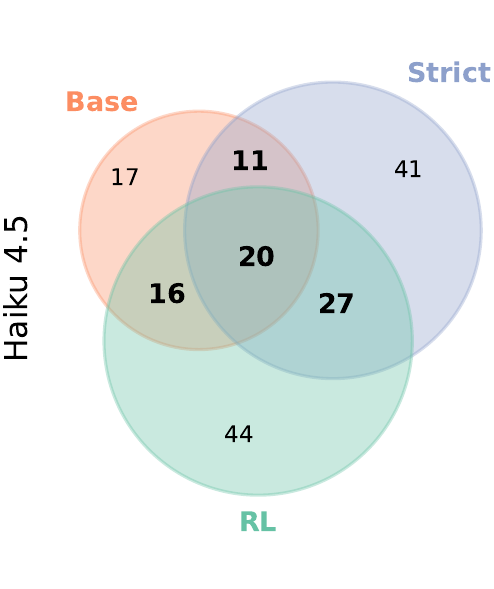}
\end{minipage}
\end{minipage}
\vspace{-6mm}
\caption{\UseMacro{FCap-verl-venn}}
\label{fig:verl-venn}
\end{figure*}

Table~\ref{tab:verl-lev-norm-eval-rust-repos} shows the mean \levmetric from the \reimplemented \before
functions to the \editor's output under four subsets of \tasks.
\textbf{All \tasks} covers every \task. Each \textbf{Both pass}
column is restricted to the \tasks on which the two \editors it names both produce
an edit that passes all tests. We leave the \editor not named out of the comparison and show only dashes there.
Figure~\ref{fig:verl-venn} visualizes those populations for
Table~\ref{tab:verl-lev-norm-eval-rust-repos}: each
restricted comparison is an overlap of two circles, and the
center region holds the \tasks all three solve.

We report \levmetric
at character granularity to show how much of the code is
edited. For a given \task, we min-max
normalize \baselabel, \strictlabel, and \rllabel within a pool of edits generated by all 7 models used in this work: the \baselabel, \strictlabel, \rllabel, \qwen,
\gpto, \olmol, and \haiku.
As in Section~\ref{sec:quantify-crossedit}, the normalization is defined on only some
\tasks, which leaves
\num{\UseMacro{fig-verl-comparison-allmean-rustfmt-minmax-qwen-lev-char-n}}
\tasks for \qwen,
\num{\UseMacro{fig-verl-comparison-allmean-rustfmt-minmax-gpt-lev-char-n}}
for \gpto,
\num{\UseMacro{fig-verl-comparison-allmean-rustfmt-minmax-olmo7b-lev-char-n}}
for \olmos,
\num{\UseMacro{fig-verl-comparison-allmean-rustfmt-minmax-olmo32b-lev-char-n}}
for \olmol, and
\num{\UseMacro{fig-verl-comparison-allmean-rustfmt-minmax-haiku-lev-char-n}}
for \haiku.
We report absolute distances and distances at line-level granularity in Appendix~\ref{sec:app-verl-lev-raw}.

\MyPara{Our framework meaningfully reduces the edit size of \olmos (Table~\ref{tab:verl-lev-norm-eval-rust-repos})}
On the All \tasks columns, \Tool halves the edit distance on
functions \reimplemented by each of the
\numofopeneditors open-weight \rewriters, demonstrating the effectiveness of our framework.
One natural concern is that the \rlmodel may edit less only because it
produces worse edits. The restricted columns rule this out. Where
\baselabel and \rllabel both succeed, the \rlmodel edits less
on every \rewriter. Where
\strictlabel and \rllabel both succeed, the \rlmodel
again edits less on \qwen, \gpto, \olmol, and \haiku \reimplementations. It performs worse only on \olmos, on a
population of \num{\UseMacro{fig-verl-lev-prompt-pass-rustfmt-minmax-olmo7b-lev-char-n}} \tasks. The
\rlmodel therefore makes smaller edits than both the \basemodel and the
\minprompt on \othercode while achieving the same successful outcome.

\MyPara{Prompt engineering cannot consistently lower edit size
(Table~\ref{tab:verl-lev-norm-eval-rust-repos})}
On the All \tasks columns, the \minprompt lowers the edit distance only on \qwen, \gpto, and \olmol \reimplementations, by far less
than the \rlmodel does, and actually raises it on \olmos and \haiku. On the
\basestrictpass column, where the \baselabel and \strictlabel edits
both pass every test, it edits less than the \basemodel only on \olmol. This demonstrates that \Tool's edit size reduction cannot be achieved by prompt engineering.

\subsection{Effect on Edit Success}
\label{sec:results-edit-success}

\begin{table}[t]
\centering
\small
\setlength{\tabcolsep}{3pt}
\begin{tabular}{l c c c}
\toprule
\shortstack[l]{\textbf{Implementor} \\ \quad\textbf{Editor}} & \UseMacro{TCol-build} & \UseMacro{TCol-alltest} & \UseMacro{TCol-meantest-wide} \\
\midrule
\multicolumn{4}{@{}l}{\textbf{\qwen}} \\
\quad \baselabel & \BiggerBold{\UseMacro{fig-verl-comparison-allmean-base-qwen-build}}{\UseMacro{fig-verl-comparison-allmean-verl-step62-qwen-build}} & \BiggerBold{\UseMacro{fig-verl-comparison-allmean-base-qwen-tests}}{\UseMacro{fig-verl-comparison-allmean-verl-step62-qwen-tests}} & \BiggerBold{\UseMacro{fig-verl-comparison-allmean-base-qwen-mean-tests}}{\UseMacro{fig-verl-comparison-allmean-verl-step62-qwen-mean-tests}} \\
\quad \strictlabel & \UseMacro{fig-verl-comparison-allmean-prompt-qwen-build} & \UseMacro{fig-verl-comparison-allmean-prompt-qwen-tests} & \UseMacro{fig-verl-comparison-allmean-prompt-qwen-mean-tests} \\
\quad \rllabel & \BiggerBold{\UseMacro{fig-verl-comparison-allmean-verl-step62-qwen-build}}{\UseMacro{fig-verl-comparison-allmean-base-qwen-build}} & \BiggerBold{\UseMacro{fig-verl-comparison-allmean-verl-step62-qwen-tests}}{\UseMacro{fig-verl-comparison-allmean-base-qwen-tests}} & \BiggerBold{\UseMacro{fig-verl-comparison-allmean-verl-step62-qwen-mean-tests}}{\UseMacro{fig-verl-comparison-allmean-base-qwen-mean-tests}} \\
\midrule
\multicolumn{4}{@{}l}{\textbf{\gpto}} \\
\quad \baselabel & \BiggerBold{\UseMacro{fig-verl-comparison-allmean-base-gpt-build}}{\UseMacro{fig-verl-comparison-allmean-verl-step62-gpt-build}} & \BiggerBold{\UseMacro{fig-verl-comparison-allmean-base-gpt-tests}}{\UseMacro{fig-verl-comparison-allmean-verl-step62-gpt-tests}} & \BiggerBold{\UseMacro{fig-verl-comparison-allmean-base-gpt-mean-tests}}{\UseMacro{fig-verl-comparison-allmean-verl-step62-gpt-mean-tests}} \\
\quad \strictlabel & \UseMacro{fig-verl-comparison-allmean-prompt-gpt-build} & \UseMacro{fig-verl-comparison-allmean-prompt-gpt-tests} & \UseMacro{fig-verl-comparison-allmean-prompt-gpt-mean-tests} \\
\quad \rllabel & \BiggerBold{\UseMacro{fig-verl-comparison-allmean-verl-step62-gpt-build}}{\UseMacro{fig-verl-comparison-allmean-base-gpt-build}} & \BiggerBold{\UseMacro{fig-verl-comparison-allmean-verl-step62-gpt-tests}}{\UseMacro{fig-verl-comparison-allmean-base-gpt-tests}} & \BiggerBold{\UseMacro{fig-verl-comparison-allmean-verl-step62-gpt-mean-tests}}{\UseMacro{fig-verl-comparison-allmean-base-gpt-mean-tests}} \\
\midrule
\multicolumn{4}{@{}l}{\textbf{\olmos}} \\
\quad \baselabel & \BiggerBold{\UseMacro{fig-verl-comparison-allmean-base-olmo7b-build}}{\UseMacro{fig-verl-comparison-allmean-verl-step62-olmo7b-build}} & \BiggerBold{\UseMacro{fig-verl-comparison-allmean-base-olmo7b-tests}}{\UseMacro{fig-verl-comparison-allmean-verl-step62-olmo7b-tests}} & \BiggerBold{\UseMacro{fig-verl-comparison-allmean-base-olmo7b-mean-tests}}{\UseMacro{fig-verl-comparison-allmean-verl-step62-olmo7b-mean-tests}} \\
\quad \strictlabel & \UseMacro{fig-verl-comparison-allmean-prompt-olmo7b-build} & \UseMacro{fig-verl-comparison-allmean-prompt-olmo7b-tests} & \UseMacro{fig-verl-comparison-allmean-prompt-olmo7b-mean-tests} \\
\quad \rllabel & \BiggerBold{\UseMacro{fig-verl-comparison-allmean-verl-step62-olmo7b-build}}{\UseMacro{fig-verl-comparison-allmean-base-olmo7b-build}} & \BiggerBold{\UseMacro{fig-verl-comparison-allmean-verl-step62-olmo7b-tests}}{\UseMacro{fig-verl-comparison-allmean-base-olmo7b-tests}} & \BiggerBold{\UseMacro{fig-verl-comparison-allmean-verl-step62-olmo7b-mean-tests}}{\UseMacro{fig-verl-comparison-allmean-base-olmo7b-mean-tests}} \\
\midrule
\multicolumn{4}{@{}l}{\textbf{\olmol}} \\
\quad \baselabel & \BiggerBold{\UseMacro{fig-verl-comparison-allmean-base-olmo32b-build}}{\UseMacro{fig-verl-comparison-allmean-verl-step62-olmo32b-build}} & \BiggerBold{\UseMacro{fig-verl-comparison-allmean-base-olmo32b-tests}}{\UseMacro{fig-verl-comparison-allmean-verl-step62-olmo32b-tests}} & \BiggerBold{\UseMacro{fig-verl-comparison-allmean-base-olmo32b-mean-tests}}{\UseMacro{fig-verl-comparison-allmean-verl-step62-olmo32b-mean-tests}} \\
\quad \strictlabel & \UseMacro{fig-verl-comparison-allmean-prompt-olmo32b-build} & \UseMacro{fig-verl-comparison-allmean-prompt-olmo32b-tests} & \UseMacro{fig-verl-comparison-allmean-prompt-olmo32b-mean-tests} \\
\quad \rllabel & \BiggerBold{\UseMacro{fig-verl-comparison-allmean-verl-step62-olmo32b-build}}{\UseMacro{fig-verl-comparison-allmean-base-olmo32b-build}} & \BiggerBold{\UseMacro{fig-verl-comparison-allmean-verl-step62-olmo32b-tests}}{\UseMacro{fig-verl-comparison-allmean-base-olmo32b-tests}} & \BiggerBold{\UseMacro{fig-verl-comparison-allmean-verl-step62-olmo32b-mean-tests}}{\UseMacro{fig-verl-comparison-allmean-base-olmo32b-mean-tests}} \\
\midrule
\multicolumn{4}{@{}l}{\textbf{\haiku}} \\
\quad \baselabel & \BiggerBold{\UseMacro{fig-verl-comparison-allmean-base-haiku-build}}{\UseMacro{fig-verl-comparison-allmean-verl-step62-haiku-build}} & \BiggerBold{\UseMacro{fig-verl-comparison-allmean-base-haiku-tests}}{\UseMacro{fig-verl-comparison-allmean-verl-step62-haiku-tests}} & \BiggerBold{\UseMacro{fig-verl-comparison-allmean-base-haiku-mean-tests}}{\UseMacro{fig-verl-comparison-allmean-verl-step62-haiku-mean-tests}} \\
\quad \strictlabel & \UseMacro{fig-verl-comparison-allmean-prompt-haiku-build} & \UseMacro{fig-verl-comparison-allmean-prompt-haiku-tests} & \UseMacro{fig-verl-comparison-allmean-prompt-haiku-mean-tests} \\
\quad \rllabel & \BiggerBold{\UseMacro{fig-verl-comparison-allmean-verl-step62-haiku-build}}{\UseMacro{fig-verl-comparison-allmean-base-haiku-build}} & \BiggerBold{\UseMacro{fig-verl-comparison-allmean-verl-step62-haiku-tests}}{\UseMacro{fig-verl-comparison-allmean-base-haiku-tests}} & \BiggerBold{\UseMacro{fig-verl-comparison-allmean-verl-step62-haiku-mean-tests}}{\UseMacro{fig-verl-comparison-allmean-base-haiku-mean-tests}} \\
\bottomrule
\end{tabular}
\caption{\UseMacro{TCap-verl-comparison-all-mean-eval-rust-repos}}
\label{tab:verl-comparison-all-mean-eval-rust-repos}
\end{table}

\MyPara{Smaller edits do not come at the cost of edit success (Table~\ref{tab:verl-comparison-all-mean-eval-rust-repos})}
We have shown that \Tool teaches the model to edit less. Does this
come at the cost of edit quality? We report the
percentage of edits that successfully build
(\UseMacro{TColInline-build}), the percentage that pass
every collected test
(\UseMacro{TColInline-alltest}), and the mean of the fraction of tests passed for each task
(\UseMacro{TColInline-meantest}). \rllabel outperforms \baselabel on all three measures for every \rewriter
other than \olmos itself.
This exception is expected. \Tool's training process teaches the \basemodel to
better edit \othercode and limit the excessive edits, so it is not expected to improve the task success rate on \selfediting, where the model already introduces a smaller number of excessive edits. The \rlmodel still keeps most of its performance in terms of the mean fraction of tests passed. In practice, when handling code written by \olmos itself, a user could unload the LoRA adapter and fall back to the
\basemodel for that \task. \rllabel also outperforms \strictlabel on
every \rewriter, showing that the success rate improvement cannot be achieved by prompt engineering alone. Here, we note that
success rates are low in every row, which is due to the editing capability of the small \basemodel, \olmos, rather than the training.

\begin{table}[t]
\centering
\small
\setlength{\tabcolsep}{3pt}
\begin{tabular}{l c c c}
\toprule
\shortstack[l]{\textbf{Implementor} \\ \quad\textbf{Editor}} & \UseMacro{TCol-build} & \UseMacro{TCol-build-nc} & \UseMacro{TCol-build-wc} \\
\midrule
\multicolumn{4}{@{}l}{\textbf{\qwen}} \\
\quad \baselabel & \LargestBoldSafe{\UseMacro{fig-verl-buildnc-base-qwen-build}}{\UseMacro{fig-verl-buildnc-prompt-qwen-build}}{\UseMacro{fig-verl-buildnc-verl-step62-qwen-build}} & \LargestBoldSafe{\UseMacro{fig-verl-buildnc-base-qwen-no-change}}{\UseMacro{fig-verl-buildnc-prompt-qwen-no-change}}{\UseMacro{fig-verl-buildnc-verl-step62-qwen-no-change}} & \LargestBoldSafe{\UseMacro{fig-verl-buildnc-base-qwen-build-change}}{\UseMacro{fig-verl-buildnc-prompt-qwen-build-change}}{\UseMacro{fig-verl-buildnc-verl-step62-qwen-build-change}} \\
\quad \strictlabel & \LargestBoldSafe{\UseMacro{fig-verl-buildnc-prompt-qwen-build}}{\UseMacro{fig-verl-buildnc-base-qwen-build}}{\UseMacro{fig-verl-buildnc-verl-step62-qwen-build}} & \LargestBoldSafe{\UseMacro{fig-verl-buildnc-prompt-qwen-no-change}}{\UseMacro{fig-verl-buildnc-base-qwen-no-change}}{\UseMacro{fig-verl-buildnc-verl-step62-qwen-no-change}} & \LargestBoldSafe{\UseMacro{fig-verl-buildnc-prompt-qwen-build-change}}{\UseMacro{fig-verl-buildnc-base-qwen-build-change}}{\UseMacro{fig-verl-buildnc-verl-step62-qwen-build-change}} \\
\quad \rllabel & \LargestBoldSafe{\UseMacro{fig-verl-buildnc-verl-step62-qwen-build}}{\UseMacro{fig-verl-buildnc-base-qwen-build}}{\UseMacro{fig-verl-buildnc-prompt-qwen-build}} & \LargestBoldSafe{\UseMacro{fig-verl-buildnc-verl-step62-qwen-no-change}}{\UseMacro{fig-verl-buildnc-base-qwen-no-change}}{\UseMacro{fig-verl-buildnc-prompt-qwen-no-change}} & \LargestBoldSafe{\UseMacro{fig-verl-buildnc-verl-step62-qwen-build-change}}{\UseMacro{fig-verl-buildnc-base-qwen-build-change}}{\UseMacro{fig-verl-buildnc-prompt-qwen-build-change}} \\
\midrule
\multicolumn{4}{@{}l}{\textbf{\gpto}} \\
\quad \baselabel & \LargestBoldSafe{\UseMacro{fig-verl-buildnc-base-gpt-build}}{\UseMacro{fig-verl-buildnc-prompt-gpt-build}}{\UseMacro{fig-verl-buildnc-verl-step62-gpt-build}} & \LargestBoldSafe{\UseMacro{fig-verl-buildnc-base-gpt-no-change}}{\UseMacro{fig-verl-buildnc-prompt-gpt-no-change}}{\UseMacro{fig-verl-buildnc-verl-step62-gpt-no-change}} & \LargestBoldSafe{\UseMacro{fig-verl-buildnc-base-gpt-build-change}}{\UseMacro{fig-verl-buildnc-prompt-gpt-build-change}}{\UseMacro{fig-verl-buildnc-verl-step62-gpt-build-change}} \\
\quad \strictlabel & \LargestBoldSafe{\UseMacro{fig-verl-buildnc-prompt-gpt-build}}{\UseMacro{fig-verl-buildnc-base-gpt-build}}{\UseMacro{fig-verl-buildnc-verl-step62-gpt-build}} & \LargestBoldSafe{\UseMacro{fig-verl-buildnc-prompt-gpt-no-change}}{\UseMacro{fig-verl-buildnc-base-gpt-no-change}}{\UseMacro{fig-verl-buildnc-verl-step62-gpt-no-change}} & \LargestBoldSafe{\UseMacro{fig-verl-buildnc-prompt-gpt-build-change}}{\UseMacro{fig-verl-buildnc-base-gpt-build-change}}{\UseMacro{fig-verl-buildnc-verl-step62-gpt-build-change}} \\
\quad \rllabel & \LargestBoldSafe{\UseMacro{fig-verl-buildnc-verl-step62-gpt-build}}{\UseMacro{fig-verl-buildnc-base-gpt-build}}{\UseMacro{fig-verl-buildnc-prompt-gpt-build}} & \LargestBoldSafe{\UseMacro{fig-verl-buildnc-verl-step62-gpt-no-change}}{\UseMacro{fig-verl-buildnc-base-gpt-no-change}}{\UseMacro{fig-verl-buildnc-prompt-gpt-no-change}} & \LargestBoldSafe{\UseMacro{fig-verl-buildnc-verl-step62-gpt-build-change}}{\UseMacro{fig-verl-buildnc-base-gpt-build-change}}{\UseMacro{fig-verl-buildnc-prompt-gpt-build-change}} \\
\midrule
\multicolumn{4}{@{}l}{\textbf{\olmos}} \\
\quad \baselabel & \LargestBoldSafe{\UseMacro{fig-verl-buildnc-base-olmo7b-build}}{\UseMacro{fig-verl-buildnc-prompt-olmo7b-build}}{\UseMacro{fig-verl-buildnc-verl-step62-olmo7b-build}} & \LargestBoldSafe{\UseMacro{fig-verl-buildnc-base-olmo7b-no-change}}{\UseMacro{fig-verl-buildnc-prompt-olmo7b-no-change}}{\UseMacro{fig-verl-buildnc-verl-step62-olmo7b-no-change}} & \LargestBoldSafe{\UseMacro{fig-verl-buildnc-base-olmo7b-build-change}}{\UseMacro{fig-verl-buildnc-prompt-olmo7b-build-change}}{\UseMacro{fig-verl-buildnc-verl-step62-olmo7b-build-change}} \\
\quad \strictlabel & \LargestBoldSafe{\UseMacro{fig-verl-buildnc-prompt-olmo7b-build}}{\UseMacro{fig-verl-buildnc-base-olmo7b-build}}{\UseMacro{fig-verl-buildnc-verl-step62-olmo7b-build}} & \LargestBoldSafe{\UseMacro{fig-verl-buildnc-prompt-olmo7b-no-change}}{\UseMacro{fig-verl-buildnc-base-olmo7b-no-change}}{\UseMacro{fig-verl-buildnc-verl-step62-olmo7b-no-change}} & \LargestBoldSafe{\UseMacro{fig-verl-buildnc-prompt-olmo7b-build-change}}{\UseMacro{fig-verl-buildnc-base-olmo7b-build-change}}{\UseMacro{fig-verl-buildnc-verl-step62-olmo7b-build-change}} \\
\quad \rllabel & \LargestBoldSafe{\UseMacro{fig-verl-buildnc-verl-step62-olmo7b-build}}{\UseMacro{fig-verl-buildnc-base-olmo7b-build}}{\UseMacro{fig-verl-buildnc-prompt-olmo7b-build}} & \LargestBoldSafe{\UseMacro{fig-verl-buildnc-verl-step62-olmo7b-no-change}}{\UseMacro{fig-verl-buildnc-base-olmo7b-no-change}}{\UseMacro{fig-verl-buildnc-prompt-olmo7b-no-change}} & \LargestBoldSafe{\UseMacro{fig-verl-buildnc-verl-step62-olmo7b-build-change}}{\UseMacro{fig-verl-buildnc-base-olmo7b-build-change}}{\UseMacro{fig-verl-buildnc-prompt-olmo7b-build-change}} \\
\midrule
\multicolumn{4}{@{}l}{\textbf{\olmol}} \\
\quad \baselabel & \LargestBoldSafe{\UseMacro{fig-verl-buildnc-base-olmo32b-build}}{\UseMacro{fig-verl-buildnc-prompt-olmo32b-build}}{\UseMacro{fig-verl-buildnc-verl-step62-olmo32b-build}} & \LargestBoldSafe{\UseMacro{fig-verl-buildnc-base-olmo32b-no-change}}{\UseMacro{fig-verl-buildnc-prompt-olmo32b-no-change}}{\UseMacro{fig-verl-buildnc-verl-step62-olmo32b-no-change}} & \LargestBoldSafe{\UseMacro{fig-verl-buildnc-base-olmo32b-build-change}}{\UseMacro{fig-verl-buildnc-prompt-olmo32b-build-change}}{\UseMacro{fig-verl-buildnc-verl-step62-olmo32b-build-change}} \\
\quad \strictlabel & \LargestBoldSafe{\UseMacro{fig-verl-buildnc-prompt-olmo32b-build}}{\UseMacro{fig-verl-buildnc-base-olmo32b-build}}{\UseMacro{fig-verl-buildnc-verl-step62-olmo32b-build}} & \LargestBoldSafe{\UseMacro{fig-verl-buildnc-prompt-olmo32b-no-change}}{\UseMacro{fig-verl-buildnc-base-olmo32b-no-change}}{\UseMacro{fig-verl-buildnc-verl-step62-olmo32b-no-change}} & \LargestBoldSafe{\UseMacro{fig-verl-buildnc-prompt-olmo32b-build-change}}{\UseMacro{fig-verl-buildnc-base-olmo32b-build-change}}{\UseMacro{fig-verl-buildnc-verl-step62-olmo32b-build-change}} \\
\quad \rllabel & \LargestBoldSafe{\UseMacro{fig-verl-buildnc-verl-step62-olmo32b-build}}{\UseMacro{fig-verl-buildnc-base-olmo32b-build}}{\UseMacro{fig-verl-buildnc-prompt-olmo32b-build}} & \LargestBoldSafe{\UseMacro{fig-verl-buildnc-verl-step62-olmo32b-no-change}}{\UseMacro{fig-verl-buildnc-base-olmo32b-no-change}}{\UseMacro{fig-verl-buildnc-prompt-olmo32b-no-change}} & \LargestBoldSafe{\UseMacro{fig-verl-buildnc-verl-step62-olmo32b-build-change}}{\UseMacro{fig-verl-buildnc-base-olmo32b-build-change}}{\UseMacro{fig-verl-buildnc-prompt-olmo32b-build-change}} \\
\midrule
\multicolumn{4}{@{}l}{\textbf{\haiku}} \\
\quad \baselabel & \LargestBoldSafe{\UseMacro{fig-verl-buildnc-base-haiku-build}}{\UseMacro{fig-verl-buildnc-prompt-haiku-build}}{\UseMacro{fig-verl-buildnc-verl-step62-haiku-build}} & \LargestBoldSafe{\UseMacro{fig-verl-buildnc-base-haiku-no-change}}{\UseMacro{fig-verl-buildnc-prompt-haiku-no-change}}{\UseMacro{fig-verl-buildnc-verl-step62-haiku-no-change}} & \LargestBoldSafe{\UseMacro{fig-verl-buildnc-base-haiku-build-change}}{\UseMacro{fig-verl-buildnc-prompt-haiku-build-change}}{\UseMacro{fig-verl-buildnc-verl-step62-haiku-build-change}} \\
\quad \strictlabel & \LargestBoldSafe{\UseMacro{fig-verl-buildnc-prompt-haiku-build}}{\UseMacro{fig-verl-buildnc-base-haiku-build}}{\UseMacro{fig-verl-buildnc-verl-step62-haiku-build}} & \LargestBoldSafe{\UseMacro{fig-verl-buildnc-prompt-haiku-no-change}}{\UseMacro{fig-verl-buildnc-base-haiku-no-change}}{\UseMacro{fig-verl-buildnc-verl-step62-haiku-no-change}} & \LargestBoldSafe{\UseMacro{fig-verl-buildnc-prompt-haiku-build-change}}{\UseMacro{fig-verl-buildnc-base-haiku-build-change}}{\UseMacro{fig-verl-buildnc-verl-step62-haiku-build-change}} \\
\quad \rllabel & \LargestBoldSafe{\UseMacro{fig-verl-buildnc-verl-step62-haiku-build}}{\UseMacro{fig-verl-buildnc-base-haiku-build}}{\UseMacro{fig-verl-buildnc-prompt-haiku-build}} & \LargestBoldSafe{\UseMacro{fig-verl-buildnc-verl-step62-haiku-no-change}}{\UseMacro{fig-verl-buildnc-base-haiku-no-change}}{\UseMacro{fig-verl-buildnc-prompt-haiku-no-change}} & \LargestBoldSafe{\UseMacro{fig-verl-buildnc-verl-step62-haiku-build-change}}{\UseMacro{fig-verl-buildnc-base-haiku-build-change}}{\UseMacro{fig-verl-buildnc-prompt-haiku-build-change}} \\
\bottomrule
\end{tabular}
\caption{\UseMacro{TCap-verl-buildnc-eval-rust-repos}}
\label{tab:verl-buildnc-eval-rust-repos}
\end{table}

\MyPara{The build advantage holds after excluding trivial no-change responses (Table~\ref{tab:verl-buildnc-eval-rust-repos})}
However, build metrics can be inflated if the model returns the
\reimplementation unchanged, since an unmodified \before
function may already build trivially. To isolate this effect, we split the \tasks into two disjoint subsets: \tasks
where the \rlmodel makes no change, and \tasks where the \rlmodel
actually modifies the \reimplementation. A noticeable portion (\UseMacro{fig-verl-buildnc-pooled-nochange-share}\%) of the \rlmodel's outputs falls in the no-change category. This is within reason: $R_{\text{sim}}$ rewards minimal change, so the
\rlmodel becomes more conservative and more willing to leave the \reimplementation as-is when it does not know how to complete the edit request. After excluding these no-change responses and restricting to \tasks where the \rlmodel's outputs actually modify the \reimplementation, the \rlmodel still achieves a higher build rate than the \basemodel for every \rewriter except \olmos itself, showing that its build advantage does not come solely from declining to edit.
We also show this for the mean fraction of tests passed
(\UseMacro{TColInline-meantest}) in Appendix~\ref{sec:app-verl-meantestnc}.

\subsection{Qualitative Analysis}
\label{sec:results-qualitative}

To understand how the \rlmodel's edits differ from the \basemodel's,
we inspected edits for two sets of \tasks: \tasks where only the \rlmodel
generates successful edits, and \tasks where only the \basemodel generates
successful edits. Figure~\ref{fig:scopedrift-endianness} and
Figure~\ref{fig:rename-cookedbytestring} in
Appendix~\ref{sec:app-qualitative} show one example of each.

\MyPara{The \rlmodel keeps edits in scope}
We see that in cases where the \rlmodel generates successful edits but the
\basemodel fails, the \basemodel tends to add code outside the given
edit scope, such as new \CodeIn{struct} definitions or unrelated helper functions
that the \PR did not request, while the \rlmodel learns to keep its
edit in scope. For example, in the function \texttt{endianness} the \task is to add
a new \CodeIn{match} arm to handle a new type of architecture. The
\basemodel adds a hallucinated \texttt{enum Architecture} with invalid
syntax, causing a build failure. On the other hand, the \rlmodel limits its
editing scope to the \CodeIn{match} block and adds only the
required arm.

\MyPara{The \simreward penalty can make the \rlmodel under-edit on identifier propagation}
The downside of the training shows up where the \basemodel
passes but the \rlmodel fails. We observe that the penalty from \simreward
makes the \rlmodel more likely to refuse to make requested edits if the \task
seems too minor. This is especially prevalent when an identifier
is renamed and the rename needs to be propagated. The \rlmodel often fails
to apply the needed propagation. In \texttt{cooked\_byte\_string}, the \task is to propagate changes from \texttt{LexError} to
\texttt{Reject}. The \basemodel completes that \task by making the simple
replacement, while the \rlmodel keeps the old return type and additionally
adds a new character to skip for \CodeIn{next\_byte}.

\section{Conclusion}

We show that \LLMs edit code authored by other \LLMs excessively.
We build a \rust pull-request corpus that allows varying the \rewriter of the \beforecode, and find that
the \numofopeneditors open-weight \LLMs we study edit \othercode more than their
own \selfcode on
\UseMacro{fig-pr-lev-selfcross-mwu-open-n-right-direction} of
\UseMacro{fig-pr-lev-selfcross-mwu-open-n-tests} model pairings. To address this excessive editing, we
introduce \Tool, a post-training framework that pairs a \simreward
penalizing large edits with an \execreward scoring edit
success, optimized via GRPO. Training \olmos with \Tool roughly halves
its edit distance on \reimplementations by every \rewriter while
improving build and all-tests pass rates on every \rewriter but
\olmos itself. We show that this improvement cannot be achieved by prompt engineering. In fact, the prompt engineering approach fails to reduce edit distance or improve edit success consistently.

\section{Limitations}

\MyPara{Single programming language}
Our corpus and training data are drawn
exclusively from Rust crates on \github. We acknowledge that languages
with weaker typing or different grammar features, such as Python or
JavaScript, may exhibit different \crossediting patterns. We have not
verified that the \selfedit/\crossedit gap, or the
effectiveness of \Tool's reward can generalize to other programming languages.

\MyPara{Function-level scope}
In order to limit the edit scope and the context size, we decompose
each \PR into per-function edit \tasks where both the \before and
\after bodies contain between 20 and 200 non-empty, non-comment lines.
Edits that, in practice, could span multiple functions or multiple
files are outside the scope of our project. \Tool's behavior in those
settings is not studied in this work.

\MyPara{Limited model pool}
Due to budget constraints, the study uses \numofeditors \LLMs (\qwen,
\gpto, \olmol, \olmos, \haiku), only one of which is closed-source,
and we train only \olmos with \Tool. Larger closed-source models such
as GPT-5 and Claude Opus are not evaluated. The training framework are
also not tested on these large closed-source models.

\section*{Acknowledgments}
We thank the anonymous reviewers
for helpful feedback. This work was supported in part by the U.S. National Science
Foundation (NSF) Nos. CCF-2217696, CCF-2313027, CCF-2403036; Cisco; and AMD (University Program AI \& HPC Cluster). Any opinions, findings, and conclusions or recommendations expressed in this material
are those of the authors and do not necessarily reflect the views of the sponsoring entities.

\bibliography{bib}

@inproceedings{Cassano2023CanItEdit,
  title     = {Can It Edit? Evaluating the Ability of Large Language Models to Follow Code Editing Instructions},
  author    = {Federico Cassano and Luisa Li and Akul Sethi and Noah Shinn and Abby Brennan-Jones and Jacob Ginesin and Edward Berman and George Chakhnashvili and Anton Lozhkov and Carolyn Jane Anderson and Arjun Guha},
  booktitle = {Conference on Language Modeling (COLM)},
  year      = {2024},
  url       = {https://openreview.net/forum?id=D06yk3DBas},
}

@inproceedings{chi2026editbench,
  title     = {{EDIT}-Bench: Evaluating {LLM} Abilities to Perform Real-World Instructed Code Edits},
  author    = {Wayne Chi and Valerie Chen and Ryan Shar and Aditya Mittal and Jenny Liang and Wei-Lin Chiang and Anastasios Nikolas Angelopoulos and Ion Stoica and Graham Neubig and Ameet Talwalkar and Chris Donahue},
  booktitle = {International Conference on Learning Representations (ICLR)},
  year      = {2026},
  url       = {https://openreview.net/forum?id=FtL9eEmU6v},
}

@article{xiaAgentlessDemystifyingLLMbased2024,
  title   = {Demystifying {LLM}-Based Software Engineering Agents},
  author  = {Xia, Chunqiu Steven and Deng, Yinlin and Dunn, Soren and Zhang, Lingming},
  journal = {Proceedings of the ACM on Software Engineering (PACMSE)},
  volume  = {2},
  number  = {FSE},
  pages   = {801--824},
  year    = {2025},
  doi     = {10.1145/3715754},
}

@article{chen2021humaneval,
  title  = {Evaluating Large Language Models Trained on Code},
  author = {     Mark Chen and
                  Jerry Tworek and
                  Heewoo Jun and
                  Qiming Yuan and
                  Henrique Pond{\'{e}} de Oliveira Pinto and
                  Jared Kaplan and
                  Harri Edwards and
                  Yuri Burda and
                  Nicholas Joseph and
                  Greg Brockman and
                  Alex Ray and
                  Raul Puri and
                  Gretchen Krueger and
                  Michael Petrov and
                  Heidy Khlaaf and
                  Girish Sastry and
                  Pamela Mishkin and
                  Brooke Chan and
                  Scott Gray and
                  Nick Ryder and
                  Mikhail Pavlov and
                  Alethea Power and
                  Lukasz Kaiser and
                  Mohammad Bavarian and
                  Clemens Winter and
                  Philippe Tillet and
                  Felipe Petroski Such and
                  Dave Cummings and
                  Matthias Plappert and
                  Fotios Chantzis and
                  Elizabeth Barnes and
                  Ariel Herbert{-}Voss and
                  William Hebgen Guss and
                  Alex Nichol and
                  Alex Paino and
                  Nikolas Tezak and
                  Jie Tang and
                  Igor Babuschkin and
                  Suchir Balaji and
                  Shantanu Jain and
                  William Saunders and
                  Christopher Hesse and
                  Andrew N. Carr and
                  Jan Leike and
                  Joshua Achiam and
                  Vedant Misra and
                  Evan Morikawa and
                  Alec Radford and
                  Matthew Knight and
                  Miles Brundage and
                  Mira Murati and
                  Katie Mayer and
                  Peter Welinder and
                  Bob McGrew and
                  Dario Amodei and
                  Sam McCandlish and
                  Ilya Sutskever and
                  Wojciech Zaremba
  },
  journal= {arXiv preprint arXiv:2107.03374},
  year   = {2021},
  doi = {10.48550/arXiv.2107.03374}
}

@article{austin2021mbpp,
  title  = {Program Synthesis with Large Language Models},
  author = {Austin, Jacob and Odena, Augustus and Nye, Maxwell and Bosma, Maarten and Michalewski, Henryk and Dohan, David and Jiang, Ellen and Cai, Carrie and Terry, Michael and Le, Quoc and Sutton, Charles},
  journal= {arXiv preprint arXiv:2108.07732},
  year   = {2021},
  doi = {10.48550/arXiv.2108.07732}
}

@inproceedings{nijkamp2022codegen,
  title     = {CodeGen: An Open Large Language Model for Code with Multi-Turn Program Synthesis},
  author    = {Nijkamp, Erik and Pang, Bo and Hayashi, Hiroaki and Tu, Lifu and Wang, Huan and Zhou, Yingbo and Savarese, Silvio and Xiong, Caiming},
  booktitle = {International Conference on Learning Representations (ICLR)},
  year      = {2023},
  url       = {https://openreview.net/forum?id=iaYcJKpY2B_}
}

@article{roziere2023codellama,
  title  = {Code Llama: Open Foundation Models for Code},
  author = {Baptiste Rozi{\`{e}}re and
                  Jonas Gehring and
                  Fabian Gloeckle and
                  Sten Sootla and
                  Itai Gat and
                  Xiaoqing Ellen Tan and
                  Yossi Adi and
                  Jingyu Liu and
                  Romain Sauvestre and
                  Tal Remez and
                  J{\'{e}}r{\'{e}}my Rapin and
                  Artyom Kozhevnikov and
                  Ivan Evtimov and
                  Joanna Bitton and
                  Manish Bhatt and
                  Cristian Canton{-}Ferrer and
                  Aaron Grattafiori and
                  Wenhan Xiong and
                  Alexandre D{\'{e}}fossez and
                  Jade Copet and
                  Faisal Azhar and
                  Hugo Touvron and
                  Louis Martin and
                  Nicolas Usunier and
                  Thomas Scialom and
                  Gabriel Synnaeve
                  },
  journal= {arXiv preprint arXiv:2308.12950},
  year   = {2023},
  doi = {10.48550/arXiv.2308.12950}
}

@article{li2026gr3,
  title   = {Tackling Length Inflation Without Trade-offs: Group Relative Reward Rescaling for Reinforcement Learning},
  author  = {Li, Zichao and Lou, Jie and Dong, Fangchen and Fan, Zhiyuan and Ren, Mengjie and Lin, Hongyu and Han, Xianpei and Zhang, Debing and Sun, Le and Lu, Yaojie and Yu, Xing},
  journal = {arXiv preprint arXiv:2603.10535},
  year    = {2026},
  doi     = {10.48550/arXiv.2603.10535}
}

@article{hui2024qwen25coder,
  title  = {Qwen2.5-{C}oder Technical Report},
  author = {Binyuan Hui and
                  Jian Yang and
                  Zeyu Cui and
                  Jiaxi Yang and
                  Dayiheng Liu and
                  Lei Zhang and
                  Tianyu Liu and
                  Jiajun Zhang and
                  Bowen Yu and
                  Kai Dang and
                  An Yang and
                  Rui Men and
                  Fei Huang and
                  Xingzhang Ren and
                  Xuancheng Ren and
                  Jingren Zhou and
                  Junyang Lin},
  journal= {arXiv preprint arXiv:2409.12186},
  year   = {2024},
  doi = {10.48550/arXiv.2409.12186}
}

@article{yang2025qwen3,
  title  = {Qwen3 Technical Report},
  author = {An Yang and Anfeng Li and Baosong Yang and Beichen Zhang and Binyuan Hui and Bo Zheng and Bowen Yu and Chang Gao and Chengen Huang and Chenxu Lv and Chujie Zheng and Dayiheng Liu and Fan Zhou and Fei Huang and Feng Hu and Hao Ge and Haoran Wei and Huan Lin and Jialong Tang and Jian Yang and Jianhong Tu and Jianwei Zhang and Jianxin Yang and Jiaxi Yang and Jing Zhou and Jingren Zhou and Junyang Lin and Kai Dang and Keqin Bao and Kexin Yang and Le Yu and Lianghao Deng and Mei Li and Mingfeng Xue and Mingze Li and Pei Zhang and Peng Wang and Qin Zhu and Rui Men and Ruize Gao and Shixuan Liu and Shuang Luo and Tianhao Li and Tianyi Tang and Wenbiao Yin and Xingzhang Ren and Xinyu Wang and Xinyu Zhang and Xuancheng Ren and Yang Fan and Yang Su and Yichang Zhang and Yinger Zhang and Yu Wan and Yuqiong Liu and Zekun Wang and Zeyu Cui and Zhenru Zhang and Zhipeng Zhou and Zihan Qiu},
  journal= {arXiv preprint arXiv:2505.09388},
  year   = {2025},
  doi = {10.48550/arXiv.2505.09388}
}

@article{openai2025gptoss,
  title  = {gpt-oss-120b \& gpt-oss-20b Model Card},
  author = {{OpenAI}},
  journal= {arXiv preprint arXiv:2508.10925},
  year   = {2025},
  doi = {10.48550/arXiv.2508.10925}
}

@article{olmoteam2025olmo3,
  title  = {{Olmo} 3},
  author = {Allyson Ettinger and Amanda Bertsch and Bailey Kuehl and David Graham and David Heineman and Dirk Groeneveld and Faeze Brahman and Finbarr Timbers and Hamish Ivison and Jacob Morrison and Jake Poznanski and Kyle Lo and Luca Soldaini and Matt Jordan and Mayee Chen and Michael Noukhovitch and Nathan Lambert and Pete Walsh and Pradeep Dasigi and Robert Berry and Saumya Malik and Saurabh Shah and Scott Geng and Shane Arora and Shashank Gupta and Taira Anderson and Teng Xiao and Tyler Murray and Tyler Romero and Victoria Graf and Akari Asai and Akshita Bhagia and Alexander Wettig and Alisa Liu and Aman Rangapur and Chloe Anastasiades and Costa Huang and Dustin Schwenk and Harsh Trivedi and Ian Magnusson and Jaron Lochner and Jiacheng Liu and Lester James V. Miranda and Maarten Sap and Malia Morgan and Michael Schmitz and Michal Guerquin and Michael Wilson and Regan Huff and Ronan Le Bras and Rui Xin and Rulin Shao and Sam Skjonsberg and Shannon Zejiang Shen and Shuyue Stella Li and Tucker Wilde and Valentina Pyatkin and Will Merrill and Yapei Chang and Yuling Gu and Zhiyuan Zeng and Ashish Sabharwal and Luke Zettlemoyer and Pang Wei Koh and Ali Farhadi and Noah A. Smith and Hannaneh Hajishirzi},
  journal= {arXiv preprint arXiv:2512.13961},
  year   = {2025},
  doi = {10.48550/arXiv.2512.13961}
}

@misc{anthropic2025haiku,
  title        = {{Claude Haiku 4.5} System Card},
  author       = {{Anthropic}},
  year         = {2025},
  howpublished = {\url{https://www.anthropic.com/claude-haiku-4-5-system-card}}
}

@inproceedings{ouyang2022instructgpt,
  title  = {Training Language Models to Follow Instructions with Human Feedback},
  author = {Ouyang, Long and Wu, Jeffrey and Jiang, Xu and Almeida, Diogo and Wainwright, Carroll and Mishkin, Pamela and Zhang, Chong and Agarwal, Sandhini and Slama, Katarina and Ray, Alex and Schulman, John and Hilton, Jacob and Kelton, Fraser and Miller, Luke and Simens, Maddie and Askell, Amanda and Welinder, Peter and Christiano, Paul F and Leike, Jan and Lowe, Ryan},
  booktitle = {Advances in Neural Information Processing Systems (NeurIPS)},
  year   = {2022},
  url = {https://papers.nips.cc/paper_files/paper/2022/hash/b1efde53be364a73914f58805a001731-Abstract-Conference.html}
}

@article{shao2024deepseekmath,
  title  = {DeepSeekMath: Pushing the Limits of Mathematical Reasoning in Open Language Models},
  author = {Shao, Zhihong and Wang, Peiyi and Zhu, Qihao and Xu, Runxin and Song, Junxiao and Bi, Xiao and Zhang, Haowei and Zhang, Mingchuan and Li, Y. K. and Wu, Y. and Guo, Daya},
  journal= {arXiv preprint arXiv:2402.03300},
  year   = {2024},
  doi = {10.48550/arXiv.2402.03300}
}

@inproceedings{hu2022lora,
  title     = {{LoRA}: Low-Rank Adaptation of Large Language Models},
  author    = {Hu, Edward J. and Shen, Yelong and Wallis, Phillip and Allen-Zhu, Zeyuan and Li, Yuanzhi and Wang, Shean and Wang, Lu and Chen, Weizhu},
  booktitle = {International Conference on Learning Representations (ICLR)},
  year      = {2022},
  url = {https://openreview.net/forum?id=nZeVKeeFYf9}
}

@inproceedings{sheng2025verl,
  title     = {{HybridFlow}: A Flexible and Efficient {RLHF} Framework},
  author    = {Sheng, Guangming and Zhang, Chi and Ye, Zilingfeng and Wu, Xibin and Zhang, Wang and Zhang, Ru and Peng, Yanghua and Lin, Haibin and Wu, Chuan},
  booktitle = {European Conference on Computer Systems (EuroSys)},
  year      = {2025},
  pages     = {1279--1297},
  doi       = {10.1145/3689031.3696075}
}

@inproceedings{just2014defects4j,
  title  = {Defects4J: A Database of Existing Faults to Enable Controlled Testing Studies for Java Programs},
  author = {Just, Ren{\'e} and Jalali, Darioush and Ernst, Michael D.},
  booktitle = {International Symposium on Software Testing and Analysis (ISSTA)},
  year   = {2014},
  pages  = {437--440},
  doi    = {10.1145/2610384.2628055}
}

@inproceedings{widyasari2020bugsinpy,
  title  = {BugsInPy: A Database of Existing Bugs in Python Programs to Enable Controlled Testing and Debugging Studies},
  author = {Widyasari, Ratnadira and Sim, Sheng Qin and Lok, Camellia and Qi, Haodi and Phan, Jack and Tay, Qijin and Tan, Constance and Wee, Fiona and Tan, Jodie Ethelda and Yieh, Yuheng and Goh, Brian and Thung, Ferdian and Kang, Hong Jin and Hoang, Thong and Lo, David and Ouh, Eng Lieh},
  booktitle = {Joint European Software Engineering Conference and Symposium on the Foundations of Software Engineering (ESEC/FSE)},
  year   = {2020},
  pages  = {1556--1560},
  doi    = {10.1145/3368089.3417943}
}

@inproceedings{xia2022alpharepair,
  title  = {Less Training, More Repairing Please: Revisiting Automated Program Repair via Zero-Shot Learning},
  author = {Xia, Chunqiu Steven and Zhang, Lingming},
  booktitle = {Joint European Software Engineering Conference and Symposium on the Foundations of Software Engineering (ESEC/FSE)},
  year   = {2022},
  pages  = {959--971},
  doi    = {10.1145/3540250.3549101}
}

@inproceedings{xia2023apr,
  title  = {Automated Program Repair in the Era of Large Pre-Trained Language Models},
  author = {Xia, Chunqiu Steven and Wei, Yuxiang and Zhang, Lingming},
  booktitle = {International Conference on Software Engineering (ICSE)},
  year   = {2023},
  pages  = {1482--1494},
  doi    = {10.1109/ICSE48619.2023.00129}
}

@article{tufano2019empirical,
  title   = {An Empirical Study on Learning Bug-Fixing Patches in the Wild via Neural Machine Translation},
  author  = {Tufano, Michele and Watson, Cody and Bavota, Gabriele and Di Penta, Massimiliano and White, Martin and Poshyvanyk, Denys},
  journal = {ACM Transactions on Software Engineering and Methodology (TOSEM)},
  volume  = {28},
  number  = {4},
  pages   = {19:1--19:29},
  year    = {2019},
  doi     = {10.1145/3340544}
}

@inproceedings{cui2025refactorbench,
  title  = {RefactorBench: Evaluating Stateful Reasoning in Language Agents Through Code},
  author = {Gautam, Dhruv and Garg, Spandan and Jang, Jinu and Sundaresan, Neel and Moghaddam, Roshanak Zilouchian},
  booktitle = {International Conference on Learning Representations (ICLR)},
  year   = {2025},
  url    = {https://openreview.net/forum?id=NiNIthntx7}
}

@inproceedings{yang2024sweagent,
  title  = {SWE-agent: Agent-Computer Interfaces Enable Automated Software Engineering},
  author = {Yang, John and Jimenez, Carlos E. and Wettig, Alexander and Lieret, Kilian and Yao, Shunyu and Narasimhan, Karthik and Press, Ofir},
  booktitle = {Advances in Neural Information Processing Systems (NeurIPS)},
  year   = {2024},
  url    = {https://papers.nips.cc/paper_files/paper/2024/hash/5a7c947568c1b1328ccc5230172e1e7c-Abstract-Conference.html}
}

@inproceedings{liang2024survey,
  title  = {A Large-Scale Survey on the Usability of {AI} Programming Assistants: Successes and Challenges},
  author = {Liang, Jenny T. and Yang, Chenyang and Myers, Brad A.},
  booktitle = {International Conference on Software Engineering (ICSE)},
  year   = {2024},
  pages  = {52:1--52:13},
  doi    = {10.1145/3597503.3608128}
}

@inproceedings{cai2026oneisnotenough,
  title={One Is Not Enough: How People Use Multiple AI Models in Everyday Life},
  author={Pyo, Seunghwa and Lee, Donggun and Rhee, Jungwoo and Park, Soobin and Lim, Youn-kyung},
  booktitle={CHI Conference on Human Factors in Computing Systems (CHI EA)},
  pages={494:1--494:6},
  year={2026},
  doi={10.1145/3772363.3798682}
}

@misc{crates-io,
  author       = {{The Rust Foundation}},
  title        = {crates.io},
  year         = {2026},
  howpublished = {\url{https://crates.io}},
}

@inproceedings{ZhangETAL22CoditT5,
  author    = {Zhang, Jiyang and Panthaplackel, Sheena and Nie, Pengyu and Li, Junyi Jessy and Gligoric, Milos},
  title     = {Codit{T}5: Pretraining for Source Code and Natural Language Editing},
  booktitle = {International Conference on Automated Software Engineering (ASE)},
  year      = {2022},
  pages     = {22:1--22:12},
  doi       = {10.1145/3551349.3556955},
}

@inproceedings{le2022coderl,
  title     = {{CodeRL}: Mastering Code Generation through Pretrained Models and Deep Reinforcement Learning},
  author    = {Le, Hung and Wang, Yue and Gotmare, Akhilesh Deepak and Savarese, Silvio and Hoi, Steven C. H.},
  booktitle = {Advances in Neural Information Processing Systems (NeurIPS)},
  year      = {2022},
  url       = {https://papers.nips.cc/paper_files/paper/2022/hash/8636419dea1aa9fbd25fc4248e702da4-Abstract-Conference.html},
}

@inproceedings{madaan2023selfrefine,
    title = "Self-Refine: Iterative Refinement with Self-Feedback",
    author = "Madaan, Aman and
      Tandon, Niket and
      Gupta, Prakhar and
      Hallinan, Skyler and
      Gao, Luyu and
      Wiegreffe, Sarah and
      Alon, Uri and
      Dziri, Nouha and
      Prabhumoye, Shrimai and
      Yang, Yiming and
      Gupta, Shashank and
      Majumder, Bodhisattwa Prasad and
      Hermann, Katherine and
      Welleck, Sean and
      Yazdanbakhsh, Amir and
      Clark, Peter",
    booktitle = "Advances in Neural Information Processing Systems (NeurIPS)",
    year = "2023",
    url = "https://papers.nips.cc/paper_files/paper/2023/hash/91edff07232fb1b55a505a9e9f6c0ff3-Abstract-Conference.html",
}

@article{liu2023rltf,
      title={{RLTF}: Reinforcement Learning from Unit Test Feedback},
      author={Jiate Liu and Yiqin Zhu and Kaiwen Xiao and Qiang Fu and Xiao Han and Yang Wei and Deheng Ye},
      journal={Transactions on Machine Learning Research (TMLR)},
      issn={2835-8856},
      year={2023},
      url={https://openreview.net/forum?id=hjYmsV6nXZ}
}

@inproceedings{dou2024stepcoder,
    title = {{S}tep{C}oder: Improving Code Generation with Reinforcement Learning from Compiler Feedback},
    author = {Dou, Shihan  and
      Liu, Yan  and
      Jia, Haoxiang  and
      Zhou, Enyu  and
      Xiong, Limao  and
      Shan, Junjie  and
      Huang, Caishuang  and
      Wang, Xiao  and
      Fan, Xiaoran  and
      Xi, Zhiheng  and
      Zhou, Yuhao  and
      Ji, Tao  and
      Zheng, Rui  and
      Zhang, Qi  and
      Gui, Tao  and
      Huang, Xuanjing},
    booktitle = {Annual Meeting of the Association for Computational Linguistics (ACL)},
    year = {2024},
    pages = {4571--4585},
    doi = {10.18653/v1/2024.acl-long.251},
}

\newpage
\clearpage

\appendix
\raggedbottom

\section*{Appendix}

\section{Corpus Filter}
\label{sec:app-corpus-attrition}
Table~\ref{tab:corpus-attrition-before} and Table~\ref{tab:corpus-attrition-after} count how many functions survive each filter of Section~\ref{sec:char-crossediting}, from the \PRs we collect to the \tasks each \rewriter contributes.

\MyPara{Developer-written functions
(Table~\ref{tab:corpus-attrition-before})} In this table, \emph{\PR
and function-level filters} is the starting count, the
\PRs must change at most five files with at least 75\% of them in
Rust. Each function must already exist before the \PR and must not be
completely removed after the \PR. Finally, the functions must have
between 20 and 200 non-empty, non-comment lines. \emph{README
available} keeps the functions whose repository has a README.
\emph{Tests available} keeps the functions for which the line-coverage
collector finds tests that execute them. \emph{\IDCallsites} available
keeps functions that we can find their call cites inside their own
repository. \emph{\IDFuncdesc available} keeps functions whose code
and associated context fit inside the
\num{\UseMacro{corpus-eval-rust-repos-funcdesc-ctx}}-token window we
used to generated the descriptions.

\begin{table}[h]
\centering
\small
\setlength{\tabcolsep}{3pt}
\begin{tabular}{@{}lc@{}}
\toprule
\textbf{Filter} & \textbf{Functions} \\
\midrule
\PR and function-level filters & \num{\UseMacro{corpus-eval-rust-repos-split-functions}} \\
README available & \num{\UseMacro{corpus-eval-rust-repos-readme-functions}} \\
Tests available & \num{\UseMacro{corpus-eval-rust-repos-pretests-functions}} \\
\IDCallsites available & \num{\UseMacro{corpus-eval-rust-repos-target-after-callsites}} \\
\IDFuncdesc available & \num{\UseMacro{corpus-eval-rust-repos-draft-target-functions}} \\
\bottomrule
\end{tabular}
\caption{\UseMacro{TCap-corpus-attrition-before}}
\label{tab:corpus-attrition-before}
\end{table}

\MyPara{Model
\reimplementations (Table~\ref{tab:corpus-attrition-after})} These
filters come from Section~\ref{sec:reimplement}, and each \rewriter
now has its own count.
\emph{Pass \reimplementation test} keeps the \LLM \reimplementations
that pass every test collected for the \before function after the
\draft and its repair rounds. \emph{Overlap below 50\%} keeps
\reimplementations whose overlap with the \before function is at most
50\%, to prevent models from generating functions memorized during
training and to prevent style leakage. \emph{Fail edit test} keeps
only \reimplementations that fail at least one \after test, since an
\reimplementation that already passes them needs no edit.
\emph{\IDPRdesc available} keeps \tasks whose \PR diff and context fit
the window we use to generate the \PRdesc. The bold row is the final
set of edit \tasks. The filters that min-max normalization needs,
which keep only \tasks where every \editor produced an edit and
edits by each models were not all the same size, are not shown here. They are only applied to
normalized \levmetric and the resulting number is different for
different sets of \editors. We report those counts in
Section~\ref{sec:quantify-crossedit} and
Section~\ref{sec:results-edit-size}.

\begin{table}[h]
\centering
\small
\setlength{\tabcolsep}{3pt}
\sisetup{detect-weight=true, detect-family=true}
\begin{tabular}{@{}>{\raggedright\arraybackslash}p{42pt}ccccc@{}}
\toprule
& \textbf{\UseMacro{TCol-qwen}} & \textbf{\UseMacro{TCol-gpto}} & \textbf{\UseMacro{TCol-olmos}} & \textbf{\UseMacro{TCol-olmol}} & \textbf{\UseMacro{TCol-haiku}} \\
\midrule
Pass \reimplementation test & \num{\UseMacro{edit-filter-eval-rust-repos-qwen-wide-passing}} & \num{\UseMacro{edit-filter-eval-rust-repos-gpt-wide-passing}} & \num{\UseMacro{edit-filter-eval-rust-repos-olmo7b-wide-passing}} & \num{\UseMacro{edit-filter-eval-rust-repos-olmo32b-wide-passing}} & \num{\UseMacro{edit-filter-eval-rust-repos-haiku-wide-passing}} \\
\midrule
Overlap below 50\% & \num{\UseMacro{edit-filter-eval-rust-repos-qwen-wide-after-overlap}} & \num{\UseMacro{edit-filter-eval-rust-repos-gpt-wide-after-overlap}} & \num{\UseMacro{edit-filter-eval-rust-repos-olmo7b-wide-after-overlap}} & \num{\UseMacro{edit-filter-eval-rust-repos-olmo32b-wide-after-overlap}} & \num{\UseMacro{edit-filter-eval-rust-repos-haiku-wide-after-overlap}} \\
\midrule
Fail edit test & \num{\UseMacro{edit-filter-eval-rust-repos-qwen-after-presolved}} & \num{\UseMacro{edit-filter-eval-rust-repos-gpt-after-presolved}} & \num{\UseMacro{edit-filter-eval-rust-repos-olmo7b-after-presolved}} & \num{\UseMacro{edit-filter-eval-rust-repos-olmo32b-after-presolved}} & \num{\UseMacro{edit-filter-eval-rust-repos-haiku-after-presolved}} \\
\midrule
\IDPRdesc available & \textbf{\num{\UseMacro{edit-filter-eval-rust-repos-qwen-after-nodesc}}} & \textbf{\num{\UseMacro{edit-filter-eval-rust-repos-gpt-after-nodesc}}} & \textbf{\num{\UseMacro{edit-filter-eval-rust-repos-olmo7b-after-nodesc}}} & \textbf{\num{\UseMacro{edit-filter-eval-rust-repos-olmo32b-after-nodesc}}} & \textbf{\num{\UseMacro{edit-filter-eval-rust-repos-haiku-after-nodesc}}} \\
\bottomrule
\end{tabular}
\caption{\UseMacro{TCap-corpus-attrition-after}}
\label{tab:corpus-attrition-after}
\end{table}

\section{Edit Success Across \Crossediting}
\label{sec:app-edit-success-matrix}

Table~\ref{tab:pr-edit-pass-rate-pivot} shows the percentage of edits
that can pass every test, over the same set of \rewriter and \editor
pairs between the five models used in
Section~\ref{sec:char-crossediting}. These numbers are calculated on
all tasks we created after the \reimplementation process. If a model
fails to produce an edit, we count it as a failure. Here, unlike the
pattern we observed for edit amount, we do not observe edits from \selfediting perform better than \crossedit consistently. More interestingly, 3 out of 5 models perform the best
on \olmol \reimplementations and all models perform the worst on
\haiku \reimplementations. How \crossediting affects a model's ability
to make correct edits is an
interesting question that we leave to future work.

\begin{table}[h]
\centering
\small
\setlength{\tabcolsep}{3pt}
\begin{tabular}{l c c c c c}
\toprule
\multirow{2}{*}{\textbf{Impl.}} & \multicolumn{5}{c}{\textbf{Editor}} \\
\cmidrule(lr){2-6}
& \textbf{\UseMacro{TCol-qwen}} & \textbf{\UseMacro{TCol-gpto}} & \textbf{\UseMacro{TCol-olmos}} & \textbf{\UseMacro{TCol-olmol}} & \textbf{\UseMacro{TCol-haiku}} \\
\midrule
\textbf{\UseMacro{TColInline-qwen}} & \UseMacro{fig-pr-edit-pass-rate-pivot-qwen3.5-thinking:35b-a3b-q4-k-m-qwen3.5-thinking:35b-a3b-q4-k-m-tests} & \UseMacro{fig-pr-edit-pass-rate-pivot-qwen3.5-thinking:35b-a3b-q4-k-m-gpt-oss:20b-q4-k-m-tests} & \UseMacro{fig-pr-edit-pass-rate-pivot-qwen3.5-thinking:35b-a3b-q4-k-m-olmo3-thinking:7b.think-q4-k-m-tests} & \UseMacro{fig-pr-edit-pass-rate-pivot-qwen3.5-thinking:35b-a3b-q4-k-m-olmo3.1-thinking:32b.think-q4-k-m-tests} & \textbf{\UseMacro{fig-pr-edit-pass-rate-pivot-qwen3.5-thinking:35b-a3b-q4-k-m-claude-haiku-4-5-tests}} \\
\textbf{\UseMacro{TColInline-gpto}} & \UseMacro{fig-pr-edit-pass-rate-pivot-gpt-oss:20b-q4-k-m-qwen3.5-thinking:35b-a3b-q4-k-m-tests} & \UseMacro{fig-pr-edit-pass-rate-pivot-gpt-oss:20b-q4-k-m-gpt-oss:20b-q4-k-m-tests} & \UseMacro{fig-pr-edit-pass-rate-pivot-gpt-oss:20b-q4-k-m-olmo3-thinking:7b.think-q4-k-m-tests} & \UseMacro{fig-pr-edit-pass-rate-pivot-gpt-oss:20b-q4-k-m-olmo3.1-thinking:32b.think-q4-k-m-tests} & \UseMacro{fig-pr-edit-pass-rate-pivot-gpt-oss:20b-q4-k-m-claude-haiku-4-5-tests} \\
\textbf{\UseMacro{TColInline-olmos}} & \UseMacro{fig-pr-edit-pass-rate-pivot-olmo3-thinking:7b.think-q4-k-m-qwen3.5-thinking:35b-a3b-q4-k-m-tests} & \UseMacro{fig-pr-edit-pass-rate-pivot-olmo3-thinking:7b.think-q4-k-m-gpt-oss:20b-q4-k-m-tests} & \textbf{\UseMacro{fig-pr-edit-pass-rate-pivot-olmo3-thinking:7b.think-q4-k-m-olmo3-thinking:7b.think-q4-k-m-tests}} & \UseMacro{fig-pr-edit-pass-rate-pivot-olmo3-thinking:7b.think-q4-k-m-olmo3.1-thinking:32b.think-q4-k-m-tests} & \UseMacro{fig-pr-edit-pass-rate-pivot-olmo3-thinking:7b.think-q4-k-m-claude-haiku-4-5-tests} \\
\textbf{\UseMacro{TColInline-olmol}} & \textbf{\UseMacro{fig-pr-edit-pass-rate-pivot-olmo3.1-thinking:32b.think-q4-k-m-qwen3.5-thinking:35b-a3b-q4-k-m-tests}} & \textbf{\UseMacro{fig-pr-edit-pass-rate-pivot-olmo3.1-thinking:32b.think-q4-k-m-gpt-oss:20b-q4-k-m-tests}} & \UseMacro{fig-pr-edit-pass-rate-pivot-olmo3.1-thinking:32b.think-q4-k-m-olmo3-thinking:7b.think-q4-k-m-tests} & \textbf{\UseMacro{fig-pr-edit-pass-rate-pivot-olmo3.1-thinking:32b.think-q4-k-m-olmo3.1-thinking:32b.think-q4-k-m-tests}} & \UseMacro{fig-pr-edit-pass-rate-pivot-olmo3.1-thinking:32b.think-q4-k-m-claude-haiku-4-5-tests} \\
\textbf{\UseMacro{TColInline-haiku}} & \UseMacro{fig-pr-edit-pass-rate-pivot-claude-haiku-4-5-qwen3.5-thinking:35b-a3b-q4-k-m-tests} & \UseMacro{fig-pr-edit-pass-rate-pivot-claude-haiku-4-5-gpt-oss:20b-q4-k-m-tests} & \UseMacro{fig-pr-edit-pass-rate-pivot-claude-haiku-4-5-olmo3-thinking:7b.think-q4-k-m-tests} & \UseMacro{fig-pr-edit-pass-rate-pivot-claude-haiku-4-5-olmo3.1-thinking:32b.think-q4-k-m-tests} & \UseMacro{fig-pr-edit-pass-rate-pivot-claude-haiku-4-5-claude-haiku-4-5-tests} \\
\bottomrule
\end{tabular}
\caption{\UseMacro{TCap-pr-edit-pass-rate-pivot}}
\label{tab:pr-edit-pass-rate-pivot}
\end{table}

\clearpage

\section{Mean Fraction of Tests Passed Excluding No-Change Responses}
\label{sec:app-verl-meantestnc}

As described in Section~\ref{sec:results-edit-success}, the mean fraction of tests passed
(\UseMacro{TColInline-meantest}) can also potentially be inflated if the model returns
the \reimplementation unchanged. Here, we also split the \tasks into two disjoint subsets: \tasks
where the \rlmodel makes no change
(\UseMacro{TColInline-meantest-nc}), and \tasks where the \rlmodel
actually modifies the \reimplementation
(\UseMacro{TColInline-meantest-wc}). In Table~\ref{tab:verl-meantestnc-eval-rust-repos}, we observe that making no change
does not always work in the \rlmodel's favor. On three of the five \rewriters, \rllabel does not have the highest mean fraction of tests passed on the no-change subset. In contrast, on the with-change subset,
\rllabel's mean fraction of tests passed exceeds the \baselabel's for every
\rewriter, even for \olmos itself, showing that its mean test advantage
does not originate from declining to edit.

\begin{table}[h]
\centering
\small
\setlength{\tabcolsep}{3pt}
\begin{tabular}{l c c c}
\toprule
\shortstack[l]{\textbf{Implementor} \\ \quad\textbf{Editor}} & \UseMacro{TCol-meantest} & \UseMacro{TCol-meantest-nc} & \UseMacro{TCol-meantest-wc} \\
\midrule
\multicolumn{4}{@{}l}{\textbf{\qwen}} \\
\quad \baselabel & \LargestBoldSafe{\UseMacro{fig-verl-meantestnc-base-qwen-mean-tests}}{\UseMacro{fig-verl-meantestnc-prompt-qwen-mean-tests}}{\UseMacro{fig-verl-meantestnc-verl-step62-qwen-mean-tests}} & \LargestBoldSafe{\UseMacro{fig-verl-meantestnc-base-qwen-no-change}}{\UseMacro{fig-verl-meantestnc-prompt-qwen-no-change}}{\UseMacro{fig-verl-meantestnc-verl-step62-qwen-no-change}} & \LargestBoldSafe{\UseMacro{fig-verl-meantestnc-base-qwen-mean-tests-change}}{\UseMacro{fig-verl-meantestnc-prompt-qwen-mean-tests-change}}{\UseMacro{fig-verl-meantestnc-verl-step62-qwen-mean-tests-change}} \\
\quad \strictlabel & \LargestBoldSafe{\UseMacro{fig-verl-meantestnc-prompt-qwen-mean-tests}}{\UseMacro{fig-verl-meantestnc-base-qwen-mean-tests}}{\UseMacro{fig-verl-meantestnc-verl-step62-qwen-mean-tests}} & \LargestBoldSafe{\UseMacro{fig-verl-meantestnc-prompt-qwen-no-change}}{\UseMacro{fig-verl-meantestnc-base-qwen-no-change}}{\UseMacro{fig-verl-meantestnc-verl-step62-qwen-no-change}} & \LargestBoldSafe{\UseMacro{fig-verl-meantestnc-prompt-qwen-mean-tests-change}}{\UseMacro{fig-verl-meantestnc-base-qwen-mean-tests-change}}{\UseMacro{fig-verl-meantestnc-verl-step62-qwen-mean-tests-change}} \\
\quad \rllabel & \LargestBoldSafe{\UseMacro{fig-verl-meantestnc-verl-step62-qwen-mean-tests}}{\UseMacro{fig-verl-meantestnc-base-qwen-mean-tests}}{\UseMacro{fig-verl-meantestnc-prompt-qwen-mean-tests}} & \LargestBoldSafe{\UseMacro{fig-verl-meantestnc-verl-step62-qwen-no-change}}{\UseMacro{fig-verl-meantestnc-base-qwen-no-change}}{\UseMacro{fig-verl-meantestnc-prompt-qwen-no-change}} & \LargestBoldSafe{\UseMacro{fig-verl-meantestnc-verl-step62-qwen-mean-tests-change}}{\UseMacro{fig-verl-meantestnc-base-qwen-mean-tests-change}}{\UseMacro{fig-verl-meantestnc-prompt-qwen-mean-tests-change}} \\
\midrule
\multicolumn{4}{@{}l}{\textbf{\gpto}} \\
\quad \baselabel & \LargestBoldSafe{\UseMacro{fig-verl-meantestnc-base-gpt-mean-tests}}{\UseMacro{fig-verl-meantestnc-prompt-gpt-mean-tests}}{\UseMacro{fig-verl-meantestnc-verl-step62-gpt-mean-tests}} & \LargestBoldSafe{\UseMacro{fig-verl-meantestnc-base-gpt-no-change}}{\UseMacro{fig-verl-meantestnc-prompt-gpt-no-change}}{\UseMacro{fig-verl-meantestnc-verl-step62-gpt-no-change}} & \LargestBoldSafe{\UseMacro{fig-verl-meantestnc-base-gpt-mean-tests-change}}{\UseMacro{fig-verl-meantestnc-prompt-gpt-mean-tests-change}}{\UseMacro{fig-verl-meantestnc-verl-step62-gpt-mean-tests-change}} \\
\quad \strictlabel & \LargestBoldSafe{\UseMacro{fig-verl-meantestnc-prompt-gpt-mean-tests}}{\UseMacro{fig-verl-meantestnc-base-gpt-mean-tests}}{\UseMacro{fig-verl-meantestnc-verl-step62-gpt-mean-tests}} & \LargestBoldSafe{\UseMacro{fig-verl-meantestnc-prompt-gpt-no-change}}{\UseMacro{fig-verl-meantestnc-base-gpt-no-change}}{\UseMacro{fig-verl-meantestnc-verl-step62-gpt-no-change}} & \LargestBoldSafe{\UseMacro{fig-verl-meantestnc-prompt-gpt-mean-tests-change}}{\UseMacro{fig-verl-meantestnc-base-gpt-mean-tests-change}}{\UseMacro{fig-verl-meantestnc-verl-step62-gpt-mean-tests-change}} \\
\quad \rllabel & \LargestBoldSafe{\UseMacro{fig-verl-meantestnc-verl-step62-gpt-mean-tests}}{\UseMacro{fig-verl-meantestnc-base-gpt-mean-tests}}{\UseMacro{fig-verl-meantestnc-prompt-gpt-mean-tests}} & \LargestBoldSafe{\UseMacro{fig-verl-meantestnc-verl-step62-gpt-no-change}}{\UseMacro{fig-verl-meantestnc-base-gpt-no-change}}{\UseMacro{fig-verl-meantestnc-prompt-gpt-no-change}} & \LargestBoldSafe{\UseMacro{fig-verl-meantestnc-verl-step62-gpt-mean-tests-change}}{\UseMacro{fig-verl-meantestnc-base-gpt-mean-tests-change}}{\UseMacro{fig-verl-meantestnc-prompt-gpt-mean-tests-change}} \\
\midrule
\multicolumn{4}{@{}l}{\textbf{\olmos}} \\
\quad \baselabel & \LargestBoldSafe{\UseMacro{fig-verl-meantestnc-base-olmo7b-mean-tests}}{\UseMacro{fig-verl-meantestnc-prompt-olmo7b-mean-tests}}{\UseMacro{fig-verl-meantestnc-verl-step62-olmo7b-mean-tests}} & \LargestBoldSafe{\UseMacro{fig-verl-meantestnc-base-olmo7b-no-change}}{\UseMacro{fig-verl-meantestnc-prompt-olmo7b-no-change}}{\UseMacro{fig-verl-meantestnc-verl-step62-olmo7b-no-change}} & \LargestBoldSafe{\UseMacro{fig-verl-meantestnc-base-olmo7b-mean-tests-change}}{\UseMacro{fig-verl-meantestnc-prompt-olmo7b-mean-tests-change}}{\UseMacro{fig-verl-meantestnc-verl-step62-olmo7b-mean-tests-change}} \\
\quad \strictlabel & \LargestBoldSafe{\UseMacro{fig-verl-meantestnc-prompt-olmo7b-mean-tests}}{\UseMacro{fig-verl-meantestnc-base-olmo7b-mean-tests}}{\UseMacro{fig-verl-meantestnc-verl-step62-olmo7b-mean-tests}} & \LargestBoldSafe{\UseMacro{fig-verl-meantestnc-prompt-olmo7b-no-change}}{\UseMacro{fig-verl-meantestnc-base-olmo7b-no-change}}{\UseMacro{fig-verl-meantestnc-verl-step62-olmo7b-no-change}} & \LargestBoldSafe{\UseMacro{fig-verl-meantestnc-prompt-olmo7b-mean-tests-change}}{\UseMacro{fig-verl-meantestnc-base-olmo7b-mean-tests-change}}{\UseMacro{fig-verl-meantestnc-verl-step62-olmo7b-mean-tests-change}} \\
\quad \rllabel & \LargestBoldSafe{\UseMacro{fig-verl-meantestnc-verl-step62-olmo7b-mean-tests}}{\UseMacro{fig-verl-meantestnc-base-olmo7b-mean-tests}}{\UseMacro{fig-verl-meantestnc-prompt-olmo7b-mean-tests}} & \LargestBoldSafe{\UseMacro{fig-verl-meantestnc-verl-step62-olmo7b-no-change}}{\UseMacro{fig-verl-meantestnc-base-olmo7b-no-change}}{\UseMacro{fig-verl-meantestnc-prompt-olmo7b-no-change}} & \LargestBoldSafe{\UseMacro{fig-verl-meantestnc-verl-step62-olmo7b-mean-tests-change}}{\UseMacro{fig-verl-meantestnc-base-olmo7b-mean-tests-change}}{\UseMacro{fig-verl-meantestnc-prompt-olmo7b-mean-tests-change}} \\
\midrule
\multicolumn{4}{@{}l}{\textbf{\olmol}} \\
\quad \baselabel & \LargestBoldSafe{\UseMacro{fig-verl-meantestnc-base-olmo32b-mean-tests}}{\UseMacro{fig-verl-meantestnc-prompt-olmo32b-mean-tests}}{\UseMacro{fig-verl-meantestnc-verl-step62-olmo32b-mean-tests}} & \LargestBoldSafe{\UseMacro{fig-verl-meantestnc-base-olmo32b-no-change}}{\UseMacro{fig-verl-meantestnc-prompt-olmo32b-no-change}}{\UseMacro{fig-verl-meantestnc-verl-step62-olmo32b-no-change}} & \LargestBoldSafe{\UseMacro{fig-verl-meantestnc-base-olmo32b-mean-tests-change}}{\UseMacro{fig-verl-meantestnc-prompt-olmo32b-mean-tests-change}}{\UseMacro{fig-verl-meantestnc-verl-step62-olmo32b-mean-tests-change}} \\
\quad \strictlabel & \LargestBoldSafe{\UseMacro{fig-verl-meantestnc-prompt-olmo32b-mean-tests}}{\UseMacro{fig-verl-meantestnc-base-olmo32b-mean-tests}}{\UseMacro{fig-verl-meantestnc-verl-step62-olmo32b-mean-tests}} & \LargestBoldSafe{\UseMacro{fig-verl-meantestnc-prompt-olmo32b-no-change}}{\UseMacro{fig-verl-meantestnc-base-olmo32b-no-change}}{\UseMacro{fig-verl-meantestnc-verl-step62-olmo32b-no-change}} & \LargestBoldSafe{\UseMacro{fig-verl-meantestnc-prompt-olmo32b-mean-tests-change}}{\UseMacro{fig-verl-meantestnc-base-olmo32b-mean-tests-change}}{\UseMacro{fig-verl-meantestnc-verl-step62-olmo32b-mean-tests-change}} \\
\quad \rllabel & \LargestBoldSafe{\UseMacro{fig-verl-meantestnc-verl-step62-olmo32b-mean-tests}}{\UseMacro{fig-verl-meantestnc-base-olmo32b-mean-tests}}{\UseMacro{fig-verl-meantestnc-prompt-olmo32b-mean-tests}} & \LargestBoldSafe{\UseMacro{fig-verl-meantestnc-verl-step62-olmo32b-no-change}}{\UseMacro{fig-verl-meantestnc-base-olmo32b-no-change}}{\UseMacro{fig-verl-meantestnc-prompt-olmo32b-no-change}} & \LargestBoldSafe{\UseMacro{fig-verl-meantestnc-verl-step62-olmo32b-mean-tests-change}}{\UseMacro{fig-verl-meantestnc-base-olmo32b-mean-tests-change}}{\UseMacro{fig-verl-meantestnc-prompt-olmo32b-mean-tests-change}} \\
\midrule
\multicolumn{4}{@{}l}{\textbf{\haiku}} \\
\quad \baselabel & \LargestBoldSafe{\UseMacro{fig-verl-meantestnc-base-haiku-mean-tests}}{\UseMacro{fig-verl-meantestnc-prompt-haiku-mean-tests}}{\UseMacro{fig-verl-meantestnc-verl-step62-haiku-mean-tests}} & \LargestBoldSafe{\UseMacro{fig-verl-meantestnc-base-haiku-no-change}}{\UseMacro{fig-verl-meantestnc-prompt-haiku-no-change}}{\UseMacro{fig-verl-meantestnc-verl-step62-haiku-no-change}} & \LargestBoldSafe{\UseMacro{fig-verl-meantestnc-base-haiku-mean-tests-change}}{\UseMacro{fig-verl-meantestnc-prompt-haiku-mean-tests-change}}{\UseMacro{fig-verl-meantestnc-verl-step62-haiku-mean-tests-change}} \\
\quad \strictlabel & \LargestBoldSafe{\UseMacro{fig-verl-meantestnc-prompt-haiku-mean-tests}}{\UseMacro{fig-verl-meantestnc-base-haiku-mean-tests}}{\UseMacro{fig-verl-meantestnc-verl-step62-haiku-mean-tests}} & \LargestBoldSafe{\UseMacro{fig-verl-meantestnc-prompt-haiku-no-change}}{\UseMacro{fig-verl-meantestnc-base-haiku-no-change}}{\UseMacro{fig-verl-meantestnc-verl-step62-haiku-no-change}} & \LargestBoldSafe{\UseMacro{fig-verl-meantestnc-prompt-haiku-mean-tests-change}}{\UseMacro{fig-verl-meantestnc-base-haiku-mean-tests-change}}{\UseMacro{fig-verl-meantestnc-verl-step62-haiku-mean-tests-change}} \\
\quad \rllabel & \LargestBoldSafe{\UseMacro{fig-verl-meantestnc-verl-step62-haiku-mean-tests}}{\UseMacro{fig-verl-meantestnc-base-haiku-mean-tests}}{\UseMacro{fig-verl-meantestnc-prompt-haiku-mean-tests}} & \LargestBoldSafe{\UseMacro{fig-verl-meantestnc-verl-step62-haiku-no-change}}{\UseMacro{fig-verl-meantestnc-base-haiku-no-change}}{\UseMacro{fig-verl-meantestnc-prompt-haiku-no-change}} & \LargestBoldSafe{\UseMacro{fig-verl-meantestnc-verl-step62-haiku-mean-tests-change}}{\UseMacro{fig-verl-meantestnc-base-haiku-mean-tests-change}}{\UseMacro{fig-verl-meantestnc-prompt-haiku-mean-tests-change}} \\
\bottomrule
\end{tabular}
\caption{\UseMacro{TCap-verl-meantestnc-eval-rust-repos}}
\label{tab:verl-meantestnc-eval-rust-repos}
\end{table}

\section{Absolute Edit Distance and Line Granularity}
\label{sec:app-verl-lev-raw}

Table~\ref{tab:verl-lev-raw-rustfmt-eval-rust-repos} and
Table~\ref{tab:verl-lev-normfull-rustfmt-eval-rust-repos} report the
comparison between \Tool and the baselines in \levmetric in absolute and in
min-max normalized form. We report the numbers at line and character
granularity. Both carry all four populations described in
Section~\ref{sec:results-edit-size}.

\MyPara{Absolute distances
(Table~\ref{tab:verl-lev-raw-rustfmt-eval-rust-repos})}
Rows are grouped by \rewriter, and inside each group we compare the 3 \editors: \baselabel, \strictlabel, and \rllabel.
\textbf{All \tasks} covers every \task. Each \textbf{Both pass}
column is restricted to the \tasks on which the two \editors it names both produce
an edit that passes all tests. We leave the \editor not named out of the comparison and show only dashes there. The populations are visualized in Figure~\ref{fig:verl-venn} in Section~\ref{sec:results-edit-size}. \emph{Chars} and \emph{Lines} are the \levmetric calculated at character and line granularity. The smallest value of
each comparison is bolded, and a dagger marks a cell averaged over ten
or fewer \tasks. We see that the \rlmodel roughly halves the edit distance in absolute numbers and it still maintains an advantage over the \basemodel and the \minprompt when restricted to tasks where \rlmodel and baselines both produce an edit that passes all tests. Interestingly, when looking at tasks where both baselines produce an edit that passes all tests, the \minprompt actually consistently performs worse. This further shows that prompt engineering alone cannot consistently reduce excessive editing.

\MyPara{Normalized
distances
(Table~\ref{tab:verl-lev-normfull-rustfmt-eval-rust-repos})} This
table min-max normalizes the \levmetric over the same \tasks, with the
pool of \editors described in Section~\ref{sec:results-edit-size}.
Table~\ref{tab:verl-lev-norm-eval-rust-repos} in the main paper
prints the \emph{Chars} column of each group from this table, while
this table adds an additional \emph{Lines} column for \levmetric at
line granularity. The results at line granularity still show that
\rllabel makes significantly smaller edits than the baselines on all
\rewriters. This shows us that \rlmodel did not simply
optimize edit size by making small edits in several locations. We can
also observe that the \minprompt still cannot lower the edit amount
consistently at line granularity. The line-level \levmetric results
prove to us that \Tool can potentially reduce the burden for reviewers
who read the edit diffs that show each lines where changes occur.

\clearpage

\begin{table*}[h]
\centering
\small
\setlength{\tabcolsep}{3pt}
\begin{tabular}{l cc cc cc cc}
\toprule
\multirow{2}{*}{\shortstack[l]{\textbf{Implementor} \\ \quad\textbf{Editor}}} & \multicolumn{2}{c}{\shortstack{\textbf{All} \\ \textbf{\tasks}}} & \multicolumn{2}{c}{\shortstack{\textbf{Base \&} \\ \textbf{Strict pass}}} & \multicolumn{2}{c}{\shortstack{\textbf{Base \&} \\ \textbf{RL pass}}} & \multicolumn{2}{c}{\shortstack{\textbf{Strict \&} \\ \textbf{RL pass}}} \\
\cmidrule(lr){2-3} \cmidrule(lr){4-5} \cmidrule(lr){6-7} \cmidrule(lr){8-9}
& \textbf{Chars} & \textbf{Lines} & \textbf{Chars} & \textbf{Lines} & \textbf{Chars} & \textbf{Lines} & \textbf{Chars} & \textbf{Lines} \\
\midrule
\multicolumn{9}{@{}l}{\textbf{\qwen}} \\
\quad \baselabel & \SmallestBoldSafe{\UseMacro{fig-verl-comparison-allmean-rustfmt-base-qwen-lev-char}}{\UseMacro{fig-verl-comparison-allmean-rustfmt-prompt-qwen-lev-char}}{\UseMacro{fig-verl-comparison-allmean-rustfmt-verl-step62-qwen-lev-char}} & \SmallestBoldSafe{\UseMacro{fig-verl-comparison-allmean-rustfmt-base-qwen-lev-line}}{\UseMacro{fig-verl-comparison-allmean-rustfmt-prompt-qwen-lev-line}}{\UseMacro{fig-verl-comparison-allmean-rustfmt-verl-step62-qwen-lev-line}} & \SmallerBoldSafe{\UseMacro{fig-verl-lev-base-strict-rustfmt-qwen-base-mean-char}}{\UseMacro{fig-verl-lev-base-strict-rustfmt-qwen-prompt-mean-char}}\UseMacro{fig-verl-lev-base-strict-rustfmt-qwen-small} & \SmallerBoldSafe{\UseMacro{fig-verl-lev-base-strict-rustfmt-qwen-base-mean-line}}{\UseMacro{fig-verl-lev-base-strict-rustfmt-qwen-prompt-mean-line}}\UseMacro{fig-verl-lev-base-strict-rustfmt-qwen-small} & \SmallerBoldSafe{\UseMacro{fig-verl-lev-both-pass-rustfmt-qwen-base-mean-char}}{\UseMacro{fig-verl-lev-both-pass-rustfmt-qwen-rl-mean-char}}\UseMacro{fig-verl-lev-both-pass-rustfmt-qwen-small} & \SmallerBoldSafe{\UseMacro{fig-verl-lev-both-pass-rustfmt-qwen-base-mean-line}}{\UseMacro{fig-verl-lev-both-pass-rustfmt-qwen-rl-mean-line}}\UseMacro{fig-verl-lev-both-pass-rustfmt-qwen-small} & -- & -- \\
\quad \strictlabel & \SmallestBoldSafe{\UseMacro{fig-verl-comparison-allmean-rustfmt-prompt-qwen-lev-char}}{\UseMacro{fig-verl-comparison-allmean-rustfmt-base-qwen-lev-char}}{\UseMacro{fig-verl-comparison-allmean-rustfmt-verl-step62-qwen-lev-char}} & \SmallestBoldSafe{\UseMacro{fig-verl-comparison-allmean-rustfmt-prompt-qwen-lev-line}}{\UseMacro{fig-verl-comparison-allmean-rustfmt-base-qwen-lev-line}}{\UseMacro{fig-verl-comparison-allmean-rustfmt-verl-step62-qwen-lev-line}} & \SmallerBoldSafe{\UseMacro{fig-verl-lev-base-strict-rustfmt-qwen-prompt-mean-char}}{\UseMacro{fig-verl-lev-base-strict-rustfmt-qwen-base-mean-char}}\UseMacro{fig-verl-lev-base-strict-rustfmt-qwen-small} & \SmallerBoldSafe{\UseMacro{fig-verl-lev-base-strict-rustfmt-qwen-prompt-mean-line}}{\UseMacro{fig-verl-lev-base-strict-rustfmt-qwen-base-mean-line}}\UseMacro{fig-verl-lev-base-strict-rustfmt-qwen-small} & -- & -- & \SmallerBoldSafe{\UseMacro{fig-verl-lev-prompt-pass-rustfmt-qwen-prompt-mean-char}}{\UseMacro{fig-verl-lev-prompt-pass-rustfmt-qwen-rl-mean-char}}\UseMacro{fig-verl-lev-prompt-pass-rustfmt-qwen-small} & \SmallerBoldSafe{\UseMacro{fig-verl-lev-prompt-pass-rustfmt-qwen-prompt-mean-line}}{\UseMacro{fig-verl-lev-prompt-pass-rustfmt-qwen-rl-mean-line}}\UseMacro{fig-verl-lev-prompt-pass-rustfmt-qwen-small} \\
\quad \rllabel & \SmallestBoldSafe{\UseMacro{fig-verl-comparison-allmean-rustfmt-verl-step62-qwen-lev-char}}{\UseMacro{fig-verl-comparison-allmean-rustfmt-base-qwen-lev-char}}{\UseMacro{fig-verl-comparison-allmean-rustfmt-prompt-qwen-lev-char}} & \SmallestBoldSafe{\UseMacro{fig-verl-comparison-allmean-rustfmt-verl-step62-qwen-lev-line}}{\UseMacro{fig-verl-comparison-allmean-rustfmt-base-qwen-lev-line}}{\UseMacro{fig-verl-comparison-allmean-rustfmt-prompt-qwen-lev-line}} & -- & -- & \SmallerBoldSafe{\UseMacro{fig-verl-lev-both-pass-rustfmt-qwen-rl-mean-char}}{\UseMacro{fig-verl-lev-both-pass-rustfmt-qwen-base-mean-char}}\UseMacro{fig-verl-lev-both-pass-rustfmt-qwen-small} & \SmallerBoldSafe{\UseMacro{fig-verl-lev-both-pass-rustfmt-qwen-rl-mean-line}}{\UseMacro{fig-verl-lev-both-pass-rustfmt-qwen-base-mean-line}}\UseMacro{fig-verl-lev-both-pass-rustfmt-qwen-small} & \SmallerBoldSafe{\UseMacro{fig-verl-lev-prompt-pass-rustfmt-qwen-rl-mean-char}}{\UseMacro{fig-verl-lev-prompt-pass-rustfmt-qwen-prompt-mean-char}}\UseMacro{fig-verl-lev-prompt-pass-rustfmt-qwen-small} & \SmallerBoldSafe{\UseMacro{fig-verl-lev-prompt-pass-rustfmt-qwen-rl-mean-line}}{\UseMacro{fig-verl-lev-prompt-pass-rustfmt-qwen-prompt-mean-line}}\UseMacro{fig-verl-lev-prompt-pass-rustfmt-qwen-small} \\
\midrule
\multicolumn{9}{@{}l}{\textbf{\gpto}} \\
\quad \baselabel & \SmallestBoldSafe{\UseMacro{fig-verl-comparison-allmean-rustfmt-base-gpt-lev-char}}{\UseMacro{fig-verl-comparison-allmean-rustfmt-prompt-gpt-lev-char}}{\UseMacro{fig-verl-comparison-allmean-rustfmt-verl-step62-gpt-lev-char}} & \SmallestBoldSafe{\UseMacro{fig-verl-comparison-allmean-rustfmt-base-gpt-lev-line}}{\UseMacro{fig-verl-comparison-allmean-rustfmt-prompt-gpt-lev-line}}{\UseMacro{fig-verl-comparison-allmean-rustfmt-verl-step62-gpt-lev-line}} & \SmallerBoldSafe{\UseMacro{fig-verl-lev-base-strict-rustfmt-gpt-base-mean-char}}{\UseMacro{fig-verl-lev-base-strict-rustfmt-gpt-prompt-mean-char}}\UseMacro{fig-verl-lev-base-strict-rustfmt-gpt-small} & \SmallerBoldSafe{\UseMacro{fig-verl-lev-base-strict-rustfmt-gpt-base-mean-line}}{\UseMacro{fig-verl-lev-base-strict-rustfmt-gpt-prompt-mean-line}}\UseMacro{fig-verl-lev-base-strict-rustfmt-gpt-small} & \SmallerBoldSafe{\UseMacro{fig-verl-lev-both-pass-rustfmt-gpt-base-mean-char}}{\UseMacro{fig-verl-lev-both-pass-rustfmt-gpt-rl-mean-char}}\UseMacro{fig-verl-lev-both-pass-rustfmt-gpt-small} & \SmallerBoldSafe{\UseMacro{fig-verl-lev-both-pass-rustfmt-gpt-base-mean-line}}{\UseMacro{fig-verl-lev-both-pass-rustfmt-gpt-rl-mean-line}}\UseMacro{fig-verl-lev-both-pass-rustfmt-gpt-small} & -- & -- \\
\quad \strictlabel & \SmallestBoldSafe{\UseMacro{fig-verl-comparison-allmean-rustfmt-prompt-gpt-lev-char}}{\UseMacro{fig-verl-comparison-allmean-rustfmt-base-gpt-lev-char}}{\UseMacro{fig-verl-comparison-allmean-rustfmt-verl-step62-gpt-lev-char}} & \SmallestBoldSafe{\UseMacro{fig-verl-comparison-allmean-rustfmt-prompt-gpt-lev-line}}{\UseMacro{fig-verl-comparison-allmean-rustfmt-base-gpt-lev-line}}{\UseMacro{fig-verl-comparison-allmean-rustfmt-verl-step62-gpt-lev-line}} & \SmallerBoldSafe{\UseMacro{fig-verl-lev-base-strict-rustfmt-gpt-prompt-mean-char}}{\UseMacro{fig-verl-lev-base-strict-rustfmt-gpt-base-mean-char}}\UseMacro{fig-verl-lev-base-strict-rustfmt-gpt-small} & \SmallerBoldSafe{\UseMacro{fig-verl-lev-base-strict-rustfmt-gpt-prompt-mean-line}}{\UseMacro{fig-verl-lev-base-strict-rustfmt-gpt-base-mean-line}}\UseMacro{fig-verl-lev-base-strict-rustfmt-gpt-small} & -- & -- & \SmallerBoldSafe{\UseMacro{fig-verl-lev-prompt-pass-rustfmt-gpt-prompt-mean-char}}{\UseMacro{fig-verl-lev-prompt-pass-rustfmt-gpt-rl-mean-char}}\UseMacro{fig-verl-lev-prompt-pass-rustfmt-gpt-small} & \SmallerBoldSafe{\UseMacro{fig-verl-lev-prompt-pass-rustfmt-gpt-prompt-mean-line}}{\UseMacro{fig-verl-lev-prompt-pass-rustfmt-gpt-rl-mean-line}}\UseMacro{fig-verl-lev-prompt-pass-rustfmt-gpt-small} \\
\quad \rllabel & \SmallestBoldSafe{\UseMacro{fig-verl-comparison-allmean-rustfmt-verl-step62-gpt-lev-char}}{\UseMacro{fig-verl-comparison-allmean-rustfmt-base-gpt-lev-char}}{\UseMacro{fig-verl-comparison-allmean-rustfmt-prompt-gpt-lev-char}} & \SmallestBoldSafe{\UseMacro{fig-verl-comparison-allmean-rustfmt-verl-step62-gpt-lev-line}}{\UseMacro{fig-verl-comparison-allmean-rustfmt-base-gpt-lev-line}}{\UseMacro{fig-verl-comparison-allmean-rustfmt-prompt-gpt-lev-line}} & -- & -- & \SmallerBoldSafe{\UseMacro{fig-verl-lev-both-pass-rustfmt-gpt-rl-mean-char}}{\UseMacro{fig-verl-lev-both-pass-rustfmt-gpt-base-mean-char}}\UseMacro{fig-verl-lev-both-pass-rustfmt-gpt-small} & \SmallerBoldSafe{\UseMacro{fig-verl-lev-both-pass-rustfmt-gpt-rl-mean-line}}{\UseMacro{fig-verl-lev-both-pass-rustfmt-gpt-base-mean-line}}\UseMacro{fig-verl-lev-both-pass-rustfmt-gpt-small} & \SmallerBoldSafe{\UseMacro{fig-verl-lev-prompt-pass-rustfmt-gpt-rl-mean-char}}{\UseMacro{fig-verl-lev-prompt-pass-rustfmt-gpt-prompt-mean-char}}\UseMacro{fig-verl-lev-prompt-pass-rustfmt-gpt-small} & \SmallerBoldSafe{\UseMacro{fig-verl-lev-prompt-pass-rustfmt-gpt-rl-mean-line}}{\UseMacro{fig-verl-lev-prompt-pass-rustfmt-gpt-prompt-mean-line}}\UseMacro{fig-verl-lev-prompt-pass-rustfmt-gpt-small} \\
\midrule
\multicolumn{9}{@{}l}{\textbf{\olmos}} \\
\quad \baselabel & \SmallestBoldSafe{\UseMacro{fig-verl-comparison-allmean-rustfmt-base-olmo7b-lev-char}}{\UseMacro{fig-verl-comparison-allmean-rustfmt-prompt-olmo7b-lev-char}}{\UseMacro{fig-verl-comparison-allmean-rustfmt-verl-step62-olmo7b-lev-char}} & \SmallestBoldSafe{\UseMacro{fig-verl-comparison-allmean-rustfmt-base-olmo7b-lev-line}}{\UseMacro{fig-verl-comparison-allmean-rustfmt-prompt-olmo7b-lev-line}}{\UseMacro{fig-verl-comparison-allmean-rustfmt-verl-step62-olmo7b-lev-line}} & \SmallerBoldSafe{\UseMacro{fig-verl-lev-base-strict-rustfmt-olmo7b-base-mean-char}}{\UseMacro{fig-verl-lev-base-strict-rustfmt-olmo7b-prompt-mean-char}}\UseMacro{fig-verl-lev-base-strict-rustfmt-olmo7b-small} & \SmallerBoldSafe{\UseMacro{fig-verl-lev-base-strict-rustfmt-olmo7b-base-mean-line}}{\UseMacro{fig-verl-lev-base-strict-rustfmt-olmo7b-prompt-mean-line}}\UseMacro{fig-verl-lev-base-strict-rustfmt-olmo7b-small} & \SmallerBoldSafe{\UseMacro{fig-verl-lev-both-pass-rustfmt-olmo7b-base-mean-char}}{\UseMacro{fig-verl-lev-both-pass-rustfmt-olmo7b-rl-mean-char}}\UseMacro{fig-verl-lev-both-pass-rustfmt-olmo7b-small} & \SmallerBoldSafe{\UseMacro{fig-verl-lev-both-pass-rustfmt-olmo7b-base-mean-line}}{\UseMacro{fig-verl-lev-both-pass-rustfmt-olmo7b-rl-mean-line}}\UseMacro{fig-verl-lev-both-pass-rustfmt-olmo7b-small} & -- & -- \\
\quad \strictlabel & \SmallestBoldSafe{\UseMacro{fig-verl-comparison-allmean-rustfmt-prompt-olmo7b-lev-char}}{\UseMacro{fig-verl-comparison-allmean-rustfmt-base-olmo7b-lev-char}}{\UseMacro{fig-verl-comparison-allmean-rustfmt-verl-step62-olmo7b-lev-char}} & \SmallestBoldSafe{\UseMacro{fig-verl-comparison-allmean-rustfmt-prompt-olmo7b-lev-line}}{\UseMacro{fig-verl-comparison-allmean-rustfmt-base-olmo7b-lev-line}}{\UseMacro{fig-verl-comparison-allmean-rustfmt-verl-step62-olmo7b-lev-line}} & \SmallerBoldSafe{\UseMacro{fig-verl-lev-base-strict-rustfmt-olmo7b-prompt-mean-char}}{\UseMacro{fig-verl-lev-base-strict-rustfmt-olmo7b-base-mean-char}}\UseMacro{fig-verl-lev-base-strict-rustfmt-olmo7b-small} & \SmallerBoldSafe{\UseMacro{fig-verl-lev-base-strict-rustfmt-olmo7b-prompt-mean-line}}{\UseMacro{fig-verl-lev-base-strict-rustfmt-olmo7b-base-mean-line}}\UseMacro{fig-verl-lev-base-strict-rustfmt-olmo7b-small} & -- & -- & \SmallerBoldSafe{\UseMacro{fig-verl-lev-prompt-pass-rustfmt-olmo7b-prompt-mean-char}}{\UseMacro{fig-verl-lev-prompt-pass-rustfmt-olmo7b-rl-mean-char}}\UseMacro{fig-verl-lev-prompt-pass-rustfmt-olmo7b-small} & \SmallerBoldSafe{\UseMacro{fig-verl-lev-prompt-pass-rustfmt-olmo7b-prompt-mean-line}}{\UseMacro{fig-verl-lev-prompt-pass-rustfmt-olmo7b-rl-mean-line}}\UseMacro{fig-verl-lev-prompt-pass-rustfmt-olmo7b-small} \\
\quad \rllabel & \SmallestBoldSafe{\UseMacro{fig-verl-comparison-allmean-rustfmt-verl-step62-olmo7b-lev-char}}{\UseMacro{fig-verl-comparison-allmean-rustfmt-base-olmo7b-lev-char}}{\UseMacro{fig-verl-comparison-allmean-rustfmt-prompt-olmo7b-lev-char}} & \SmallestBoldSafe{\UseMacro{fig-verl-comparison-allmean-rustfmt-verl-step62-olmo7b-lev-line}}{\UseMacro{fig-verl-comparison-allmean-rustfmt-base-olmo7b-lev-line}}{\UseMacro{fig-verl-comparison-allmean-rustfmt-prompt-olmo7b-lev-line}} & -- & -- & \SmallerBoldSafe{\UseMacro{fig-verl-lev-both-pass-rustfmt-olmo7b-rl-mean-char}}{\UseMacro{fig-verl-lev-both-pass-rustfmt-olmo7b-base-mean-char}}\UseMacro{fig-verl-lev-both-pass-rustfmt-olmo7b-small} & \SmallerBoldSafe{\UseMacro{fig-verl-lev-both-pass-rustfmt-olmo7b-rl-mean-line}}{\UseMacro{fig-verl-lev-both-pass-rustfmt-olmo7b-base-mean-line}}\UseMacro{fig-verl-lev-both-pass-rustfmt-olmo7b-small} & \SmallerBoldSafe{\UseMacro{fig-verl-lev-prompt-pass-rustfmt-olmo7b-rl-mean-char}}{\UseMacro{fig-verl-lev-prompt-pass-rustfmt-olmo7b-prompt-mean-char}}\UseMacro{fig-verl-lev-prompt-pass-rustfmt-olmo7b-small} & \SmallerBoldSafe{\UseMacro{fig-verl-lev-prompt-pass-rustfmt-olmo7b-rl-mean-line}}{\UseMacro{fig-verl-lev-prompt-pass-rustfmt-olmo7b-prompt-mean-line}}\UseMacro{fig-verl-lev-prompt-pass-rustfmt-olmo7b-small} \\
\midrule
\multicolumn{9}{@{}l}{\textbf{\olmol}} \\
\quad \baselabel & \SmallestBoldSafe{\UseMacro{fig-verl-comparison-allmean-rustfmt-base-olmo32b-lev-char}}{\UseMacro{fig-verl-comparison-allmean-rustfmt-prompt-olmo32b-lev-char}}{\UseMacro{fig-verl-comparison-allmean-rustfmt-verl-step62-olmo32b-lev-char}} & \SmallestBoldSafe{\UseMacro{fig-verl-comparison-allmean-rustfmt-base-olmo32b-lev-line}}{\UseMacro{fig-verl-comparison-allmean-rustfmt-prompt-olmo32b-lev-line}}{\UseMacro{fig-verl-comparison-allmean-rustfmt-verl-step62-olmo32b-lev-line}} & \SmallerBoldSafe{\UseMacro{fig-verl-lev-base-strict-rustfmt-olmo32b-base-mean-char}}{\UseMacro{fig-verl-lev-base-strict-rustfmt-olmo32b-prompt-mean-char}}\UseMacro{fig-verl-lev-base-strict-rustfmt-olmo32b-small} & \SmallerBoldSafe{\UseMacro{fig-verl-lev-base-strict-rustfmt-olmo32b-base-mean-line}}{\UseMacro{fig-verl-lev-base-strict-rustfmt-olmo32b-prompt-mean-line}}\UseMacro{fig-verl-lev-base-strict-rustfmt-olmo32b-small} & \SmallerBoldSafe{\UseMacro{fig-verl-lev-both-pass-rustfmt-olmo32b-base-mean-char}}{\UseMacro{fig-verl-lev-both-pass-rustfmt-olmo32b-rl-mean-char}}\UseMacro{fig-verl-lev-both-pass-rustfmt-olmo32b-small} & \SmallerBoldSafe{\UseMacro{fig-verl-lev-both-pass-rustfmt-olmo32b-base-mean-line}}{\UseMacro{fig-verl-lev-both-pass-rustfmt-olmo32b-rl-mean-line}}\UseMacro{fig-verl-lev-both-pass-rustfmt-olmo32b-small} & -- & -- \\
\quad \strictlabel & \SmallestBoldSafe{\UseMacro{fig-verl-comparison-allmean-rustfmt-prompt-olmo32b-lev-char}}{\UseMacro{fig-verl-comparison-allmean-rustfmt-base-olmo32b-lev-char}}{\UseMacro{fig-verl-comparison-allmean-rustfmt-verl-step62-olmo32b-lev-char}} & \SmallestBoldSafe{\UseMacro{fig-verl-comparison-allmean-rustfmt-prompt-olmo32b-lev-line}}{\UseMacro{fig-verl-comparison-allmean-rustfmt-base-olmo32b-lev-line}}{\UseMacro{fig-verl-comparison-allmean-rustfmt-verl-step62-olmo32b-lev-line}} & \SmallerBoldSafe{\UseMacro{fig-verl-lev-base-strict-rustfmt-olmo32b-prompt-mean-char}}{\UseMacro{fig-verl-lev-base-strict-rustfmt-olmo32b-base-mean-char}}\UseMacro{fig-verl-lev-base-strict-rustfmt-olmo32b-small} & \SmallerBoldSafe{\UseMacro{fig-verl-lev-base-strict-rustfmt-olmo32b-prompt-mean-line}}{\UseMacro{fig-verl-lev-base-strict-rustfmt-olmo32b-base-mean-line}}\UseMacro{fig-verl-lev-base-strict-rustfmt-olmo32b-small} & -- & -- & \SmallerBoldSafe{\UseMacro{fig-verl-lev-prompt-pass-rustfmt-olmo32b-prompt-mean-char}}{\UseMacro{fig-verl-lev-prompt-pass-rustfmt-olmo32b-rl-mean-char}}\UseMacro{fig-verl-lev-prompt-pass-rustfmt-olmo32b-small} & \SmallerBoldSafe{\UseMacro{fig-verl-lev-prompt-pass-rustfmt-olmo32b-prompt-mean-line}}{\UseMacro{fig-verl-lev-prompt-pass-rustfmt-olmo32b-rl-mean-line}}\UseMacro{fig-verl-lev-prompt-pass-rustfmt-olmo32b-small} \\
\quad \rllabel & \SmallestBoldSafe{\UseMacro{fig-verl-comparison-allmean-rustfmt-verl-step62-olmo32b-lev-char}}{\UseMacro{fig-verl-comparison-allmean-rustfmt-base-olmo32b-lev-char}}{\UseMacro{fig-verl-comparison-allmean-rustfmt-prompt-olmo32b-lev-char}} & \SmallestBoldSafe{\UseMacro{fig-verl-comparison-allmean-rustfmt-verl-step62-olmo32b-lev-line}}{\UseMacro{fig-verl-comparison-allmean-rustfmt-base-olmo32b-lev-line}}{\UseMacro{fig-verl-comparison-allmean-rustfmt-prompt-olmo32b-lev-line}} & -- & -- & \SmallerBoldSafe{\UseMacro{fig-verl-lev-both-pass-rustfmt-olmo32b-rl-mean-char}}{\UseMacro{fig-verl-lev-both-pass-rustfmt-olmo32b-base-mean-char}}\UseMacro{fig-verl-lev-both-pass-rustfmt-olmo32b-small} & \SmallerBoldSafe{\UseMacro{fig-verl-lev-both-pass-rustfmt-olmo32b-rl-mean-line}}{\UseMacro{fig-verl-lev-both-pass-rustfmt-olmo32b-base-mean-line}}\UseMacro{fig-verl-lev-both-pass-rustfmt-olmo32b-small} & \SmallerBoldSafe{\UseMacro{fig-verl-lev-prompt-pass-rustfmt-olmo32b-rl-mean-char}}{\UseMacro{fig-verl-lev-prompt-pass-rustfmt-olmo32b-prompt-mean-char}}\UseMacro{fig-verl-lev-prompt-pass-rustfmt-olmo32b-small} & \SmallerBoldSafe{\UseMacro{fig-verl-lev-prompt-pass-rustfmt-olmo32b-rl-mean-line}}{\UseMacro{fig-verl-lev-prompt-pass-rustfmt-olmo32b-prompt-mean-line}}\UseMacro{fig-verl-lev-prompt-pass-rustfmt-olmo32b-small} \\
\bottomrule
\end{tabular}
\caption{\UseMacro{TCap-verl-lev-raw-rustfmt-eval-rust-repos}}
\label{tab:verl-lev-raw-rustfmt-eval-rust-repos}
\end{table*}

\begin{table*}[h]
\centering
\small
\setlength{\tabcolsep}{3pt}
\begin{tabular}{l cc cc cc cc}
\toprule
\multirow{2}{*}{\shortstack[l]{\textbf{Implementor} \\ \quad\textbf{Editor}}} & \multicolumn{2}{c}{\shortstack{\textbf{All} \\ \textbf{\tasks}}} & \multicolumn{2}{c}{\shortstack{\textbf{Base \&} \\ \textbf{Strict pass}}} & \multicolumn{2}{c}{\shortstack{\textbf{Base \&} \\ \textbf{RL pass}}} & \multicolumn{2}{c}{\shortstack{\textbf{Strict \&} \\ \textbf{RL pass}}} \\
\cmidrule(lr){2-3} \cmidrule(lr){4-5} \cmidrule(lr){6-7} \cmidrule(lr){8-9}
& \textbf{Chars} & \textbf{Lines} & \textbf{Chars} & \textbf{Lines} & \textbf{Chars} & \textbf{Lines} & \textbf{Chars} & \textbf{Lines} \\
\midrule
\multicolumn{9}{@{}l}{\textbf{\qwen}} \\
\quad \baselabel & \SmallestBoldSafe{\UseMacro{fig-verl-comparison-allmean-rustfmt-minmax-base-qwen-lev-char}}{\UseMacro{fig-verl-comparison-allmean-rustfmt-minmax-prompt-qwen-lev-char}}{\UseMacro{fig-verl-comparison-allmean-rustfmt-minmax-rl-qwen-lev-char}} & \SmallestBoldSafe{\UseMacro{fig-verl-comparison-allmean-rustfmt-minmax-base-qwen-lev-line}}{\UseMacro{fig-verl-comparison-allmean-rustfmt-minmax-prompt-qwen-lev-line}}{\UseMacro{fig-verl-comparison-allmean-rustfmt-minmax-rl-qwen-lev-line}} & \SmallerBoldSafe{\UseMacro{fig-verl-lev-base-strict-rustfmt-minmax-base-qwen-lev-char}}{\UseMacro{fig-verl-lev-base-strict-rustfmt-minmax-prompt-qwen-lev-char}}\UseMacro{fig-verl-lev-base-strict-rustfmt-minmax-qwen-lev-char-small} & \SmallerBoldSafe{\UseMacro{fig-verl-lev-base-strict-rustfmt-minmax-base-qwen-lev-line}}{\UseMacro{fig-verl-lev-base-strict-rustfmt-minmax-prompt-qwen-lev-line}}\UseMacro{fig-verl-lev-base-strict-rustfmt-minmax-qwen-lev-line-small} & \SmallerBoldSafe{\UseMacro{fig-verl-lev-both-pass-rustfmt-minmax-base-qwen-lev-char}}{\UseMacro{fig-verl-lev-both-pass-rustfmt-minmax-rl-qwen-lev-char}}\UseMacro{fig-verl-lev-both-pass-rustfmt-minmax-qwen-lev-char-small} & \SmallerBoldSafe{\UseMacro{fig-verl-lev-both-pass-rustfmt-minmax-base-qwen-lev-line}}{\UseMacro{fig-verl-lev-both-pass-rustfmt-minmax-rl-qwen-lev-line}}\UseMacro{fig-verl-lev-both-pass-rustfmt-minmax-qwen-lev-line-small} & -- & -- \\
\quad \strictlabel & \SmallestBoldSafe{\UseMacro{fig-verl-comparison-allmean-rustfmt-minmax-prompt-qwen-lev-char}}{\UseMacro{fig-verl-comparison-allmean-rustfmt-minmax-base-qwen-lev-char}}{\UseMacro{fig-verl-comparison-allmean-rustfmt-minmax-rl-qwen-lev-char}} & \SmallestBoldSafe{\UseMacro{fig-verl-comparison-allmean-rustfmt-minmax-prompt-qwen-lev-line}}{\UseMacro{fig-verl-comparison-allmean-rustfmt-minmax-base-qwen-lev-line}}{\UseMacro{fig-verl-comparison-allmean-rustfmt-minmax-rl-qwen-lev-line}} & \SmallerBoldSafe{\UseMacro{fig-verl-lev-base-strict-rustfmt-minmax-prompt-qwen-lev-char}}{\UseMacro{fig-verl-lev-base-strict-rustfmt-minmax-base-qwen-lev-char}}\UseMacro{fig-verl-lev-base-strict-rustfmt-minmax-qwen-lev-char-small} & \SmallerBoldSafe{\UseMacro{fig-verl-lev-base-strict-rustfmt-minmax-prompt-qwen-lev-line}}{\UseMacro{fig-verl-lev-base-strict-rustfmt-minmax-base-qwen-lev-line}}\UseMacro{fig-verl-lev-base-strict-rustfmt-minmax-qwen-lev-line-small} & -- & -- & \SmallerBoldSafe{\UseMacro{fig-verl-lev-prompt-pass-rustfmt-minmax-prompt-qwen-lev-char}}{\UseMacro{fig-verl-lev-prompt-pass-rustfmt-minmax-rl-qwen-lev-char}}\UseMacro{fig-verl-lev-prompt-pass-rustfmt-minmax-qwen-lev-char-small} & \SmallerBoldSafe{\UseMacro{fig-verl-lev-prompt-pass-rustfmt-minmax-prompt-qwen-lev-line}}{\UseMacro{fig-verl-lev-prompt-pass-rustfmt-minmax-rl-qwen-lev-line}}\UseMacro{fig-verl-lev-prompt-pass-rustfmt-minmax-qwen-lev-line-small} \\
\quad \rllabel & \SmallestBoldSafe{\UseMacro{fig-verl-comparison-allmean-rustfmt-minmax-rl-qwen-lev-char}}{\UseMacro{fig-verl-comparison-allmean-rustfmt-minmax-base-qwen-lev-char}}{\UseMacro{fig-verl-comparison-allmean-rustfmt-minmax-prompt-qwen-lev-char}} & \SmallestBoldSafe{\UseMacro{fig-verl-comparison-allmean-rustfmt-minmax-rl-qwen-lev-line}}{\UseMacro{fig-verl-comparison-allmean-rustfmt-minmax-base-qwen-lev-line}}{\UseMacro{fig-verl-comparison-allmean-rustfmt-minmax-prompt-qwen-lev-line}} & -- & -- & \SmallerBoldSafe{\UseMacro{fig-verl-lev-both-pass-rustfmt-minmax-rl-qwen-lev-char}}{\UseMacro{fig-verl-lev-both-pass-rustfmt-minmax-base-qwen-lev-char}}\UseMacro{fig-verl-lev-both-pass-rustfmt-minmax-qwen-lev-char-small} & \SmallerBoldSafe{\UseMacro{fig-verl-lev-both-pass-rustfmt-minmax-rl-qwen-lev-line}}{\UseMacro{fig-verl-lev-both-pass-rustfmt-minmax-base-qwen-lev-line}}\UseMacro{fig-verl-lev-both-pass-rustfmt-minmax-qwen-lev-line-small} & \SmallerBoldSafe{\UseMacro{fig-verl-lev-prompt-pass-rustfmt-minmax-rl-qwen-lev-char}}{\UseMacro{fig-verl-lev-prompt-pass-rustfmt-minmax-prompt-qwen-lev-char}}\UseMacro{fig-verl-lev-prompt-pass-rustfmt-minmax-qwen-lev-char-small} & \SmallerBoldSafe{\UseMacro{fig-verl-lev-prompt-pass-rustfmt-minmax-rl-qwen-lev-line}}{\UseMacro{fig-verl-lev-prompt-pass-rustfmt-minmax-prompt-qwen-lev-line}}\UseMacro{fig-verl-lev-prompt-pass-rustfmt-minmax-qwen-lev-line-small} \\
\midrule
\multicolumn{9}{@{}l}{\textbf{\gpto}} \\
\quad \baselabel & \SmallestBoldSafe{\UseMacro{fig-verl-comparison-allmean-rustfmt-minmax-base-gpt-lev-char}}{\UseMacro{fig-verl-comparison-allmean-rustfmt-minmax-prompt-gpt-lev-char}}{\UseMacro{fig-verl-comparison-allmean-rustfmt-minmax-rl-gpt-lev-char}} & \SmallestBoldSafe{\UseMacro{fig-verl-comparison-allmean-rustfmt-minmax-base-gpt-lev-line}}{\UseMacro{fig-verl-comparison-allmean-rustfmt-minmax-prompt-gpt-lev-line}}{\UseMacro{fig-verl-comparison-allmean-rustfmt-minmax-rl-gpt-lev-line}} & \SmallerBoldSafe{\UseMacro{fig-verl-lev-base-strict-rustfmt-minmax-base-gpt-lev-char}}{\UseMacro{fig-verl-lev-base-strict-rustfmt-minmax-prompt-gpt-lev-char}}\UseMacro{fig-verl-lev-base-strict-rustfmt-minmax-gpt-lev-char-small} & \SmallerBoldSafe{\UseMacro{fig-verl-lev-base-strict-rustfmt-minmax-base-gpt-lev-line}}{\UseMacro{fig-verl-lev-base-strict-rustfmt-minmax-prompt-gpt-lev-line}}\UseMacro{fig-verl-lev-base-strict-rustfmt-minmax-gpt-lev-line-small} & \SmallerBoldSafe{\UseMacro{fig-verl-lev-both-pass-rustfmt-minmax-base-gpt-lev-char}}{\UseMacro{fig-verl-lev-both-pass-rustfmt-minmax-rl-gpt-lev-char}}\UseMacro{fig-verl-lev-both-pass-rustfmt-minmax-gpt-lev-char-small} & \SmallerBoldSafe{\UseMacro{fig-verl-lev-both-pass-rustfmt-minmax-base-gpt-lev-line}}{\UseMacro{fig-verl-lev-both-pass-rustfmt-minmax-rl-gpt-lev-line}}\UseMacro{fig-verl-lev-both-pass-rustfmt-minmax-gpt-lev-line-small} & -- & -- \\
\quad \strictlabel & \SmallestBoldSafe{\UseMacro{fig-verl-comparison-allmean-rustfmt-minmax-prompt-gpt-lev-char}}{\UseMacro{fig-verl-comparison-allmean-rustfmt-minmax-base-gpt-lev-char}}{\UseMacro{fig-verl-comparison-allmean-rustfmt-minmax-rl-gpt-lev-char}} & \SmallestBoldSafe{\UseMacro{fig-verl-comparison-allmean-rustfmt-minmax-prompt-gpt-lev-line}}{\UseMacro{fig-verl-comparison-allmean-rustfmt-minmax-base-gpt-lev-line}}{\UseMacro{fig-verl-comparison-allmean-rustfmt-minmax-rl-gpt-lev-line}} & \SmallerBoldSafe{\UseMacro{fig-verl-lev-base-strict-rustfmt-minmax-prompt-gpt-lev-char}}{\UseMacro{fig-verl-lev-base-strict-rustfmt-minmax-base-gpt-lev-char}}\UseMacro{fig-verl-lev-base-strict-rustfmt-minmax-gpt-lev-char-small} & \SmallerBoldSafe{\UseMacro{fig-verl-lev-base-strict-rustfmt-minmax-prompt-gpt-lev-line}}{\UseMacro{fig-verl-lev-base-strict-rustfmt-minmax-base-gpt-lev-line}}\UseMacro{fig-verl-lev-base-strict-rustfmt-minmax-gpt-lev-line-small} & -- & -- & \SmallerBoldSafe{\UseMacro{fig-verl-lev-prompt-pass-rustfmt-minmax-prompt-gpt-lev-char}}{\UseMacro{fig-verl-lev-prompt-pass-rustfmt-minmax-rl-gpt-lev-char}}\UseMacro{fig-verl-lev-prompt-pass-rustfmt-minmax-gpt-lev-char-small} & \SmallerBoldSafe{\UseMacro{fig-verl-lev-prompt-pass-rustfmt-minmax-prompt-gpt-lev-line}}{\UseMacro{fig-verl-lev-prompt-pass-rustfmt-minmax-rl-gpt-lev-line}}\UseMacro{fig-verl-lev-prompt-pass-rustfmt-minmax-gpt-lev-line-small} \\
\quad \rllabel & \SmallestBoldSafe{\UseMacro{fig-verl-comparison-allmean-rustfmt-minmax-rl-gpt-lev-char}}{\UseMacro{fig-verl-comparison-allmean-rustfmt-minmax-base-gpt-lev-char}}{\UseMacro{fig-verl-comparison-allmean-rustfmt-minmax-prompt-gpt-lev-char}} & \SmallestBoldSafe{\UseMacro{fig-verl-comparison-allmean-rustfmt-minmax-rl-gpt-lev-line}}{\UseMacro{fig-verl-comparison-allmean-rustfmt-minmax-base-gpt-lev-line}}{\UseMacro{fig-verl-comparison-allmean-rustfmt-minmax-prompt-gpt-lev-line}} & -- & -- & \SmallerBoldSafe{\UseMacro{fig-verl-lev-both-pass-rustfmt-minmax-rl-gpt-lev-char}}{\UseMacro{fig-verl-lev-both-pass-rustfmt-minmax-base-gpt-lev-char}}\UseMacro{fig-verl-lev-both-pass-rustfmt-minmax-gpt-lev-char-small} & \SmallerBoldSafe{\UseMacro{fig-verl-lev-both-pass-rustfmt-minmax-rl-gpt-lev-line}}{\UseMacro{fig-verl-lev-both-pass-rustfmt-minmax-base-gpt-lev-line}}\UseMacro{fig-verl-lev-both-pass-rustfmt-minmax-gpt-lev-line-small} & \SmallerBoldSafe{\UseMacro{fig-verl-lev-prompt-pass-rustfmt-minmax-rl-gpt-lev-char}}{\UseMacro{fig-verl-lev-prompt-pass-rustfmt-minmax-prompt-gpt-lev-char}}\UseMacro{fig-verl-lev-prompt-pass-rustfmt-minmax-gpt-lev-char-small} & \SmallerBoldSafe{\UseMacro{fig-verl-lev-prompt-pass-rustfmt-minmax-rl-gpt-lev-line}}{\UseMacro{fig-verl-lev-prompt-pass-rustfmt-minmax-prompt-gpt-lev-line}}\UseMacro{fig-verl-lev-prompt-pass-rustfmt-minmax-gpt-lev-line-small} \\
\midrule
\multicolumn{9}{@{}l}{\textbf{\olmos}} \\
\quad \baselabel & \SmallestBoldSafe{\UseMacro{fig-verl-comparison-allmean-rustfmt-minmax-base-olmo7b-lev-char}}{\UseMacro{fig-verl-comparison-allmean-rustfmt-minmax-prompt-olmo7b-lev-char}}{\UseMacro{fig-verl-comparison-allmean-rustfmt-minmax-rl-olmo7b-lev-char}} & \SmallestBoldSafe{\UseMacro{fig-verl-comparison-allmean-rustfmt-minmax-base-olmo7b-lev-line}}{\UseMacro{fig-verl-comparison-allmean-rustfmt-minmax-prompt-olmo7b-lev-line}}{\UseMacro{fig-verl-comparison-allmean-rustfmt-minmax-rl-olmo7b-lev-line}} & \SmallerBoldSafe{\UseMacro{fig-verl-lev-base-strict-rustfmt-minmax-base-olmo7b-lev-char}}{\UseMacro{fig-verl-lev-base-strict-rustfmt-minmax-prompt-olmo7b-lev-char}}\UseMacro{fig-verl-lev-base-strict-rustfmt-minmax-olmo7b-lev-char-small} & \SmallerBoldSafe{\UseMacro{fig-verl-lev-base-strict-rustfmt-minmax-base-olmo7b-lev-line}}{\UseMacro{fig-verl-lev-base-strict-rustfmt-minmax-prompt-olmo7b-lev-line}}\UseMacro{fig-verl-lev-base-strict-rustfmt-minmax-olmo7b-lev-line-small} & \SmallerBoldSafe{\UseMacro{fig-verl-lev-both-pass-rustfmt-minmax-base-olmo7b-lev-char}}{\UseMacro{fig-verl-lev-both-pass-rustfmt-minmax-rl-olmo7b-lev-char}}\UseMacro{fig-verl-lev-both-pass-rustfmt-minmax-olmo7b-lev-char-small} & \SmallerBoldSafe{\UseMacro{fig-verl-lev-both-pass-rustfmt-minmax-base-olmo7b-lev-line}}{\UseMacro{fig-verl-lev-both-pass-rustfmt-minmax-rl-olmo7b-lev-line}}\UseMacro{fig-verl-lev-both-pass-rustfmt-minmax-olmo7b-lev-line-small} & -- & -- \\
\quad \strictlabel & \SmallestBoldSafe{\UseMacro{fig-verl-comparison-allmean-rustfmt-minmax-prompt-olmo7b-lev-char}}{\UseMacro{fig-verl-comparison-allmean-rustfmt-minmax-base-olmo7b-lev-char}}{\UseMacro{fig-verl-comparison-allmean-rustfmt-minmax-rl-olmo7b-lev-char}} & \SmallestBoldSafe{\UseMacro{fig-verl-comparison-allmean-rustfmt-minmax-prompt-olmo7b-lev-line}}{\UseMacro{fig-verl-comparison-allmean-rustfmt-minmax-base-olmo7b-lev-line}}{\UseMacro{fig-verl-comparison-allmean-rustfmt-minmax-rl-olmo7b-lev-line}} & \SmallerBoldSafe{\UseMacro{fig-verl-lev-base-strict-rustfmt-minmax-prompt-olmo7b-lev-char}}{\UseMacro{fig-verl-lev-base-strict-rustfmt-minmax-base-olmo7b-lev-char}}\UseMacro{fig-verl-lev-base-strict-rustfmt-minmax-olmo7b-lev-char-small} & \SmallerBoldSafe{\UseMacro{fig-verl-lev-base-strict-rustfmt-minmax-prompt-olmo7b-lev-line}}{\UseMacro{fig-verl-lev-base-strict-rustfmt-minmax-base-olmo7b-lev-line}}\UseMacro{fig-verl-lev-base-strict-rustfmt-minmax-olmo7b-lev-line-small} & -- & -- & \SmallerBoldSafe{\UseMacro{fig-verl-lev-prompt-pass-rustfmt-minmax-prompt-olmo7b-lev-char}}{\UseMacro{fig-verl-lev-prompt-pass-rustfmt-minmax-rl-olmo7b-lev-char}}\UseMacro{fig-verl-lev-prompt-pass-rustfmt-minmax-olmo7b-lev-char-small} & \SmallerBoldSafe{\UseMacro{fig-verl-lev-prompt-pass-rustfmt-minmax-prompt-olmo7b-lev-line}}{\UseMacro{fig-verl-lev-prompt-pass-rustfmt-minmax-rl-olmo7b-lev-line}}\UseMacro{fig-verl-lev-prompt-pass-rustfmt-minmax-olmo7b-lev-line-small} \\
\quad \rllabel & \SmallestBoldSafe{\UseMacro{fig-verl-comparison-allmean-rustfmt-minmax-rl-olmo7b-lev-char}}{\UseMacro{fig-verl-comparison-allmean-rustfmt-minmax-base-olmo7b-lev-char}}{\UseMacro{fig-verl-comparison-allmean-rustfmt-minmax-prompt-olmo7b-lev-char}} & \SmallestBoldSafe{\UseMacro{fig-verl-comparison-allmean-rustfmt-minmax-rl-olmo7b-lev-line}}{\UseMacro{fig-verl-comparison-allmean-rustfmt-minmax-base-olmo7b-lev-line}}{\UseMacro{fig-verl-comparison-allmean-rustfmt-minmax-prompt-olmo7b-lev-line}} & -- & -- & \SmallerBoldSafe{\UseMacro{fig-verl-lev-both-pass-rustfmt-minmax-rl-olmo7b-lev-char}}{\UseMacro{fig-verl-lev-both-pass-rustfmt-minmax-base-olmo7b-lev-char}}\UseMacro{fig-verl-lev-both-pass-rustfmt-minmax-olmo7b-lev-char-small} & \SmallerBoldSafe{\UseMacro{fig-verl-lev-both-pass-rustfmt-minmax-rl-olmo7b-lev-line}}{\UseMacro{fig-verl-lev-both-pass-rustfmt-minmax-base-olmo7b-lev-line}}\UseMacro{fig-verl-lev-both-pass-rustfmt-minmax-olmo7b-lev-line-small} & \SmallerBoldSafe{\UseMacro{fig-verl-lev-prompt-pass-rustfmt-minmax-rl-olmo7b-lev-char}}{\UseMacro{fig-verl-lev-prompt-pass-rustfmt-minmax-prompt-olmo7b-lev-char}}\UseMacro{fig-verl-lev-prompt-pass-rustfmt-minmax-olmo7b-lev-char-small} & \SmallerBoldSafe{\UseMacro{fig-verl-lev-prompt-pass-rustfmt-minmax-rl-olmo7b-lev-line}}{\UseMacro{fig-verl-lev-prompt-pass-rustfmt-minmax-prompt-olmo7b-lev-line}}\UseMacro{fig-verl-lev-prompt-pass-rustfmt-minmax-olmo7b-lev-line-small} \\
\midrule
\multicolumn{9}{@{}l}{\textbf{\olmol}} \\
\quad \baselabel & \SmallestBoldSafe{\UseMacro{fig-verl-comparison-allmean-rustfmt-minmax-base-olmo32b-lev-char}}{\UseMacro{fig-verl-comparison-allmean-rustfmt-minmax-prompt-olmo32b-lev-char}}{\UseMacro{fig-verl-comparison-allmean-rustfmt-minmax-rl-olmo32b-lev-char}} & \SmallestBoldSafe{\UseMacro{fig-verl-comparison-allmean-rustfmt-minmax-base-olmo32b-lev-line}}{\UseMacro{fig-verl-comparison-allmean-rustfmt-minmax-prompt-olmo32b-lev-line}}{\UseMacro{fig-verl-comparison-allmean-rustfmt-minmax-rl-olmo32b-lev-line}} & \SmallerBoldSafe{\UseMacro{fig-verl-lev-base-strict-rustfmt-minmax-base-olmo32b-lev-char}}{\UseMacro{fig-verl-lev-base-strict-rustfmt-minmax-prompt-olmo32b-lev-char}}\UseMacro{fig-verl-lev-base-strict-rustfmt-minmax-olmo32b-lev-char-small} & \SmallerBoldSafe{\UseMacro{fig-verl-lev-base-strict-rustfmt-minmax-base-olmo32b-lev-line}}{\UseMacro{fig-verl-lev-base-strict-rustfmt-minmax-prompt-olmo32b-lev-line}}\UseMacro{fig-verl-lev-base-strict-rustfmt-minmax-olmo32b-lev-line-small} & \SmallerBoldSafe{\UseMacro{fig-verl-lev-both-pass-rustfmt-minmax-base-olmo32b-lev-char}}{\UseMacro{fig-verl-lev-both-pass-rustfmt-minmax-rl-olmo32b-lev-char}}\UseMacro{fig-verl-lev-both-pass-rustfmt-minmax-olmo32b-lev-char-small} & \SmallerBoldSafe{\UseMacro{fig-verl-lev-both-pass-rustfmt-minmax-base-olmo32b-lev-line}}{\UseMacro{fig-verl-lev-both-pass-rustfmt-minmax-rl-olmo32b-lev-line}}\UseMacro{fig-verl-lev-both-pass-rustfmt-minmax-olmo32b-lev-line-small} & -- & -- \\
\quad \strictlabel & \SmallestBoldSafe{\UseMacro{fig-verl-comparison-allmean-rustfmt-minmax-prompt-olmo32b-lev-char}}{\UseMacro{fig-verl-comparison-allmean-rustfmt-minmax-base-olmo32b-lev-char}}{\UseMacro{fig-verl-comparison-allmean-rustfmt-minmax-rl-olmo32b-lev-char}} & \SmallestBoldSafe{\UseMacro{fig-verl-comparison-allmean-rustfmt-minmax-prompt-olmo32b-lev-line}}{\UseMacro{fig-verl-comparison-allmean-rustfmt-minmax-base-olmo32b-lev-line}}{\UseMacro{fig-verl-comparison-allmean-rustfmt-minmax-rl-olmo32b-lev-line}} & \SmallerBoldSafe{\UseMacro{fig-verl-lev-base-strict-rustfmt-minmax-prompt-olmo32b-lev-char}}{\UseMacro{fig-verl-lev-base-strict-rustfmt-minmax-base-olmo32b-lev-char}}\UseMacro{fig-verl-lev-base-strict-rustfmt-minmax-olmo32b-lev-char-small} & \SmallerBoldSafe{\UseMacro{fig-verl-lev-base-strict-rustfmt-minmax-prompt-olmo32b-lev-line}}{\UseMacro{fig-verl-lev-base-strict-rustfmt-minmax-base-olmo32b-lev-line}}\UseMacro{fig-verl-lev-base-strict-rustfmt-minmax-olmo32b-lev-line-small} & -- & -- & \SmallerBoldSafe{\UseMacro{fig-verl-lev-prompt-pass-rustfmt-minmax-prompt-olmo32b-lev-char}}{\UseMacro{fig-verl-lev-prompt-pass-rustfmt-minmax-rl-olmo32b-lev-char}}\UseMacro{fig-verl-lev-prompt-pass-rustfmt-minmax-olmo32b-lev-char-small} & \SmallerBoldSafe{\UseMacro{fig-verl-lev-prompt-pass-rustfmt-minmax-prompt-olmo32b-lev-line}}{\UseMacro{fig-verl-lev-prompt-pass-rustfmt-minmax-rl-olmo32b-lev-line}}\UseMacro{fig-verl-lev-prompt-pass-rustfmt-minmax-olmo32b-lev-line-small} \\
\quad \rllabel & \SmallestBoldSafe{\UseMacro{fig-verl-comparison-allmean-rustfmt-minmax-rl-olmo32b-lev-char}}{\UseMacro{fig-verl-comparison-allmean-rustfmt-minmax-base-olmo32b-lev-char}}{\UseMacro{fig-verl-comparison-allmean-rustfmt-minmax-prompt-olmo32b-lev-char}} & \SmallestBoldSafe{\UseMacro{fig-verl-comparison-allmean-rustfmt-minmax-rl-olmo32b-lev-line}}{\UseMacro{fig-verl-comparison-allmean-rustfmt-minmax-base-olmo32b-lev-line}}{\UseMacro{fig-verl-comparison-allmean-rustfmt-minmax-prompt-olmo32b-lev-line}} & -- & -- & \SmallerBoldSafe{\UseMacro{fig-verl-lev-both-pass-rustfmt-minmax-rl-olmo32b-lev-char}}{\UseMacro{fig-verl-lev-both-pass-rustfmt-minmax-base-olmo32b-lev-char}}\UseMacro{fig-verl-lev-both-pass-rustfmt-minmax-olmo32b-lev-char-small} & \SmallerBoldSafe{\UseMacro{fig-verl-lev-both-pass-rustfmt-minmax-rl-olmo32b-lev-line}}{\UseMacro{fig-verl-lev-both-pass-rustfmt-minmax-base-olmo32b-lev-line}}\UseMacro{fig-verl-lev-both-pass-rustfmt-minmax-olmo32b-lev-line-small} & \SmallerBoldSafe{\UseMacro{fig-verl-lev-prompt-pass-rustfmt-minmax-rl-olmo32b-lev-char}}{\UseMacro{fig-verl-lev-prompt-pass-rustfmt-minmax-prompt-olmo32b-lev-char}}\UseMacro{fig-verl-lev-prompt-pass-rustfmt-minmax-olmo32b-lev-char-small} & \SmallerBoldSafe{\UseMacro{fig-verl-lev-prompt-pass-rustfmt-minmax-rl-olmo32b-lev-line}}{\UseMacro{fig-verl-lev-prompt-pass-rustfmt-minmax-prompt-olmo32b-lev-line}}\UseMacro{fig-verl-lev-prompt-pass-rustfmt-minmax-olmo32b-lev-line-small} \\
\bottomrule
\end{tabular}
\caption{\UseMacro{TCap-verl-lev-normfull-rustfmt-eval-rust-repos}}
\label{tab:verl-lev-normfull-rustfmt-eval-rust-repos}
\end{table*}

\clearpage

\section{Per-function Context Items}
\label{sec:app-context-items-example}
Here, we include an example of all the the context items we listed in
Table~\ref{tab:context-items}. All prompts we used in this work, including
\reimplementing, editing, repairing, utilize a subset of the shown context
items.

Our example \PR is \texttt{rust-lang}/\texttt{flate2-rs}\#484, which targets the function
\texttt{read} in the source file \texttt{src}/\texttt{gz}/\texttt{bufread.rs}.

\medskip

\bigskip
\noindent\textbf{\IDFuncsig (\Funcsig).}
\begin{lstlisting}[style=ContextSnippet]
fn read(&mut self, into: &mut [u8]) -> io::Result<usize> {
\end{lstlisting}

\bigskip
\noindent\textbf{\IDRestprdiff (\Restprdiff).}
\smallskip
\begin{lstlisting}[style=ContextSnippet]
diff --git a/src/gz/mod.rs b/src/gz/mod.rs
index 3a24a754..79b957d6 100644
--- a/src/gz/mod.rs
+++ b/src/gz/mod.rs
@@ -402,8 +402,7 @@ impl GzBuilder {
         let mut header = vec![0u8; 10];
         if let Some(v) = extra {
             flg |= FEXTRA;
-            header.push((v.len() >> 0) as u8);
-            header.push((v.len() >> 8) as u8);
+            header.extend((v.len() as u16).to_le_bytes());
             header.extend(v);
         }
         if let Some(filename) = filename {
\end{lstlisting}

\bigskip
\noindent\textbf{\IDCallees (\Callees).}
\smallskip
\begin{lstlisting}[style=ContextSnippet]
// callee 1: src/gz/mod.rs
fn parse<R: BufRead>(&mut self, r: &mut R) -> Result<()> {
        loop {
            match &mut self.state {
                GzHeaderState::Start(count, buffer) => {
                    while (*count as usize) < buffer.len() {
                        *count += read_into(r, &mut buffer[*count as usize..])? as u8;
                    }
                    // Gzip identification bytes
                    if buffer[0] != 0x1f || buffer[1] != 0x8b {
                        return Err(bad_header());
                    }
                    // Gzip compression method (8 = deflate)
                    if buffer[2] != 8 {
                        return Err(bad_header());
                    }
...
(callee truncated for space)

// callee 2: src/deflate/bufread.rs
    /// Acquires a mutable reference to the underlying stream
    ///
    /// Note that mutation of the stream may result in surprising results if
    /// this decoder is continued to be used.
    pub fn get_mut(&mut self) -> &mut R {
        &mut self.obj
    }

// callee 3: src/crc.rs
    /// Get a mutable reference to the reader that is wrapped by this `CrcReader`.
    pub fn get_mut(&mut self) -> &mut R {
        &mut self.inner
    }


// ... 25 additional callees not shown
\end{lstlisting}

\bigskip
\noindent\textbf{\IDCallsites (\Callsites).}
\smallskip
\begin{lstlisting}[style=ContextSnippet]
src/gz/bufread.rs
   428 |     }
   429 | }
   430 | 
   431 | impl<R: BufRead> Read for MultiGzDecoder<R> {
   432 |     fn read(&mut self, into: &mut [u8]) -> io::Result<usize> {
   433 |         self.0.read(into)
   434 |     }
   435 | }
   436 | 
   437 | #[cfg(test)]
   438 | mod test {

src/gz/bufread.rs
   465 |             "after decompression we obtain the original input"
   466 |         );
   467 | 
   468 |         output.clear();
   469 |         assert_eq!(
   470 |             decoder.read(&mut output).unwrap(),
   471 |             0,
   472 |             "subsequent read of decoder returns 0, but inner reader can return additional data"
   473 |         );
   474 |         let mut reader = decoder.into_inner();
   475 |         assert_eq!(
\end{lstlisting}

\bigskip
\noindent\textbf{\IDUsestmts (\Usestmts).}
\smallskip
\begin{lstlisting}[style=ContextSnippet]
use std::cmp;
use std::io;
use std::io::prelude::*;
use std::mem;
use super::{corrupt, read_into, GzBuilder, GzHeader, GzHeaderParser};
use crate::crc::CrcReader;
use crate::deflate;
use crate::Compression;
\end{lstlisting}

\bigskip
\noindent\textbf{\IDStructimpl (\Structimpl).}
\smallskip
\begin{lstlisting}[style=ContextSnippet]
impl <R: BufRead> Read for GzDecoder<R>
\end{lstlisting}

\bigskip
\noindent\textbf{\IDRepodesc (\Repodesc).}
\smallskip
\begin{lstlisting}[style=ContextSnippet]
This is a Rust library for streaming compression and decompression based on DEFLATE. It primarily supports gzip, zlib, and deflate formats. The library defaults to using `miniz_oxide` but also supports C-based backends like zlib and zlib-ng.
\end{lstlisting}

\bigskip
\noindent\textbf{\IDPRdesc (\PRdesc).}
\smallskip
\begin{lstlisting}[style=ContextSnippet]
This PR simplifies code in `src/gz/bufread.rs` and `src/gz/mod.rs` by removing unnecessary borrows and manual bit operations. In `GzDecoder::read`, the call to `finish` is updated from `finish(&buf)` to `finish(buf)` since the function signature already accepts a reference. In `GzBuilder`, manual length extraction via bit shifting is replaced with `header.extend((v.len() as u16).to_le_bytes())`, which is more idiomatic. The author notes that LLVM will optimize both versions identically. There is no functional change to the library's behavior, only improvements to code clarity and reduction of syntactic noise.
\end{lstlisting}

\bigskip
\noindent\textbf{\IDFuncdesc (\Funcdesc).}
\smallskip
\begin{lstlisting}[style=ContextSnippet]
## Function Overview

This function implements the `read` method for a gzip decompressor. Its purpose is to decompress gzip-compressed data from an inner reader and write the decompressed bytes into a provided output buffer. It returns the number of bytes successfully written to the output buffer, or an error if decompression fails.

## Inputs and Outputs

**Input:**
- `into`: A mutable slice of bytes where the decompressed data will be written

**Output:**
- On success: Returns the number of bytes written to the output buffer
- On error: Returns an error indicating decompression failure (e.g., corrupted data, invalid header)

## Behavior

The function uses a state machine to manage the decompression process through several stages:

### 1. Header Parsing State
When starting a new gzip stream, the function first parses the gzip header from the underlying reader. This header contains metadata about the compression format. Once parsing completes, the state transitions to the Body stage.

### 2. Body Reading State
In this stage, the function reads decompressed data from the inner reader:
- If the output buffer is empty, it returns 0 immediately
- It reads data from the inner reader into the output buffer
- If no more data is available (end of stream), it transitions to the Finished state
- Otherwise, it returns the number of bytes read

### 3. Finished State
After the body data is fully read, the function verifies data integrity:
- It reads any remaining buffered data from the inner reader
- It computes and verifies the CRC checksum against the expected value stored in the gzip stream
- If the checksum verification fails, it transitions to an error state
- If the `multi` flag is enabled (support for concatenated gzip streams), it checks whether more data follows
  - If more data exists, it resets and returns to the Header state to process the next stream
  - If no more data exists, it transitions to the End state
- If `multi` is disabled, it transitions to the End state after verification

### 4. Error State
If any error occurred during processing, the function returns the error and transitions to the End state.

### 5. End State
When the decompressor has finished processing all available data, subsequent read requests return 0 bytes to indicate end of stream.

## Key Features

- **State Management**: Uses a state machine to track the current phase of decompression (header, body, finished, error, or end)
- **Checksum Verification**: Validates data integrity using CRC checksums before finalizing a stream
- **Multi-Stream Support**: When enabled, can process multiple concatenated gzip streams sequentially
- **Buffer Handling**: Efficiently manages reading from the inner reader into the output buffer
- **Graceful EOF**: Returns 0 bytes when no more decompressed data is available
\end{lstlisting}

\clearpage

\section{Repair Plan Example}
\label{sec:app-instruct-plan-example}
Figure~\ref{fig:instruct-plan-example} shows one failing candidate
(Figure~\ref{fig:instruct-plan-example-candidate}) and the repair plan the
\instagent wrote for it (Figure~\ref{fig:instruct-plan-example-plan}). In the plan, the agent simply described the visibility annotation bug and the fix needed without leaking any stylistic information.

\begin{figure}[h]
\begin{subfigure}{\columnwidth}
\begin{lstlisting}[style=ContextSnippet,language=rust-pretty,numbersep=6pt]
pub fn fmt(&self, f: &mut fmt::Formatter) -> fmt::Result {
    match self {
        ParseError::GeneralParseError(s) => write!(f, "General parsing error: {}", s),
        ParseError::UnclosedOpenParen => write!(f, "Unclosed opening parenthesis"),
        ParseError::InvalidRepeat => write!(f, "Invalid repeat syntax"),
        ParseError::RecursionExceeded => write!(f, "Pattern is nested too deeply"),
        ...
        _ => write!(f, "Unknown error"),
    }
}
\end{lstlisting}
\caption{\UseMacro{FCap-instruct-plan-example-candidate}}
\label{fig:instruct-plan-example-candidate}
\end{subfigure}

\vspace{4pt}

\begin{subfigure}{\columnwidth}
\begin{lstlisting}[style=ContextSnippet]
The function incorrectly declares a visibility modifier on a trait method implementation. Trait methods like `fmt` inherit their visibility from the trait itself and cannot have explicit visibility annotations. Remove the visibility qualifier from the method definition to resolve the compilation error. This adjustment ensures the method's visibility aligns with the trait's requirements.
\end{lstlisting}
\caption{\UseMacro{FCap-instruct-plan-example-plan}}
\label{fig:instruct-plan-example-plan}
\end{subfigure}
\vspace{-8pt}
\caption{\UseMacro{FCap-instruct-plan-example}}
\label{fig:instruct-plan-example}
\end{figure}

\section{Guidelines for the \Minprompt}
\label{sec:app-edit-limit-prompt}
Here, we reproduce in Figure~\ref{fig:prompt-edit-limit-guidelines} the
guideline list that the \minprompt of Section~\ref{sec:experiment-baselines}
adds to the \editor system prompt. The remainder of that system prompt states
the same task, output format, and other rules as the plain
\editor system prompt that every other condition uses, so those guidelines are
the entire difference between the \minprompt and the \basemodel.

\section{Qualitative Examples}
\label{sec:app-qualitative}
Here we include the examples discussed in our qualitative analysis in
Section~\ref{sec:results-qualitative}.

\begin{figure}[h]
\begin{subfigure}{\columnwidth}
\begin{lstlisting}[language=rust-diff]
+(*@\label{line:endianness-base-enum}@*)pub enum Architecture {
+    Bpfel, Hexagon, X86_32(X86_32Architecture),
+    M68k, Mips32(Mips32Architecture),
+(*@\label{line:endianness-base-invalid}@*)    Riscv32(_), Riscv64(_),
+    X86_32(_), X86_64, XTensa,
+}
 pub fn endianness(self) -> Result<Endianness, ()> {
     ...
 }
\end{lstlisting}
\caption{Base edit.}
\label{fig:scopedrift-endianness-base}
\end{subfigure}
\begin{subfigure}{\columnwidth}
\vspace{4mm}
\begin{lstlisting}[language=rust-diff]
 pub fn endianness(self) -> Result<Endianness, ()> {
     match self {
         Architecture::Unknown    => Err(()),
         Architecture::Arm(arm)   => Ok(arm.endianness()),
         Architecture::Aarch64(a) => Ok(a.endianness()),
+(*@\label{line:endianness-rl-arm}@*)        Architecture::M68k       => Ok(Endianness::Little),
         ...
     }
 }
\end{lstlisting}
\caption{RL edit.}
\label{fig:scopedrift-endianness-rl}
\end{subfigure}
\caption{Scope drift on \texttt{endianness} (\texttt{bytecode} \texttt{alliance}/\texttt{target}-\texttt{lexicon} PR \#79, \qwen \rewriter).  The \basemodel prefixes a fabricated enum definition with invalid pattern syntax; the \rlmodel writes only the one new match arm.}
\label{fig:scopedrift-endianness}
\vspace{6.5mm}
\end{figure}

\begin{figure}[h]
\begin{subfigure}{\columnwidth}
\begin{lstlisting}[language=rust-diff]
-fn cooked_byte_string(mut input: Cursor) -> Result<Cursor, LexError> {
+(*@\label{line:cbs-base-sig}@*)fn cooked_byte_string(mut input: Cursor) -> Result<Cursor, Reject> {
     ...
-    _ => return Err(LexError),
+(*@\label{line:cbs-base-err}@*)    _ => return Err(Reject),
     ...
 }
\end{lstlisting}
\caption{Base edit.}
\label{fig:rename-cookedbytestring-base}
\end{subfigure}
\begin{subfigure}{\columnwidth}
\vspace{4mm}
\begin{lstlisting}[language=rust-diff]
 (*@\label{line:cbs-rl-sig}@*)fn cooked_byte_string(mut input: Cursor) -> Result<Cursor, LexError> {
     ...
     match next_byte {
-        b'n' | b'r' | b't' | b'\\' | b'\'' | b'"' | b'0' => continue,
+(*@\label{line:cbs-rl-lit}@*)        b'n' | b'r' | b't' | b'\\' | b'\\' | b'\' | b'"' | b'0' => continue,
         _ => return Err(LexError),
     }
 }
\end{lstlisting}
\caption{RL edit.}
\label{fig:rename-cookedbytestring-rl}
\end{subfigure}
\caption{Partial rename on \texttt{cooked}\texttt{\_byte}\texttt{\_string} (\texttt{dtolnay}/\texttt{proc-macro2} PR \#282, \qwen \rewriter).  The PR renames \texttt{LexError} to \texttt{Reject}; the \rlmodel keeps the old return type and additionally introduces an unterminated character literal.}
\label{fig:rename-cookedbytestring}
\end{figure}

In Figure~\ref{fig:scopedrift-endianness}, the \task is one new
\CodeIn{match} arm. The \rlmodel writes that arm and nothing else,
while the \basemodel adds a fabricated \CodeIn{enum} that does not
build. In Figure~\ref{fig:rename-cookedbytestring}, the \task is to
propagate a rename of \CodeIn{LexError} to \CodeIn{Reject}. The
\basemodel applies it, while the \rlmodel leaves the old type as is
and breaks an unrelated literal instead.

\begin{figure*}[t]
\noindent\rule{\linewidth}{\heavyrulewidth}
\vspace{-4pt}
\begin{lstlisting}[style=ContextSnippet]
## Guidelines

**1. Scope** --- Change only what the fix requires. No unrelated refactors, no speculative additions, no global cleanup.

**2. Control Flow** --- Preserve existing branching, loop structures, statement order, and nesting. Don't convert between `if/else` and `match`, restructure loops, or add/remove early returns without functional reason.

**3. Types & Variables** --- Don't rename, re-sign, or restructure types/variables cosmetically. Preserve mutability, ownership, and existing access patterns. Avoid new aliases or redundant intermediates.

**4. Safety** --- Keep `unsafe` minimal and only where already required. No uninitialized memory, dummy allocations, or unsafe fallbacks in disabled `cfg` branches.

**5. Configuration** --- Preserve existing `#[cfg]` attributes and feature guards. Restrict changes to fields/blocks directly involved in the fix.

**6. API & Initialization** --- Use literals for fixed values. Retain all existing function arguments, trait bounds, and constructor patterns.

**7. Errors** --- Don't rewrite error messages, panic strings, or logs unless the error condition itself changed. Maintain existing error-handling style (`match`, `?`, `unwrap_or_else`).

**8. Formatting** --- Touch whitespace only on lines you functionally changed. No re-indenting, blank-line shuffling, or style normalization elsewhere.

**9. Comments** --- Preserve existing comments. Add new ones only for non-obvious *why* decisions. No TODOs, no meta-commentary about the diff.

**10. Before Submitting** --- Review every changed line: is it necessary for the fix? If not, revert it.
\end{lstlisting}
\vspace{-4pt}
\noindent\rule{\linewidth}{\heavyrulewidth}
\vspace{-8pt}
\caption{\UseMacro{FCap-prompt-edit-limit-guidelines}}
\label{fig:prompt-edit-limit-guidelines}
\end{figure*}

\end{document}